\documentclass{ieeeaccess}

\usepackage[dvipsnames]{xcolor}
\usepackage{moreverb,url}

\usepackage[utf8]{inputenc}
\usepackage{amsmath}
\usepackage{mathtools}
\usepackage{etoolbox}
\makeatletter
\patchcmd{\@IEEEyesnumber}
  {\stepcounter}
  {\refstepcounter}
  {}{}
\patchcmd{\@@IEEEeqnarray}
  {\stepcounter}
  {\refstepcounter}
  {}{}
\patchcmd{\@@IEEEeqnarraycr}
  {\stepcounter{IEEEsubequation}}
  {\refstepcounter{IEEEsubequation}}
  {}{}
\patchcmd{\@@IEEEeqnarraycr}
  {\stepcounter{IEEEsubequation}}
  {\refstepcounter{IEEEsubequation}}
  {}{}
\patchcmd{\@@IEEEeqnarraycr}
  {\stepcounter{IEEEequation}}
  {\refstepcounter{IEEEequation}}
  {}{}
\patchcmd{\@@IEEEeqnarraycr}
  {\stepcounter{IEEEequation}}
  {\refstepcounter{IEEEequation}}
  {}{}
\makeatother

\usepackage[colorlinks,linkcolor=black, citecolor = black, urlcolor = black,filecolor=black , pagebackref=false,hypertexnames=false, plainpages=false, ]{hyperref}

\usepackage[capitalize]{cleveref}
\crefformat{equation}{(#2#1#3)}
\Crefformat{equation}{Equation~(#2#1#3)}
\Crefname{equation}{Equation}{Equations}
\crefname{algocf}{Algorithm}{Algorithms}
\Crefname{algocf}{Algorithm}{Algorithms}

\newcommand{\citep}{\cite}

\usepackage{graphicx} 
\usepackage{amsfonts}
\usepackage{amsmath,soul}
\usepackage[font=small]{subcaption}
\usepackage{balance}
\usepackage[font=small]{caption}
\usepackage{bbm}

\usepackage{tikz}
\NewSpotColorSpace{PANTONE}
\AddSpotColor{PANTONE} {PANTONE3015C} {PANTONE\SpotSpace 3015\SpotSpace C} {1 0.3 0 0.2}
\SetPageColorSpace{PANTONE}
\usetikzlibrary{positioning}
\usetikzlibrary{shapes,arrows,fit,backgrounds}
\tikzstyle{block} = [draw, rectangle, text width=2cm, text centered, minimum height=1.2cm, node distance=3cm]

\usepackage{algorithm,algorithmic}
\usepackage{multirow}
\DeclareMathOperator*{\argmin}{\arg\!\min}
\DeclareMathOperator*{\argmax}{\arg\!\max}
\usepackage{tabulary}
\newcolumntype{K}[1]{>{\centering\arraybackslash}p{#1}}

\definecolor{darkgreen}{RGB}{34,139,34}

\usepackage[final]{changes} 
\definechangesauthor[name={J.~H.}, color={red}]{jh}
\definechangesauthor[name={M.~E.}, color={blue}]{me}
\definechangesauthor[name={S.~C.}, color={darkgreen}]{sc}

\usepackage{pifont}

\newcommand{\changed}[1]{#1}

\newcommand\BibTeX{{\rmfamily B\kern-.05em \textsc{i\kern-.025em b}\kern-.08em
T\kern-.1667em\lower.7ex\hbox{E}\kern-.125emX}}

\usepackage{etoolbox}
\newcounter{rowcntr}[table]
\renewcommand{\therowcntr}{\arabic{rowcntr}}
\newcolumntype{N}{>{\refstepcounter{rowcntr}\therowcntr}c}
\AtBeginEnvironment{tabular}{\setcounter{rowcntr}{0}}

\def\BibTeX{{\rm B\kern-.05em{\sc i\kern-.025em b}\kern-.08em
    T\kern-.1667em\lower.7ex\hbox{E}\kern-.125emX}}

\makeatletter
\newcommand{\removelatexerror}{\let\@latex@error\@gobble}
\makeatother

\begin{document}
\history{Received December 22, 2020, accepted January 6, 2021, date of publication January 8, 2021. This is the accepted manuscript; published version is available on IEEEXplore.}
\doi{10.1109/ACCESS.2021.3050338}



\title{Collision Avoidance in Pedestrian-Rich Environments with Deep Reinforcement Learning}
\author{\uppercase{Michael Everett}\authorrefmark{1},
\uppercase{Yu Fan Chen\authorrefmark{2}, and Jonathan P. How}\authorrefmark{1} \IEEEmembership{Fellow, IEEE}}
\address[1]{Massachusetts Institute of Technology, USA}
\address[2]{Facebook Reality Labs, USA (work done during PhD research at MIT)}
\tfootnote{This work is supported by Ford Motor Company, with computation support through Amazon Web Services.}

\markboth
{Everett \headeretal: Collision Avoidance in Pedestrian-Rich Environments with Deep Reinforcement Learning}
{Everett \headeretal: Collision Avoidance in Pedestrian-Rich Environments with Deep Reinforcement Learning}

\corresp{Corresponding author: Michael Everett (e-mail: mfe@ mit.edu).}





\begin{abstract}
Collision avoidance algorithms are essential for safe and efficient robot operation among pedestrians.
This work proposes using deep reinforcement (RL) learning as a framework to model the complex interactions and cooperation with nearby, decision-making agents, such as pedestrians and other robots.
Existing RL-based works assume homogeneity of agent properties, use specific motion models over short timescales, or lack a principled method to handle a large, possibly varying number of agents.
Therefore, this work develops an algorithm that learns collision avoidance among a variety of heterogeneous, non-communicating, dynamic agents without assuming they follow any particular behavior rules.
It extends our previous work by introducing a strategy using Long Short-Term Memory (LSTM) that enables the algorithm to use observations of an arbitrary number of other agents, instead of a small, fixed number of neighbors.
The proposed algorithm is shown to outperform a classical collision avoidance algorithm, another deep RL-based algorithm, and scales with the number of agents better (fewer collisions, shorter time to goal) than our previously published learning-based approach.
Analysis of the LSTM provides insights into how observations of nearby agents affect the hidden state and quantifies the performance impact of various agent ordering heuristics.
The learned policy generalizes to several applications beyond the training scenarios: formation control (arrangement into letters), demonstrations on a fleet of four multirotors and on a fully autonomous robotic vehicle capable of traveling at human walking speed among pedestrians.
\end{abstract}

\begin{keywords}
Collision Avoidance, Deep Reinforcement Learning, Motion Planning, Multiagent Systems, Decentralized Execution
\end{keywords}

\titlepgskip=-15pt

\maketitle

\section{Introduction} \label{sec:intro}

A fundamental challenge in autonomous vehicle operation is to safely negotiate interactions with other dynamic agents in the environment.
For example, it is important for self-driving cars to take other vehicles' motion into account, and for delivery robots to avoid colliding with pedestrians.
While there has been impressive progress in the past decade~\cite{kummerle2013navigation}, fully autonomous navigation remains challenging, particularly in uncertain, dynamic environments cohabited by other mobile agents.
The challenges arise because the other agents' intents and policies (i.e., goals and desired paths) are typically not known to the planning system, and, furthermore, explicit communication of such hidden quantities is often impractical due to physical limitations. These issues motivate the use of decentralized collision avoidance algorithms.

Existing work on decentralized collision avoidance can be classified into cooperative and non-cooperative methods.
Non-cooperative methods first predict the other agents' motion and then plan a collision-free path for the vehicle with respect to the other agents' predicted motion. 
However, this can lead to the \emph{freezing robot problem}~\cite{trautman_unfreezing_2010}, where the vehicle fails to find any feasible path because the other agents' predicted paths would occupy a large portion of the traversable space.
Cooperative methods address this issue by modeling interaction in the planner, such that the vehicle's action can influence the other agent's motion, thereby having all agents share the responsibility for avoiding collision.
Cooperative methods include reaction-based methods~\cite{snape_hybrid_2011,ferrer_social-aware_2013, berg_reciprocal_2011,alonso2013optimal} and trajectory-based methods~\cite{kretzschmar_socially_2016,trautman_robot_2013,kuderer_feature-based_2012}.

This work seeks to combine the best of both types of cooperative techniques -- the computational efficiency of reaction-based methods and the smooth motion of trajectory-based methods.
To this end, the work presents the collision avoidance with deep reinforcement learning (CADRL) algorithm, which tackles the aforementioned trade-off between computation time and smooth motion by using reinforcement learning (RL) to offload the expensive online computation to an offline learning procedure.
Specifically, a computationally efficient (i.e., real-time implementable) interaction rule is developed by learning a policy that implicitly encodes cooperative behaviors.

Learning the collision avoidance policy for CADRL presents several challenges.
A first key challenge is that the number of other agents in the environment can vary between timesteps or experiments, however the typical feedforward neural networks used in this domain require a fixed-dimension input.
Our prior work defines a maximum number of agents that the network can observe, and other approaches use raw sensor data as the input~\cite{long2018towards,tai2017virtual}.
This work instead uses an idea from Natural Language Processing~\cite{sutskever2014sequence,cho2014learning} to encode the varying size state of the world (e.g., positions of other agents) into a fixed-length vector, using long short-term memory (LSTM)~\cite{hochreiter1997long} cells at the network input.
This enables the algorithm to make decisions based on an arbitrary number of other agents in the robot's vicinity.

A second fundamental challenge is in finding a policy that makes realistic assumptions about other agents' belief states, policies, and intents.
This work learns a collision avoidance policy without assuming that the other agents follow any particular behavior model and without explicit assumptions on homogeneity~\citep{long2018towards} (e.g., agents of the same size and nominal speed) or specific motion models (e.g., constant velocity) over short timescales~\citep{chen_decentralized_2017,Chen17_IROS}.


The main contributions of this work are:
\begin{itemize}
    \item a new collision avoidance algorithm that greatly outperforms prior works as the number of agents in the environments is increased: a key factor in that improvement is to relax the assumptions on the other agents' behavior models during training and inference,
    \item a novel use of LSTM in that it encodes spatial representations, rather than temporal, to address the challenge that the number of neighboring agents could be large and could vary in time,
    \item simulation results that show significant improvement in solution quality compared with other recently published state-of-the-art methods (such as~\cite{berg_reciprocal_2011,Chen17_IROS,long2018towards}), and
    \item hardware experiments with aerial and ground robots to demonstrate that the proposed algorithm can be deployed in real time on robots with real sensors.
\end{itemize}
Open-source software based on this manuscript includes a pre-trained collision avoidance policy (as a ROS package) \texttt{cadrl\_ros}\footnote{\url{https://github.com/mit-acl/cadrl_ros}}, the GA3C-CADRL learning algorithm\footnote{\url{https://github.com/mit-acl/rl_collision_avoidance}}, and a simulation/training environment with several implemented policies, \texttt{gym\_collision\_avoidance}\footnote{\url{https://github.com/mit-acl/gym-collision-avoidance}}.
Videos of the experimental results are posted at \url{https://youtu.be/Bjx4ZEov0yE}.

This work is based on~\cite{chen_decentralized_2017,Chen17_IROS,Everett18_IROS} and extends them as follows: 
(i) expanded discussion and example of the limitations of the prior work,
(ii) further explanation of the proposed algorithm, including pseudo-code,
(iii) analysis on the effect of sequence ordering in LSTM, which addresses a primary gap in the prior work,
(iv) quantifying input gate activation to provide deeper intuition on why the proposed use of LSTM works,
(v) additional comparisons to model- and learning-based collision avoidance algorithms, 
(vi) ablation study of the proposed algorithm, and
(vii) experiments with formation control and on real multirotors to demonstrate generalizability of the learned policy.
\section{Background}\label{sec:background}
\subsection{Problem Formulation}\label{sec:prob_formulation}
The non-communicating, multiagent collision avoidance problem can be formulated as a sequential decision making problem~\citep{chen_decentralized_2017,Chen17_IROS}.
In an n-agent scenario ($\mathbb{N}_{\leq n}=\{1,2,\dots,n\}$), denote the joint world state, $\mathbf{s}^{jn}_t$, agent $i$'s state, $\mathbf{s}_{i,t}$, and agent $i$'s action, $\mathbf{u}_{i,t}$, $\forall i \in \mathbb{N}_{\leq n}$.
Each agent's state vector is composed of an observable and unobservable (hidden) portion, $\mathbf{s}_{i,t} = [\mathbf{s}^o_{i,t}, \, \mathbf{s}^h_{i,t}]$.
In the global frame, observable states are the agent's position, velocity, and radius, $\mathbf{s}^o = [p_x, \, p_y, \, v_x, \, v_y, \, r] \in \mathbb{R}^{5}$, and unobservable states are the goal position, preferred speed, and orientation\footnote{Other agents' positions and velocities are straightforward to estimate with a 2D Lidar, unlike human body \changed{heading angle}}, $\mathbf{s}^h = [p_{gx}, \, p_{gy}, \, v_{pref}, \, \psi] \in \mathbb{R}^{4}$.
The action is a speed and heading angle, $\mathbf{u}_t = [v_t, \, \psi_t] \in \mathbb{R}^{2}$.
The observable states of all $n-1$ other agents is denoted, $\tilde{\mathbf{S}}_{i,t}^o = \{\tilde{\mathbf{s}}^o_{j,t}: j\in \mathbb{N}_{\leq n} \setminus i\}$.
A policy, $\pi: \left( \mathbf{s}_{0:t}, \, \tilde{\mathbf{S}}_{0:t}^o \right) \mapsto \mathbf{u}_t$, is developed with the objective of minimizing expected time to goal $\mathbb{E}[t_g]$ while avoiding collision with other agents,
\begin{align}
\argmin_{\pi_i} \quad & \mathbb{E} \left[t_g | \mathbf{s}_i, \, \tilde{\mathbf{S}}^o_i,  \, \pi_i \right] \label{eqn:cost} \\ 
s.t. \quad & ||\mathbf{p}_{i,t} - \tilde{\mathbf{p}}_{j,t}||_2 \geq r_i + r_j \qquad \forall j\neq i, \forall t
		\label{eqn:con_collision} \\
	 \quad & \mathbf{p}_{i,t_g} = \mathbf{p}_{i,g} \label{eqn:con_reach_goal} \quad \forall i\\
	 \quad & \mathbf{p}_{i,t} = \mathbf{p}_{i,t-1} + \Delta t \cdot \pi_i( \mathbf{s}_{i,t-1}, \, \tilde{\mathbf{S}}^o_{i,t-1}) \forall i,
	 	\label{eqn:con_kinematics}
\end{align}
where \cref{eqn:con_collision} is the collision avoidance constraint, \cref{eqn:con_reach_goal} is the goal constraint, \cref{eqn:con_kinematics} is the agents' kinematics, 
and the expectation in \cref{eqn:cost} is with respect to the other agent's unobservable states (intents) and policies.

Although it is difficult to solve for the optimal solution of \cref{eqn:cost}-\cref{eqn:con_kinematics}, this problem formulation can be useful for understanding the limitations of the existing methods. In particular, it provides insights into the approximations/assumptions made by existing works.



\subsection{Related Work}

Most approaches to collision avoidance with dynamic obstacles employ model-predictive control (MPC) \citep{rawlings2000tutorial} in which a planner selects a minimum cost action sequence, $\mathbf{u}_{i,t:t+T}$, using a prediction of the future world state, $P(\mathbf{s}^{jn}_{t+1:t+T+1}|\mathbf{s}^{jn}_{0:t}, \mathbf{u}_{i,t:t+T})$, conditioned on the world state history, $\mathbf{s}^{jn}_{0:t}$.
While the first actions in the sequence are being implemented, the subsequent action sequence is updated by re-planning with the updated world state information (e.g., from new sensor measurements).
The prediction of future world states is either prescribed using domain knowledge (model-based approaches) or learned from examples/experiences (learning-based approaches).

\subsubsection{Model-based approaches}\label{sec:background:model_based_approaches}
Early approaches model the world as a static entity, $[v_x, \, v_y] = \mathbf{0}$, but replan quickly to try to capture the motion through updated $(p_x, p_y)$ measurements~\citep{fox_dynamic_1997}.
This leads to time-inefficient paths among dynamic obstacles, since the planner's world model does not anticipate future changes in the environment due to the obstacles' motion.

To improve the predictive model, \textit{reaction-based} methods use one-step interaction rules based on geometry or physics to ensure collision avoidance.
These methods~\citep{berg_reciprocal_2011,ferrer_social-aware_2013,alonso2013optimal} often specify a Markovian policy, $\pi( \mathbf{s}^{jn}_{0:t}) = \pi( \mathbf{s}^{jn}_t)$, that optimizes a one-step cost while satisfying collision avoidance constraints.
For instance, in velocity obstacle approaches~\citep{berg_reciprocal_2011,alonso2013optimal}, an agent chooses a collision-free velocity that is closest to its preferred velocity (i.e., directed toward its goal). 
Given this one-step nature, reaction-based methods do account for current obstacle motion, but do not anticipate the other agents' hidden intents -- they instead rely on a fast update rate to react quickly to the other agents' changes in motion. 
Although computationally efficient given these simplifications, reaction-based methods are myopic in time, which can sometimes lead to generating unnatural trajectories~\citep{trautman_robot_2013,chen_decentralized_2017}.

\textit{Trajectory-based} methods compute plans on a longer timescale to produce smoother paths but are often computationally expensive or require knowledge of unobservable states.
A subclass of non-cooperative approaches~\citep{phillips_sipp:_2011,aoude_probabilistically_2013} propagates the other agents' dynamics forward in time and then plans a collision-free path with respect to the other agents' predicted paths.
However, in crowded environments, the set of predicted paths could occupy a large portion of the space, which leads to the \emph{freezing robot problem}~\citep{trautman_unfreezing_2010}.
A key to resolving this issue is to account for interactions, such that each agent's motion can affect one another.
Thereby, a subclass of cooperative approaches~\citep{kretzschmar_socially_2016,trautman_robot_2013,kuderer_feature-based_2012} has been proposed, which solve \cref{eqn:cost}-\cref{eqn:con_kinematics} in two steps.
First, the other agents' hidden states (i.e., goals) are inferred from their observed trajectories, $\hat{\tilde{\mathbf{S}}}^h_t = f(\tilde{\mathbf{S}}^o_{0:t})$, where $f(\cdot)$ is a inference function. 
Second, a centralized path planning algorithm, $\pi( \mathbf{s}_{0:t}, \, \tilde{\mathbf{S}}^o_{0:t}) = \pi_{central}( \mathbf{s}_t, \, \tilde{\mathbf{S}}^o_t, \, \hat{\tilde{\mathbf{S}}}^h_t)$, is employed to find jointly feasible paths.
By planning/anticipating complete paths, trajectory-based methods are no longer myopic.
However, both the inference and the planning steps are computationally expensive, and need to be carried out online at each new observation (sensor update $\tilde{\mathbf{S}}^o_t$).

\subsubsection{Learning-based approaches}
Our recent works~\citep{chen_decentralized_2017,Chen17_IROS} proposed a third category that uses a reinforcement learning framework to solve \cref{eqn:cost}-\cref{eqn:con_kinematics}.
As in the reactive-based methods, we make a Markovian assumption: $\pi(\mathbf{s}^{jn}_{0:t})=\pi(\mathbf{s}^{jn}_t)$.
The expensive operation of modeling the complex interactions is learned in an offline training step, whereas the learned policy can be queried quickly online, combining the benefits of both reactive- and trajectory-based methods.
Our prior methods pre-compute a value function, $V(\mathbf{s}^{jn})$, that estimates the expected time to the goal from a given configuration, which can be used to select actions using a one-step lookahead procedure described in those works.
To avoid the lookahead procedure, this work directly optimizes a policy $\pi(\mathbf{s}^{jn})$ to select actions to minimize the expected time to the goal.
The differences from other learning-based approaches will become more clear after a brief overview of reinforcement learning.

\subsection{Reinforcement Learning} \label{sec:prob:RL}
RL~\citep{sutton_introduction_1998} is a class of machine learning methods for solving sequential decision making problems with unknown state-transition dynamics. Typically, a sequential decision making problem can be formulated as a Markov decision process (MDP), which is defined by a tuple $M=\langle S,A,P,R,\gamma\rangle$, where $S$ is the state space, $A$ is the action space, $P$ is the state-transition model, $R$ is the reward function, and $\gamma$ is a discount factor. By detailing each of these elements and relating to \cref{eqn:cost}-\cref{eqn:con_kinematics}, the following provides a RL formulation of the $n$-agent collision avoidance problem.

\paragraph*{State space} The joint world state, $\mathbf{s}^{jn}$, was defined in~\cref{sec:prob_formulation}.

\paragraph*{Action space} The choice of action space depends on the vehicle model.
A natural choice of action space for differential drive robots is a linear and angular speed (which can be converted into wheel speeds), that is, $\mathbf{u}=[s,\, \omega]$.
The action space is either discretized directly, or represented continuously by a function of discrete parameters.

\paragraph*{Reward function}  A sparse reward function is specified to award the agent for reaching its goal~\cref{eqn:con_reach_goal}, and penalize the agent for getting too close or colliding with other agents~\cref{eqn:con_collision},
\begin{align} \hspace*{-.1in}
  R(\mathbf{s}^{jn},\mathbf{u}) = 
  \begin{cases}
    1 & \text{if $\mathbf{p} = \mathbf{p}_g$} \label{eqn:reward} \\
    -0.1 + d_{min}/2 & \text{if $0 < d_{min} < 0.2$} \\
    -0.25 & \text{if $d_{min} < 0$} \\
    0 & \text{otherwise},
  \end{cases}
\end{align}
where $d_{min}$ is the distance to the closest other agent.
Optimizing the hyperparameters (e.g., -0.25) in $R_{col}$ is left for future work.
Note that we use discount $\gamma < 1$ to encourage efficiency instead of a step penalty.

\paragraph*{State transition model}
\indent A probabilistic state transition model, $P(\mathbf{s}^{jn}_{t+1}|\mathbf{s}^{jn}_t,\mathbf{u}_t)$, is determined by the agents' kinematics as defined in~\cref{eqn:con_kinematics}.
Since the other agents' actions also depend on their policies and hidden intents (e.g., goals), the system's state transition model is unknown.

\paragraph*{Value function} One method to find the optimal policy is to first find the optimal value function,
\begin{align}
V^*(\mathbf{s}_0^{jn}) &= \mathbb{E}\left[\sum_{t=0}^T \gamma^{t} \, R(\mathbf{s}^{jn}_t, \pi^*(\mathbf{s}^{jn}_t))\right],
\label{eqn:optimal_value}
\end{align}
where $\gamma\in[0,1)$ is a discount factor.
Many methods exist to estimate the value function in an offline training process~\citep{sutton_introduction_1998}.

\paragraph*{Deep Reinforcement Learning}
To estimate the high-dimensional, continuous value function (and/or associated policy), it is common to approximate with a deep neural network (DNN) parameterized by weights and biases, $\theta$, as in~\citep{mnih-dqn-2015}.
This work's notation drops the parameters except when possible, e.g., $V(\mathbf{s};\theta)=V(\mathbf{s})$.

\paragraph*{Decision-making Policy}
A value function of the current state can be implemented as a policy,
\begin{align}
& \pi^*(\mathbf{s}^{jn}_{t+1}) = \argmax_{\mathbf{u}} R(\mathbf{s}_{t}, \mathbf{u}) + \nonumber \\ 
& \qquad \quad \gamma^{\Delta t \cdot v_{pref}}\int_{\mathbf{s}_{t+1}^{jn}}P(\mathbf{s}^{jn}_{t}, \mathbf{s}^{jn}_{t+1}|\mathbf{u}) V^*(\mathbf{s}_{t+1}^{jn})d\mathbf{s}_{t+1}^{jn}. \label{eqn:optimal_policy}
\end{align}

Our previous works avoid the complexity in explicitly modeling $P(\mathbf{s}^{jn}_{t+1}|\mathbf{s}^{jn}_{t}, \mathbf{u})$ by assuming that other agents continue their current velocities, $\hat{\mathbf{V}}_t$, for a duration $\Delta t$, meaning the policy can be extracted from the value function,
\begin{align}
\hat{\mathbf{s}}^{jn}_{t+1,\mathbf{u}} &\leftarrow [f(\mathbf{s}_{t}, \Delta t \cdot \mathbf{u}), f(\tilde{\mathbf{S}}^{o}_{t}, \Delta t \cdot \hat{\mathbf{V}}_t)] \\
\pi_{CADRL}(\mathbf{s}^{jn}_{t}) &= \argmax_{\mathbf{u}} R_{col}(\mathbf{s}_{t}, \mathbf{u}) + \nonumber\\
&\qquad\qquad\qquad\gamma^{\Delta t \cdot v_{pref}} V(\hat{\mathbf{s}}^{jn}_{t+1,\mathbf{u}})\label{eqn:cadrl_policy},
\end{align}
under the simple kinematic model, $f$.

\begin{figure*}[t]
  \centering
  \begin{subfigure}{0.24\textwidth}
    \centering
    \includegraphics [trim=0 0 0 0, clip, width=\textwidth, angle = 0]{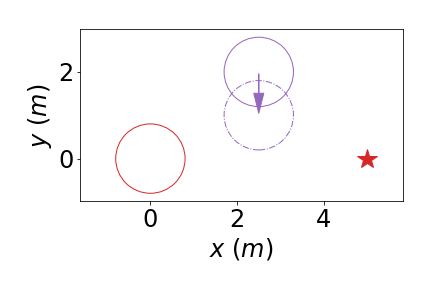}
    \caption{2-Agent Scenario\\\ }
    \label{fig:issue_cadrl_scenario} 
  \end{subfigure}
  \begin{subfigure}{0.24\textwidth}
    \centering
    \includegraphics [trim=0 0 0 0, clip, width=\textwidth, angle = 0]{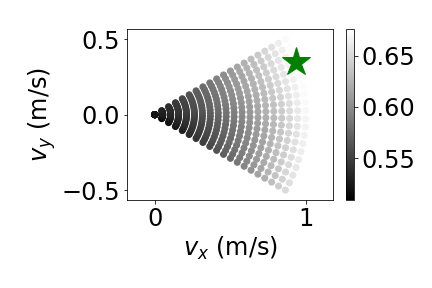}
    \caption{Time-to-Goal Estimate, $V(\hat{\mathbf{s}}^{jn}_{t+1,\mathbf{u}})$ from DNN}
    \label{fig:issue_cadrl_value} 
  \end{subfigure}
  \begin{subfigure}{0.24\textwidth}
    \centering
    \includegraphics [trim=0 0 0 0, clip, width=\textwidth, angle = 0]{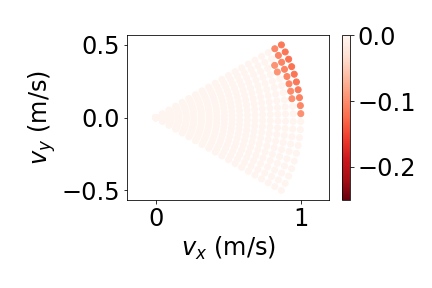}
    \caption{Collision Cost, $R_{col}(\mathbf{s}_{t}, \mathbf{u})$, using const. velocity model}
    \label{fig:issue_cadrl_collision} 
  \end{subfigure}
  \begin{subfigure}{0.24\textwidth}
    \centering
    \includegraphics [trim=0 0 0 0, clip, width=\textwidth, angle = 0]{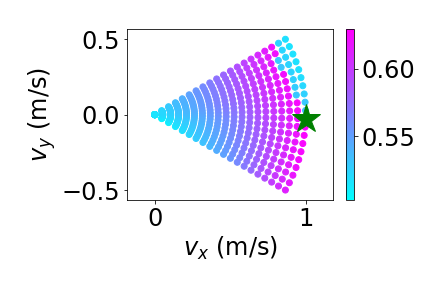}
    \caption{Objective in $\pi_{CADRL}(\mathbf{s}^{jn}_{t})$\\\ }
    \label{fig:issue_cadrl_policy} 
  \end{subfigure}
  \caption{
  \changed{
  Issue with checking collisions and state-value separately, as in~\cref{eqn:cadrl_policy}.
  In (a), the red agent's goal is at the star, and the purple agent's current velocity is in the $-y$-direction.
  In (b), the CADRL algorithm propagates the other agent forward at its current velocity (dashed purple circle), then queries the DNN for candidate future states.
  The best action (green star) is one which cuts above the purple agent, which was learned correctly by the CADRL V-Learning procedure.
  However, the constant velocity model of other agents is also used for collision checking, causing penalties of $R_{col}(\mathbf{s}_{t}, \mathbf{u})$, shown in (c).
  CADRL's policy combines these terms (d), instead choosing to go straight (green star), which is a poor choice that ignores that a cooperative purple agent likely would adjust its own velocity as well.
  This fundamental issue of checking collisions and state-values separately is addressed in this work by learning a policy directly.
  }
  }
  \label{fig:issue_cadrl}
\end{figure*}

However, the introduction of parameter $\Delta t$ leads to a difficult trade-off.
Due to the the approximation of the value function in a DNN, a sufficiently large $\Delta t$ is required such that each propagated $\hat{\mathbf{s}}_{t+1,\mathbf{u}}^{jn}$ is far enough apart, which ensures $V(\hat{\mathbf{s}}_{t+1,\mathbf{u}}^{jn})$ is not dominated by numerical noise in the network.
The implication of large $\Delta t$ is that agents are assumed to follow a constant velocity for a significant amount of time, which neglects the effects of cooperation/reactions to an agent's decisions.
As the number of agents in the environment increases, this constant velocity assumption is less likely to be valid.
Agents do not actually reach their propagated states because of the multiagent interactions.

\changed{
The impact of separately querying the value function and performing collision checking is illustrated in~\cref{fig:issue_cadrl}.
In (a), a red agent aims to reach its goal (star), and a purple agent is traveling at 1 m/s in the $-y$-direction.
Because CADRL's value function only encodes time-to-goal information, (b) depicts that the DNN appropriately recommends that the red agent should cut above the purple agent.
However, there is a second term in~\cref{eqn:cadrl_policy} to convert the value function into a policy.
This second term, the collision cost, $R_{col}(\mathbf{s}_{t}, \mathbf{u})$, shown in (c),  penalizes actions that move toward the other agent's predicted position (dashed circle).
This model-based collision checking procedure requires an assumption about other agents' behaviors, which is difficult to define ahead of time; the prior work assumed a constant-velocity model.
When the value and collision costs are combined to produce $\pi_{CADRL}(\mathbf{s}^{jn}_{t})$, the resulting objective-maximizing action is for the red agent to go straight, which will avoid a collision but be inefficient for both agents.
The challenge in defining a model for other agents' behaviors was a primary motivation for learning a value function; even with an accurate value function, this example demonstrates an additional cause of inefficient paths: an inaccurate model used in the collision checking procedure.
}

In addition to not capturing decision making behavior of other agents, our experiments suggest that $\Delta t$ is a crucial parameter to ensure convergence while training the DNNs in the previous algorithms.
If $\Delta t$ is set too small or large, the training does not converge.
A value of $\Delta t = 1$ sec was experimentally determined to enable convergence, though this number does not have much theoretical rationale.

In summary, the challenges of converting a value function to a policy, choosing the $\Delta t$ hyperparameter, and our observation that the learning stability suffered with more than 4 agents in the environment each motivate the use of a different RL framework.
\added{To address the concerns raised about $\Delta T$ propagation, this work proposes a new algorithm that does not project agents forward during policy evaluation, thus eliminating the need for tuning the $\Delta t$ hyperparameter.}

\paragraph*{Policy Learning}

Therefore, this work considers RL frameworks which generate a policy that an agent can execute directly, without any arbitrary assumptions about state transition dynamics.
A recent actor-critic algorithm called A3C~\citep{mnih2016asynchronous} uses a single DNN to approximate both the value (critic) and policy (actor) functions, and is trained with two loss terms
\begin{align}
& f_{v} = (R_t - V(\mathbf{s}^{jn}_t))^2 \label{eqn:a3c_value}, \\
& f_{\pi} = \log \pi(\textbf{u}_t|\textbf{s}_t^{jn}) (R_t - V(\textbf{s}_t^{jn})) + \beta \cdot H(\pi(\textbf{s}_t^{jn})), \label{eqn:a3c_policy}
\end{align}
where \cref{eqn:a3c_value} trains the network's value output to match the future discounted reward estimate, \added{$R_t = \sum_{i=0}^{k-1} \gamma^i r_{t+i} + \gamma^{k}V(\mathbf{s}_{t+k}^{jn})$}, over the next $k$ steps, just as in CADRL.
For the policy output in \cref{eqn:a3c_policy}, the first term penalizes actions which have high probability of occurring ($\log\pi$) that lead to a lower return than predicted by the value function $(R-V)$, and the second term encourages exploration by penalizing $\pi$'s entropy with tunable constant $\beta$.

In A3C, many threads of an agent interacting with an environment are simulated in parallel, and a policy is trained based on an intelligent fusion of all the agents' experiences.
The algorithm was shown to learn a policy that achieves super-human performance on many video games.
We specifically use GA3C~\citep{babaeizadeh2017ga3c}, a hybrid GPU/CPU implementation that efficiently queues training experiences and action predictions.
Our work builds on open-source GA3C implementations~\citep{babaeizadeh2017ga3c,omidshafiei2017crossmodal}.

Other choices for RL policy training algorithms (e.g., PPO~\citep{schulman2017proximal}, TD3~\citep{fujimoto2018addressing}) are architecturally similar to A3C.
Thus, the challenges mentioned above (varying number of agents, assumptions about other agents' behaviors) would map to future work that considers employing other RL algorithms or techniques~\citep{hessel2018rainbow} in this domain.

\subsection{Related Works using Learning}
There are several concurrent and subsequent works which use learning to solve the collision avoidance problem, categorized as non-RL, RL, and agent-level RL approaches.

Non-RL-based approaches to the collision avoidance problem include imitation learning, inverse RL, and supervised learning of prediction models.
Imitation learning approaches~\citep{tai2017virtual} learn a policy that mimics what a human pedestrian or human teleoperator~\citep{bojarski2016end} would do in the same state but require data from an expert.
Inverse RL methods learn to estimate pedestrians' cost functions, then use the cost function to inform robot plans~\citep{kim_socially_2015,kretzschmar_socially_2016}, but require real pedestrian trajectory data.
Other approaches learn to predict pedestrian paths, which improves the world model used by the planner~\citep{pfeiffer2016predicting}, but decoupling the prediction and planning steps could lead to the freezing robot problem (\cref{sec:background:model_based_approaches}).
A key advantage of RL over these methods is the ability to explore the state space through self-play, in which experiences generated in a low-fidelity simulation environment can reduce the need for expensive, real-world data collection efforts.

Within RL-based approaches, a key difference arises in the state representation: sensor-level and agent-level.
Sensor-level approaches learn to select actions directly from raw sensor readings (either 2D laserscans~\citep{long2018towards} or images~\citep{tai2017virtual}) with end-to-end training.
This leads to a large state space ($\mathbbm{R}^{w \times h \times c}$ for a camera with resolution ${w \times h}$ and $c$ channels, e.g., $480 \times 360 \times 3=5184000$), which makes training challenging.
CNNs are often used to extract low-dimensional features from this giant state space, but training such a feature extractor in simulation requires an accurate sensor simulation model.
The sensor-level approach has the advantage that both static and dynamic obstacles (including walls) can be fed into the network with a single framework.
In contrast, this work uses interpretable clustering, tracking, and multi-sensor fusion algorithms to extract an agent-level state representation from raw sensor readings.
Advantages include a much smaller state space ($\mathbb{R}^{9+5(n-1)}$) enabling faster learning convergence; a sensor-agnostic collision avoidance policy, enabling sensor upgrades without re-training; and increased introspection into decision making, so that decisions can be traced back to the sensing, clustering, tracking, or planning modules.

Within agent-level RL, a key challenge is that of representing a variable number of nearby agents in the environment at any timestep.
Typical feedforward networks used to represent the complex decision making policy for collision avoidance require a pre-determined input size.
The sensor-level methods do maintain a fixed size input (sensor resolution), but have the limitations mentioned above.
Instead, our first work trained a 2-agent value network, and proposed a mini-max rule to scale up to $n$ agents~\citep{chen_decentralized_2017}.
To account for multiagent interactions (instead of only pairwise), our next work defines a maximum number of agents that the network can handle, and pads the observation space if there are actually fewer agents in the environment~\citep{Chen17_IROS}.
However, this maximum number of agents is limited by the increased number of network parameters (and therefore training time) as more agents' states are added.
This work uses a recurrent network to convert a sequence of agent states at a particular timestep into a fixed-size representation of the world state; that representation is fed into the input of a standard feedforward network.
\added{Beyond the scope of collision avoidance, recent work~\cite{Baker2020Emergent} introduced attention mechanisms, another tool popularized in NLP, as another method for embedding the variable number of other agents' states.}

There are also differences in the reward functions used in RL-based collision avoidance approaches.
Generally, the non-zero feedback provided at each timestep by a dense reward function (e.g., \citep{long2018towards}) makes learning easier, but reward shaping quickly becomes a difficult problem in itself.
For example, balancing multiple objectives (proximity to goal, proximity to others) can introduce unexpected and undesired local minima in the reward function.
On the other hand, sparse rewards are easy to specify but require a careful initialization/exploration procedure to ensure agents will receive \textit{some} environment feedback to inform learning updates.
This work mainly uses sparse reward (arrival at goal, collision) with smooth reward function decay in near-collision states to encourage a minimum separation distance between agents.
Additional terms in the reward function are shown to reliably induce higher-level preferences (social norms) in our previous work~\citep{Chen17_IROS}.

While learning-based methods have many potential advantages over model-based approaches, learning-based approaches typically lack the guarantees (e.g., avoiding deadlock, zero collisions) desired for safety-critical applications.
A key challenge in establishing guarantees in multiagent collision avoidance is what to assume about the world (e.g., policies and dynamics of other agents).
Unrealistic or overly conservative assumptions about the world invalidate the guarantees or unnecessarily degrade the algorithm's performance: striking this balance may be possible in some domains but is particularly challenging in pedestrian-rich environments.
A survey of the active research area of Safe RL is found in~\citep{garcia2015comprehensive}.

\section{Approach} \label{sec:approach}

\subsection{GA3C-CADRL}
Recall the RL training process seeks to find the optimal policy, $\pi: \left( \mathbf{s}_t, \, \tilde{\mathbf{S}}_t^o \right) \mapsto \mathbf{u}_t$, which maps from an agent's observation of the environment to a probability distribution across actions and executes the action with highest probability.
We use a local coordinate frame (rotation-invariant) as in~\citep{chen_decentralized_2017,Chen17_IROS} and separate the state of the world in two pieces: information about the agent itself, and everything else in the world.
Information about the agent can be represented in a small, fixed number of variables.
The world, on the other hand, can be full of any number of other objects or even completely empty.
Specifically, there is one $\mathbf{s}$ vector about the agent itself and one $\tilde{\mathbf{s}}^o$ vector per other agent in the vicinity:
\begin{align}
	\mathbf{s} & = [d_g, \; v_{pref}, \; \psi, \; r] \label{eqn:agent_state} \\  
	\tilde{\mathbf{s}}^o & = [\tilde{p}_x, \; \tilde{p}_y, \; \tilde{v}_x, \; \tilde{v}_y,
		 \; \tilde{r}, \; \tilde{d}_a, \; \tilde{r}+r] \;  \label{eqn:other_state}, 
\end{align}
where $d_g=||\mathbf{p}_g - \mathbf{p}||_2$ is the agent's distance to goal, and $\tilde{d}_a=||\mathbf{p} - \tilde{\mathbf{p}}||_2$ is the distance to the other agent.

The agent's action space is composed of a speed and change in heading angle.
It is discretized into 11 actions: with a speed of $v_{pref}$ there are 6 headings evenly spaced between $\pm\pi/6$, and for speeds of $\frac{1}{2}v_{pref}$ and 0 the heading choices are $\left[-\pi/6,\,0,\,\pi/6\right]$. 
These actions are chosen to mimic real turning constraints of robotic vehicles.

This multiagent RL problem formulation is solved with GA3C in a process we call GA3C-CADRL (GPU/CPU Asynchronous Advantage Actor-Critic for Collision Avoidance with Deep RL).
Since experience generation is one of the time-intensive parts of training, this work extends GA3C to learn from multiple agents' experiences each episode.
Training batches are filled with a mix of agents' experiences ($\{\mathbf{s}^{jn}_t,\,\mathbf{u}_t,\,r_t\}$ tuples) to encourage policy gradients that improve the joint expected reward of all agents.
Our multiagent implementation of GA3C accounts for agents reaching their goals at different times and ignores experiences of agents running other policies (e.g., non-cooperative agents).

\subsection{Handling a Variable Number of Agents}\label{sec:approach:variable_num_agents}

\begin{figure}[t]
	\centering
	\includegraphics [trim=0 50 0 50, clip, width=0.5 \textwidth, page=2]{figures/ijrr/network_architecture_v5.pdf}
	\caption{LSTM unrolled to show each input. At each decision step, the agent feeds one observable state vector, $\tilde{\mathbf{s}}^o_i$, for each nearby agent, into a LSTM cell sequentially. LSTM cells store the pertinent information in the hidden states, $\mathbf{h}_i$. The final hidden state, $\mathbf{h}_n$, encodes the entire state of the other agents in a fixed-length vector, and is then fed to the feedforward portion of the network. The order of agents is sorted by decreasing distance to the ego agent, so that the closest agent has the most recent effect on $\mathbf{h}_n$.}
	\label{fig:lstm}
	\includegraphics [trim=30 40 30 40, clip, width=0.5 \textwidth, page=1]{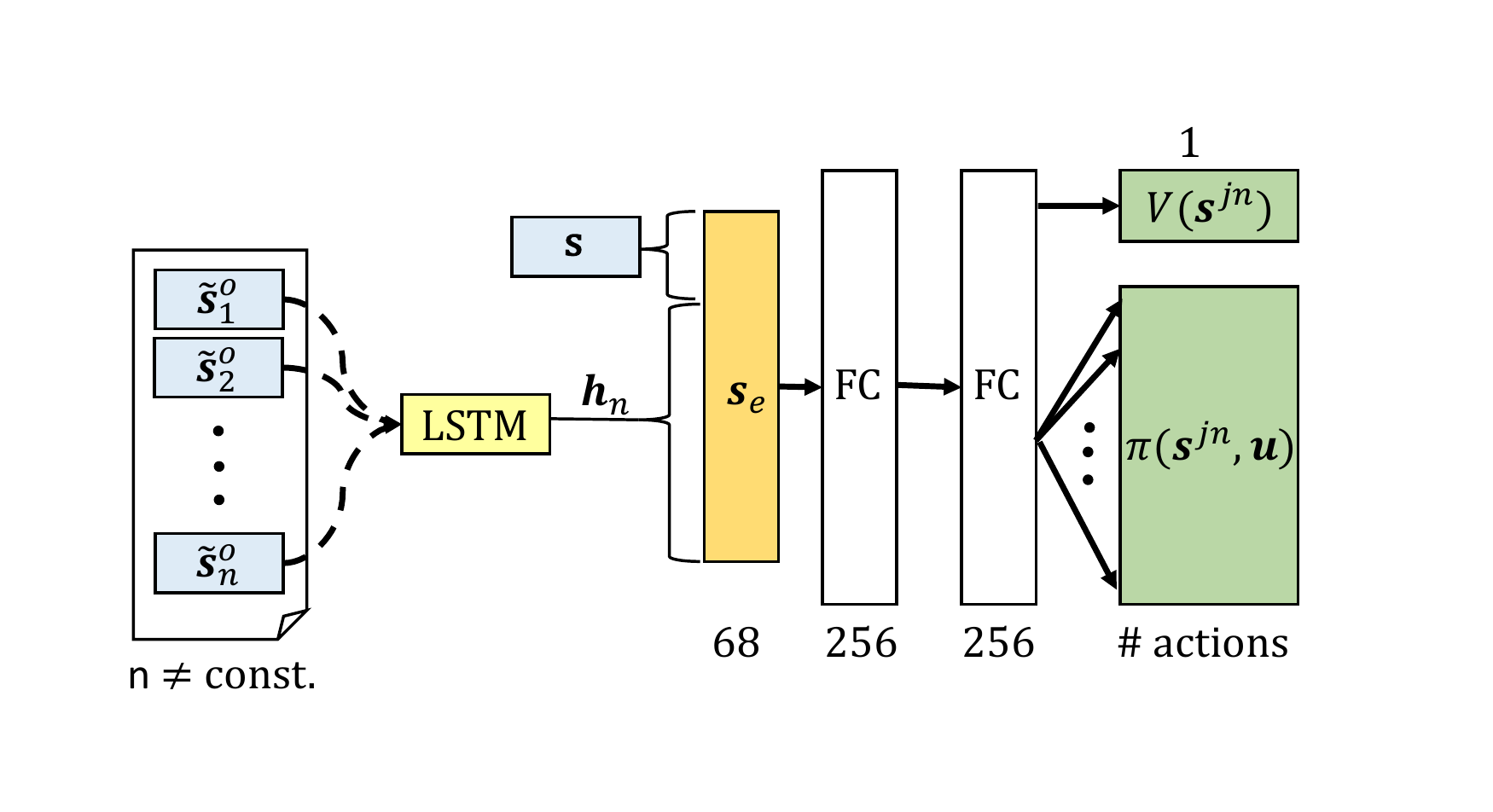}
	\caption{Network Architecture. Observable states of nearby agents, $\tilde{\mathbf{s}}^o_i$, are fed sequentially into the LSTM, as unrolled in~\cref{fig:lstm}. The final hidden state is concatenated with the agent's own state, $\mathbf{s}$, to form the vector, $\mathbf{s}_e$. For any number of agents, $\mathbf{s}_e$ contains the agent's knowledge of its own state and the state of the environment. The encoded state is fed into two fully-connected layers (FC). The outputs are a scalar value function (top, right) and policy represented as a discrete probability distribution over actions (bottom, right).}
	\label{fig:nn_arch}
\end{figure}

Recall that one key limitation of many learning-based collision avoidance methods is that the feedforward NNs typically used require a fixed-size input.
Convolutional and max-pooling layers are useful for feature extraction and can modify the input size but still convert a fixed-size input into a fixed-size output.
Recurrent NNs, where the output is produced from a combination of a stored cell state and an input, accept an arbitrary-length sequence to produce a fixed-size output.
Long short-term memory (LSTM)~\citep{hochreiter1997long} is recurrent architecture with advantageous properties for training\footnote{In practice, TensorFlow's LSTM implementation requires a known maximum sequence length, but this can be set to something bigger than the number of agents agents ever expected (e.g., 20)}.

Although LSTMs are often applied to time sequence data (e.g., pedestrian motion prediction~\citep{alahi2016social}), this paper leverages their ability to encode a sequence of information that is not time-dependent (see~\citep{Olah15} for a thorough explanation of LSTM calculations).
LSTM is parameterized by its weights, $\{W_i, W_f, W_o\}$, and biases, $\{b_i, b_f, b_o\}$, where $\{i, f, o\}$ correspond to the input, forget, and output gates.
The variable number of $\tilde{\mathbf{s}}^o_i$ vectors is a sequence of inputs that encompass everything the agent knows about the rest of the world.
As depicted in~\cref{fig:lstm}, each LSTM \textit{cell} has three inputs: the state of agent $j$ at time $t$, the previous hidden state, and the previous cell state, which are denoted $\tilde{\mathbf{s}}^o_{j,t}$, $\mathbf{h}_j$, $C_j$, respectively.
Thus, at each decision step, the agent feeds each $\tilde{\mathbf{s}}^o_i$ (observation of $i^{th}$ other agent's state) into a LSTM cell sequentially.
That is, the LSTM initially has empty states ($\mathbf{h}_0, \mathbf{C}_0$ set to zeros) and uses $\{\tilde{\mathbf{s}}^o_1, \mathbf{h}_0, C_0\}$ to generate $\{\mathbf{h}_1, C_1\}$, then feeds $\{\tilde{\mathbf{s}}^o_2, \mathbf{h}_1, C_1\}$ to produce $\{\mathbf{h}_2, C_2\}$, and so on.
As agents' states are fed in, the LSTM ``remembers'' the pertinent information in its hidden/cell states, and ``forgets'' the less important parts of the input (where the notion of memory is parameterized by the trainable LSTM weights/biases).
After inputting the final agent's state, we can interpret the LSTM's final hidden state, $\mathbf{h}_n$ as a fixed-length, encoded state of the world, for that decision step.
The LSTM contains $n$ cells, so the entire module receives inputs $\{\tilde{\mathbf{S}}^o_t, \mathbf{h}_{t-1}, \mathbf{C}_{t-1}\}$ and produces outputs $\{\mathbf{h}_n, \mathbf{C}_n\}$, and $\mathbf{h}_n$ is passed to the next network layer for decision making.

Given a sufficiently large hidden state vector, there is enough space to encode a large number of agents' states without the LSTM having to forget anything relevant.
In the case of a large number of agent states, to mitigate the impact of the agent forgetting the early states, the states are fed in reverse order of distance to the agent, meaning the closest agents (fed last) should have the biggest effect on the final hidden state, $\mathbf{h}_n$.
Because the list of agents needs to be ordered in some manner, reverse distance is one possible ordering heuristic -- we empirically compare to other possibilities in~\cref{sec:results:lstm_analysis}.

Another interpretation of the LSTM objective is that it must learn to combine an observation of a new agent with a representation of other agents (as opposed to the architectural objective of producing a fixed-length encoding of a varying size input).
This interpretation provides intuition on how an LSTM trained in 4-agent scenarios can generalize reasonably well to cases with 10 agents.

The addition of LSTM to a standard actor-critic network is visualized in~\cref{fig:nn_arch}, where the box labeled $\mathbf{s}$ is the agent's own state, and the group of boxes is the $n$ other agents' observable states, $\tilde{\mathbf{s}}^o_i$.
After passing the $n$ other agents' observable states into the LSTM, the agent's own state is concatenated with $\mathbf{h}_n$ to produce the encoded representation of the joint world state, $\mathbf{s}_e$.
Then, $\mathbf{s}_e$ is passed to a typical feedforward DNN with 2 fully-connected layers (256 hidden units each with ReLU activation).

The network produces two output types: a scalar state value (critic) and a policy composed of a probability for each action in the discrete action space (actor).
During training, the policy \textit{and} value are used for~\cref{eqn:a3c_value,eqn:a3c_policy}; during execution, only the policy is used.
During the training process, the LSTM's weights are updated to learn how to represent the variable number of other agents in a fixed-length $\mathbf{h}$ vector.
The whole network is trained end-to-end with backpropagation.

\subsection{Training the Policy}
The original CADRL and SA-CADRL (Socially Aware CADRL) algorithms used several clever tricks to enable convergence when training the networks.
Specifically, forward propagation of other agent states for $\Delta t$ seconds was a critical component that required tuning, but does not represent agents' true behaviors.
Other details include separating experiences into successful/unsuccessful sets to focus the training on cases where the agent could improve.
The new GA3C-CADRL formulation is more general, and does not require such assumptions or modifications.

The training algorithm is described in~\cref{alg:GA3C-CADRL-training}.
In this work, to train the model, the network weights are first initialized in a supervised learning phase, which converges in less than five minutes.
The initial training is done on a large, publicly released set of state-action-value tuples, $\{ \mathbf{s}^{jn}_t, \mathbf{u}_t, V(\mathbf{s}^{jn}_t; \phi_{CADRL}) \}$, from an existing CADRL solution.
The network loss combines square-error loss on the value output and softmax cross-entropy loss between the policy output and the one-hot encoding of the closest discrete action to the one in $D$, described in Lines \ref{alg:ga3c_cadrl_train:initialization_for_loop}-\ref{alg:ga3c_cadrl_train:sl_train} of~\cref{alg:GA3C-CADRL-training}.

The initialization step is necessary to enable any possibility of later generating useful RL experiences (non-initialized agents wander randomly and probabilistically almost never obtain positive reward).
Agents running the initialized GA3C-CADRL policy reach their goals reliably when there are no interactions with other agents.
However, the policy after this supervised learning process still performs poorly in collision avoidance.
This observation contrasts with CADRL, in which the initialization step was sufficient to learn a policy that performs comparably to existing reaction-based methods, due to relatively-low dimension value function combined with manual propagation of states.
Key reasons behind this contrast are the reduced structure in the GA3C-CADRL formulation (no forward propagation), and that the algorithm is now learning both a policy and value function (as opposed to just a value function), since the policy has an order of magnitude higher dimensionality than a scalar value function.

To improve the solution with RL, parallel simulation environments produce training experiences, described in {Lines \ref{alg:ga3c_cadrl_train:foreach_env}-\ref{alg:ga3c_cadrl_train:add_to_exp_queue}} of \cref{alg:GA3C-CADRL-training}.
Each episode consists of 2-10 agents, with random start and goal positions, running a random assortment of policies (Non-Cooperative, Zero Velocity, or the learned GA3C-CADRL policy at that iteration) (Line \ref{alg:ga3c_cadrl_train:random_test_case}).
Agent parameters vary between $r~\in~[0.2,\, 0.8]$m and $v_{pref} \in [0.5, \, 2.0]$m/s, chosen to be near pedestrian values.
Agents sense the environment and transform measurements to their ego frame to produce the observation vector (Lines \ref{alg:ga3c_cadrl_train:sensor_update}, \ref{alg:ga3c_cadrl_train:transform}).
Each agent sends its observation vector to the policy queue and receives an action sampled from the current iteration of the GA3C-CADRL policy (Line \ref{alg:ga3c_cadrl_train:query_all_actions}).
Agents that are not running the GA3C-CADRL policy use their own policy to overwrite $\mathbf{u}_{t,j}$ (Line \ref{alg:ga3c_cadrl_train:non_ga3c_agent_policy}).
Then, all agents that have not reached a terminal condition (collision, at goal, timed out), simultaneously move according to $\mathbf{u}_{t,j}$ (Line \ref{alg:ga3c_cadrl_train:move}).
After all agents have moved, the environment evaluates $R(\mathbf{s}^{jn}, \mathbf{u})$ for each agent, and experiences from GA3C-CADRL agents are sent to the training queue (Lines \ref{alg:ga3c_cadrl_train:rewards},\ref{alg:ga3c_cadrl_train:add_to_exp_queue}).

In another thread, experiences are popped from the queue to produce training batches (Line \ref{alg:ga3c_cadrl_train:grab_batch_from_queue}).
These experience batches are used to train a single GA3C-CADRL policy (Line \ref{alg:ga3c_cadrl_train:train_ga3c}) as in~\citep{babaeizadeh2017ga3c}.

\begin{algorithm}[t]
 \caption{GA3C-CADRL Training}
 \begin{algorithmic}[1]
 \renewcommand{\algorithmicrequire}{\textbf{Input:}}
 \renewcommand{\algorithmicensure}{\textbf{Output:}}
 \REQUIRE trajectory training set, $D$
 \ENSURE  policy network $\pi(\cdot;\theta)$
 \\ \texttt{// Initialization}
    \FOR{$N_{epochs}$} \label{alg:ga3c_cadrl_train:initialization_for_loop}
        \STATE $\{\mathbf{s}^o_{t},\tilde{\mathbf{S}}^o_{t}, \textbf{u}_t, V_t\} \leftarrow \textrm{grabBatch}(D)$  \\
        \STATE $\bar{\textbf{u}}_t \leftarrow \textrm{closestOneHot}(\textbf{u}_t)$\\
		\STATE $\mathcal{L}_V = (V_t - V(\mathbf{s}^o_{t},\tilde{\mathbf{S}}^o_{t}; \phi)^2$\\
		\STATE $\mathcal{L}_{\pi} = \textrm{softmaxCELogits}(\bar{\mathbf{u}}_t, \mathbf{s}^o_{t},\tilde{\mathbf{S}}^o_{t}, \theta)$\\
		\STATE $\pi(\cdot;\theta), V(\cdot;\phi)$ $\leftarrow$ trainNNs($\mathcal{L}_\pi, \mathcal{L}_V, \theta, \phi$) \\ \label{alg:ga3c_cadrl_train:sl_train}
    \ENDFOR
    \\\texttt{// Parallel Environment Threads} \\
	\FORALL{env} \label{alg:ga3c_cadrl_train:foreach_env}
		\STATE $\mathbf{S}_0 \leftarrow$ randomTestCase() \\ \label{alg:ga3c_cadrl_train:random_test_case}
		\WHILE{some agent not done}
			\FORALL{agent, $j$}
				\STATE $\mathbf{s}^o_g, \tilde{\mathbf{S}}^o_g \leftarrow$ sensorUpdate()\\ \label{alg:ga3c_cadrl_train:sensor_update}
				\STATE $\mathbf{s}^o, \tilde{\mathbf{S}}^o \leftarrow$ transform($\mathbf{s}^o_g, \tilde{\mathbf{S}}^o_g$)\\ \label{alg:ga3c_cadrl_train:transform}
			\ENDFOR
			\STATE $\{\mathbf{u}_{t,j}\} \sim \pi(\mathbf{s}^o_{t,j}, \tilde{\mathbf{S}}^o_{t,j}; \theta) \forall j$ \\ \label{alg:ga3c_cadrl_train:query_all_actions}
			\FORALL{not done agent, $j$}
				\IF{agent not running GA3C-CADRL} \label{alg:ga3c_cadrl_train:non_ga3c_agents}
					\STATE $\mathbf{u}_{t,j} \leftarrow \textrm{policy}(\mathbf{s}^o_{t,j}, \tilde{\mathbf{S}}^o_{t,j})$ \\ \label{alg:ga3c_cadrl_train:non_ga3c_agent_policy}
				\ENDIF
				\STATE $\mathbf{s}_{j,t+1},\tilde{\mathbf{S}}_{j,t+1}, r_{j,t} \leftarrow$ moveAgent($\mathbf{u}_{j,t}$) \\ \label{alg:ga3c_cadrl_train:move}
			\ENDFOR
			\FORALL{not done GA3C-CADRL agent, $j$}
				\STATE $r_{t,j} \leftarrow \textrm{checkRewards}(\mathbf{S}_{t+1},\mathbf{u}_{t,j})$ \\ \label{alg:ga3c_cadrl_train:rewards}
				\STATE $\textrm{addToExperienceQueue}(\mathbf{s}^o_{t,j}, \tilde{\mathbf{S}}^o_{t,j}, \mathbf{u}_{t,j}, r_{t,j})$ \\ \label{alg:ga3c_cadrl_train:add_to_exp_queue}
			\ENDFOR
		\ENDWHILE
	\ENDFOR
	\\\texttt{// Training Thread}
	\FOR{$N_{episodes}$}
		\STATE $\{\mathbf{s}^o_{t+1},\tilde{\mathbf{S}}^o_{t+1}, \mathbf{u}_t, r_t\} \leftarrow$ grabBatchFromQueue() \\ \label{alg:ga3c_cadrl_train:grab_batch_from_queue}
		\STATE $\theta, \phi \leftarrow \textrm{trainGA3C}(\theta, \phi, \{\mathbf{s}^o_{t+1},\tilde{\mathbf{S}}^o_{t+1}, \mathbf{u}_t, r_t\})$\\ \label{alg:ga3c_cadrl_train:train_ga3c}
	\ENDFOR
	\RETURN $\pi$
 \RETURN $P$ 
 \end{algorithmic}\label{alg:GA3C-CADRL-training}
\end{algorithm}

An important benefit of the new framework is that the policy can be trained on scenarios involving any number of agents, whereas the maximum number of agents had to be defined ahead of time with CADRL/SA-CADRL\footnote{Experiments suggest this number should be below about 6 for convergence}.
This work begins the RL phase with 2-4 agents in the environment, so that the policy learns the idea of collision avoidance in reasonably simple domains.
Upon convergence, a second RL phase begins with 2-10 agents in the environment.

\subsection{Policy Inference}

Inference of the trained policy for a single timestep is described in~\cref{alg:GA3C-CADRL-execution}.
As in training, GA3C-CADRL agents sense the environment, transfer to the ego frame, and select an action according to the policy (Lines \ref{alg:ga3c_cadrl_inf:sense}-\ref{alg:ga3c_cadrl_inf:query}).
Like many RL algorithms, actions are sampled from the stochastic policy during training (exploration), but the action with highest probability mass is selected during inference (exploitation).
A necessary addition for hardware is a low-level controller to track the desired speed and heading angle (Line \ref{alg:ga3c_cadrl_inf:control}).
Note that the value function is not used during inference; it is only learned to stabilize estimates of the policy gradients during training.

\begin{algorithm}[H]
 \caption{GA3C-CADRL Execution}
 \begin{algorithmic}[1]
 \renewcommand{\algorithmicrequire}{\textbf{Input:}}
 \renewcommand{\algorithmicensure}{\textbf{Output:}}
 \REQUIRE goal position, $(g_x, g_y)$
 \ENSURE  next motor commands, $\mathbf{u}$
 \STATE $\mathbf{s}^o_g, \tilde{\mathbf{S}}^o_g \leftarrow$ sensorUpdate()\\ \label{alg:ga3c_cadrl_inf:sense}
 \STATE $\mathbf{s}^o, \tilde{\mathbf{S}}^o \leftarrow$ transform($\mathbf{s}^o_g, \tilde{\mathbf{S}}^o_g$)\\ \label{alg:ga3c_cadrl_inf:transform}
 \STATE $s_{des}, \theta_{des} \leftarrow \pi(\mathbf{s}^o, \tilde{\mathbf{S}}^o)$\\ \label{alg:ga3c_cadrl_inf:query}
 \STATE $\mathbf{u}$ $\leftarrow$ control($s_{des}, \theta_{des}$) \\ \label{alg:ga3c_cadrl_inf:control}
 \RETURN $\mathbf{u}$
\end{algorithmic}\label{alg:GA3C-CADRL-execution}
\end{algorithm}



\begin{figure*}[t]
	\centering
	\includegraphics [trim=0 150 0 50, clip, width=\linewidth, page=1]{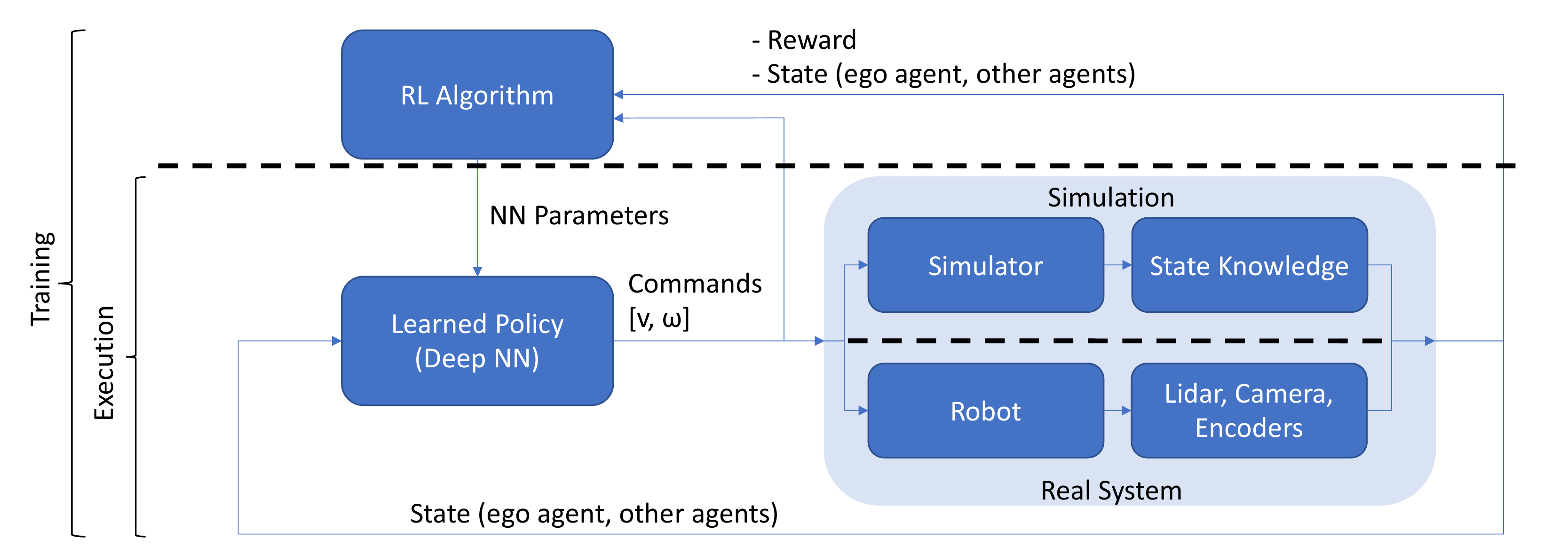}
	\caption{\added{System Architecture. During training, the policy receives state measurements to compute robot commands, and the environment returns next states and rewards. Collections of (state, action, reward) tuples enable an RL algorithm to update the parameters of the learned policy. During execution, only the blocks below the upper dashed line run (NN parameters are fixed). The key difference between executing in simulation vs. the real robot is that the robot's onboard sensors (lidar, cameras, encoders) estimate the state of the environment (e.g., agents' positions, velocities).}}
	\label{fig:system_arch}
\end{figure*}

\added{
The architecture of the training and inference steps for the simulated and real robot system are shown in~\cref{fig:system_arch}.
}

\section{Results}\label{sec:results}

\begin{figure*}[t]
	\centering
	\begin{subfigure}{0.99\textwidth}
		\centering
		\includegraphics [trim=20 0 50 60, clip, width=0.24 \textwidth, angle = 0]{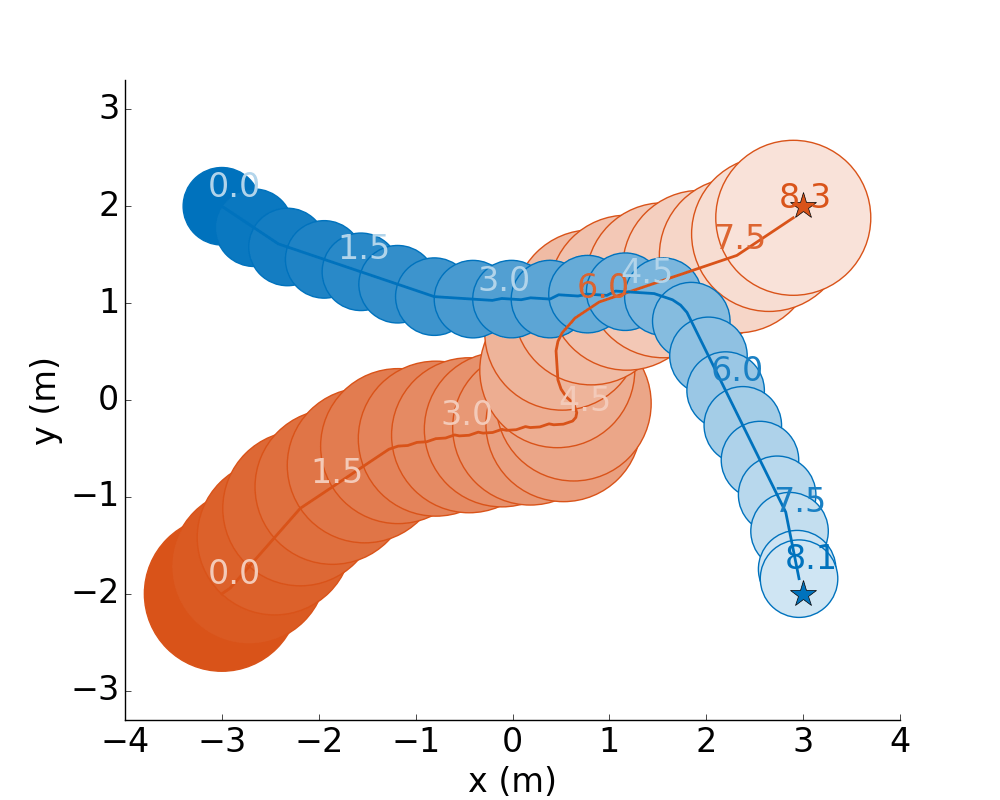}
		\includegraphics [trim=20 0 50 60, clip, width=0.24 \textwidth, angle = 0]{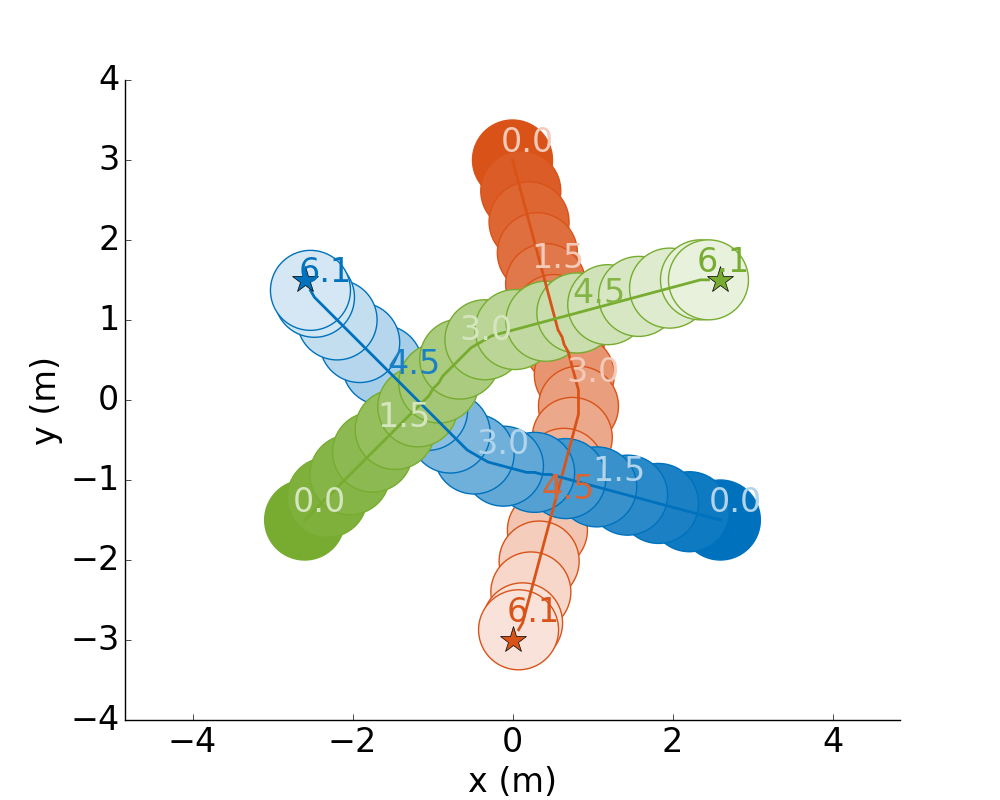}
		\includegraphics [trim=20 0 50 60, clip, width=0.24 \textwidth, angle = 0]{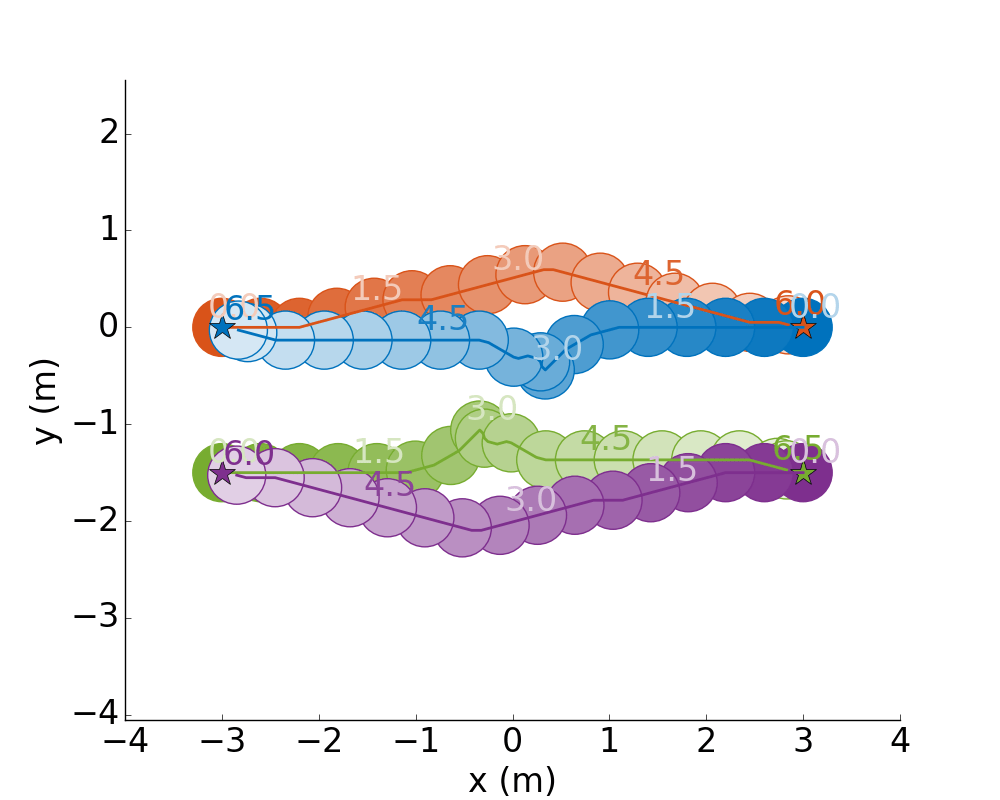}
		\includegraphics [trim=20 0 50 60, clip, width=0.24 \textwidth, angle = 0]{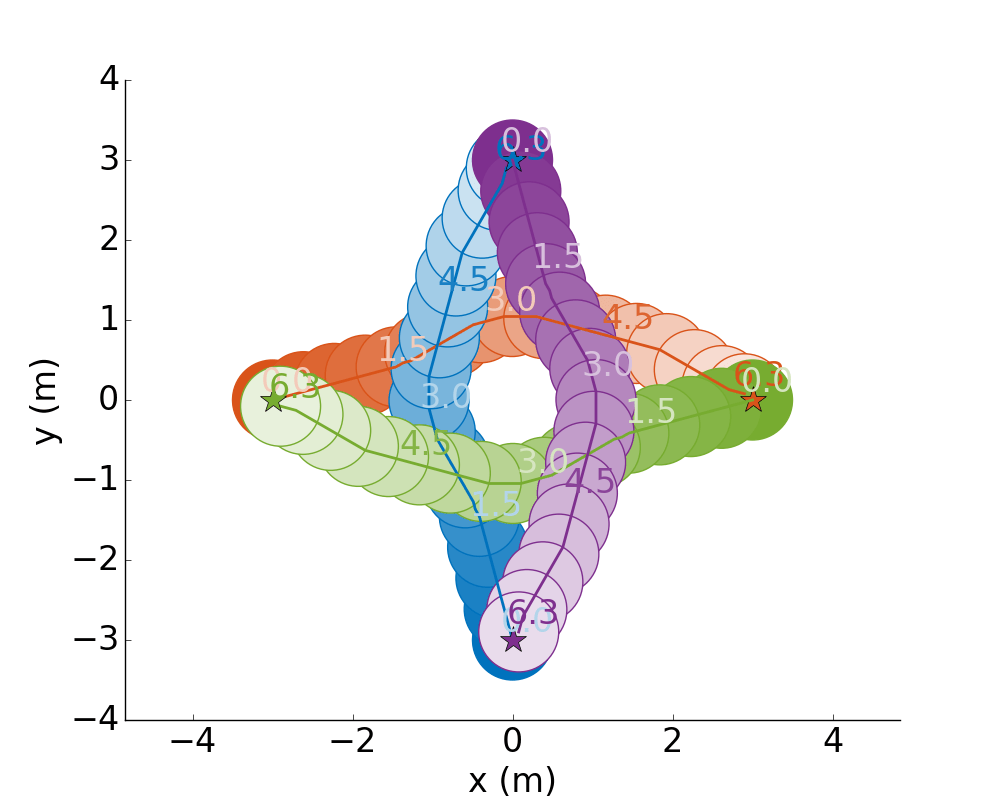}
		\caption{GA3C-CADRL trajectories with $n\in[2,3,4]$ agents}
		\label{fig:ga3c_cadrl_traj_2_agent} 
	\end{subfigure}
	\begin{subfigure}{0.99\textwidth}
		\centering
		\includegraphics [trim=20 0 50 60, clip, width=0.24 \textwidth, angle = 0]{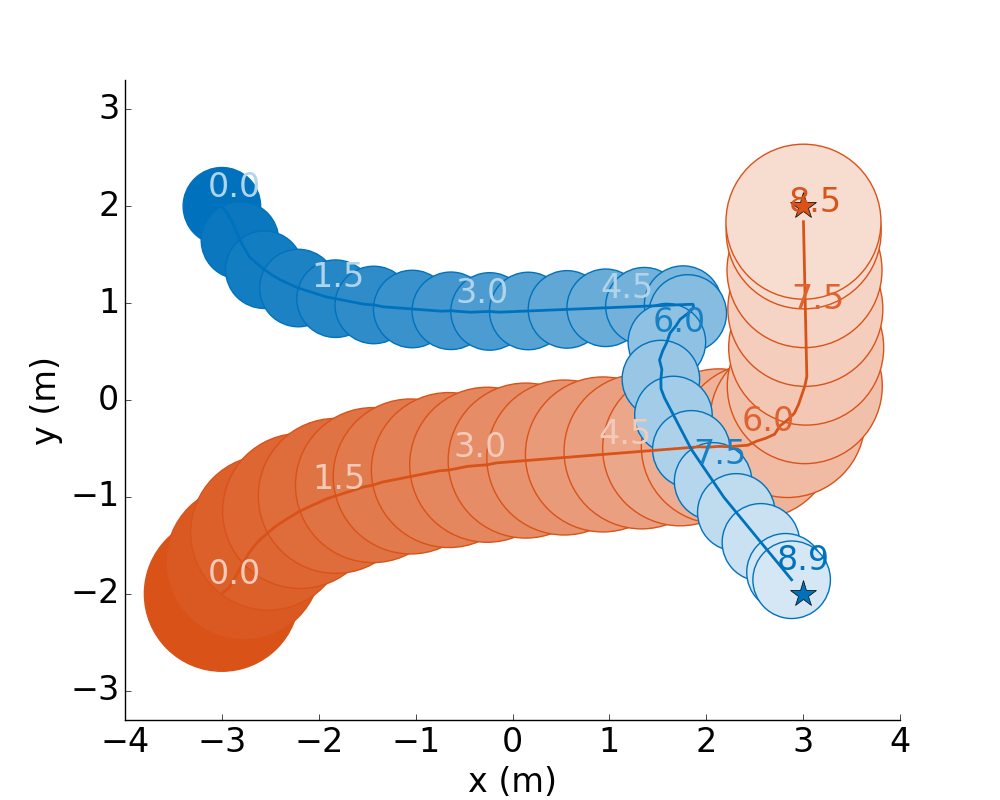}
		\includegraphics [trim=20 0 50 60, clip, width=0.24 \textwidth, angle = 0]{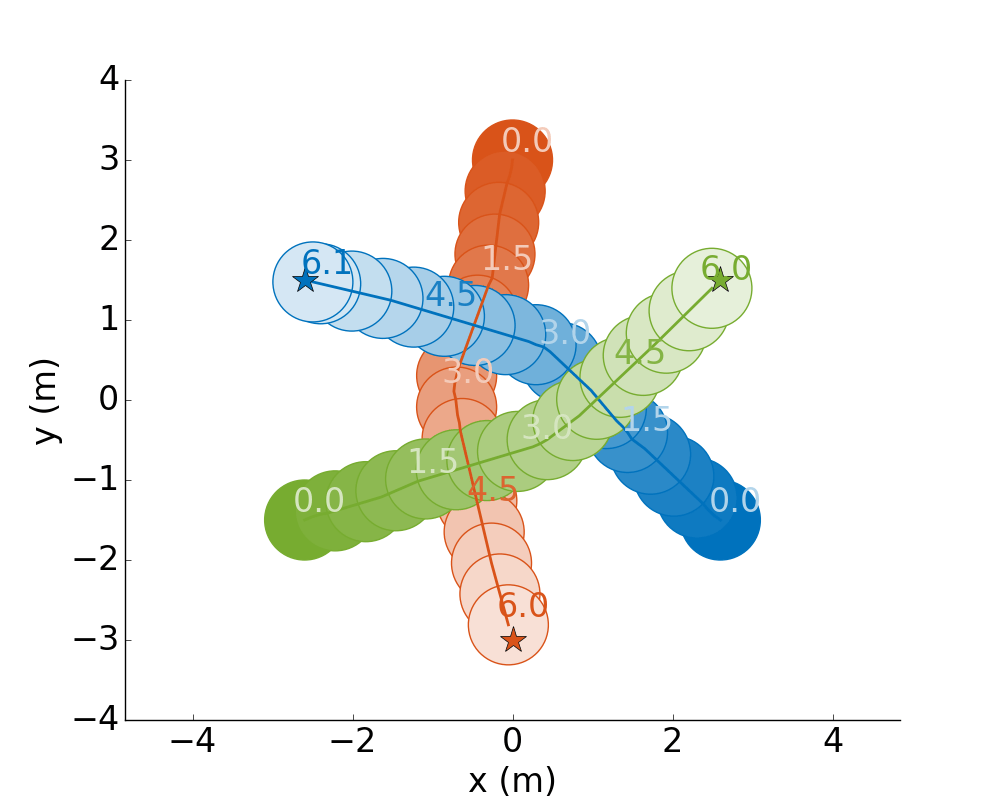}
		\includegraphics [trim=20 0 50 60, clip, width=0.24 \textwidth, angle = 0]{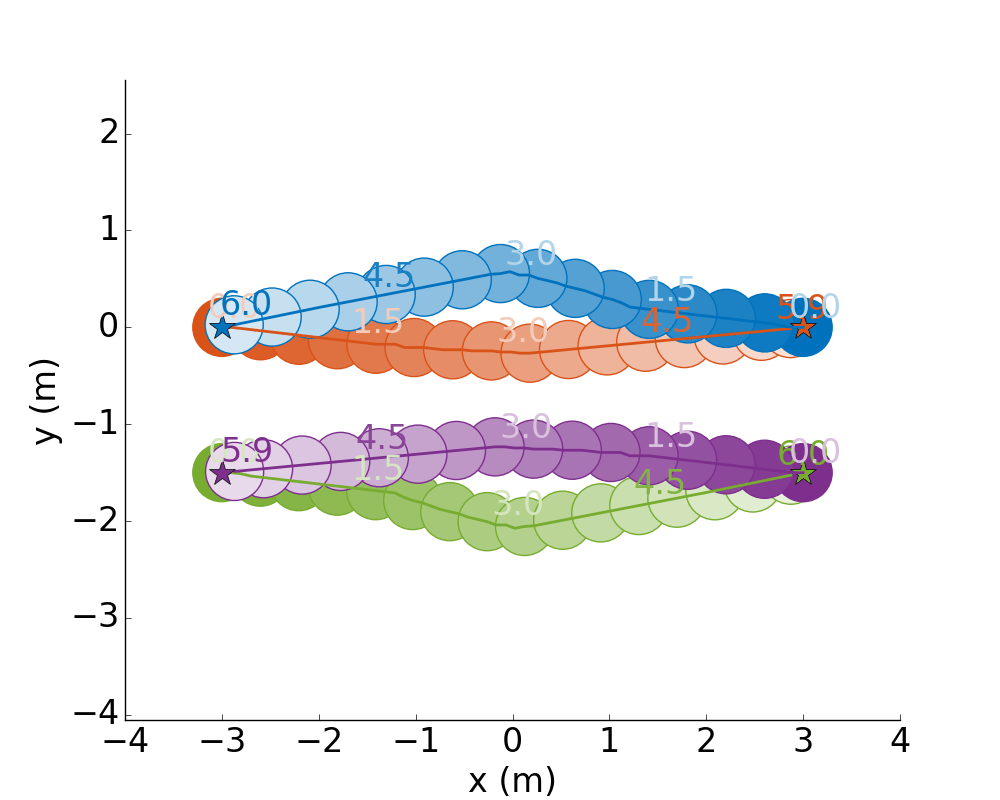}
		\includegraphics [trim=20 0 50 60, clip, width=0.24 \textwidth, angle = 0]{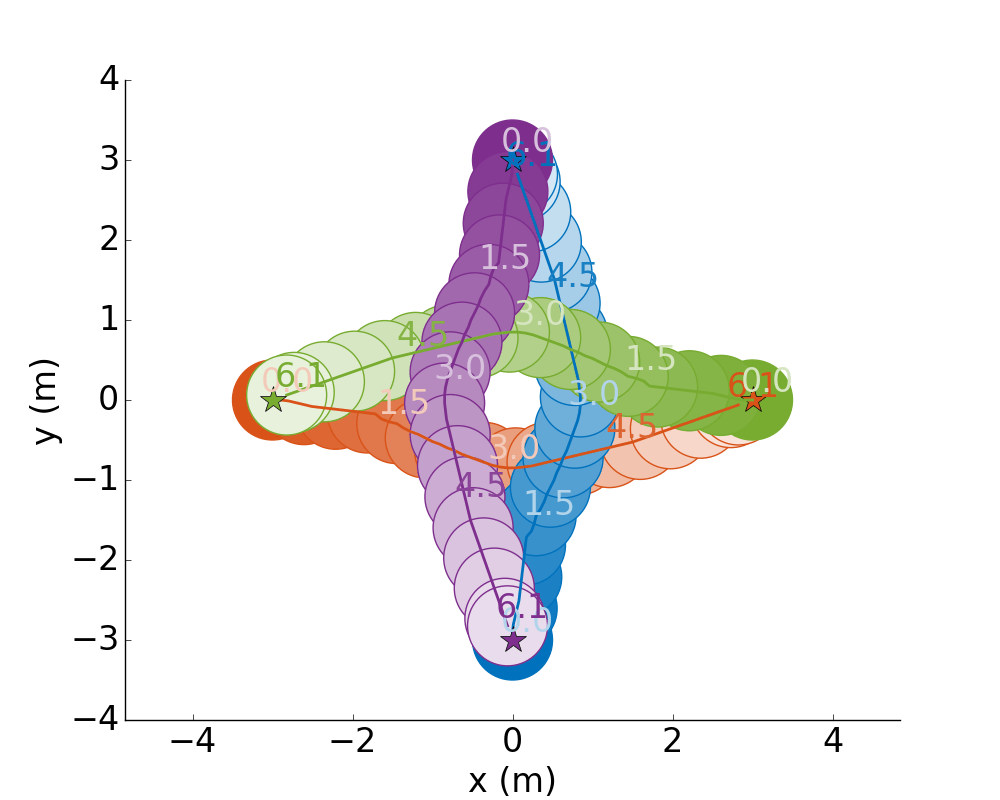}
		\caption{SA-CADRL trajectories with $n\in[2,3,4]$ agents}
		\label{fig:cadrl_traj_2_agent} 
	\end{subfigure}
	\caption{Scenarios with $n\leq4$ agents. The top row shows agents executing GA3C-CADRL-10-LSTM, and the bottom row shows same scenarios with agents using SA-CADRL. Circles lighten as time increases, the numbers represent the time at agent's position, and circle size represents agent radius. GA3C-CADRL agents are slightly less efficient, as they reach their goals slightly slower than SA-CADRL agents. However, the overall behavior is similar, and the more general GA3C-CADRL framework generates desirable behavior without many of the assumptions from SA-CADRL.}
	\label{fig:234agent_traj}
\end{figure*}

\begin{figure*}[t]
	\centering
	\begin{subfigure}{0.30\textwidth}
		\centering
		\includegraphics [trim=20 0 50 40, clip, width=\textwidth, angle = 0]{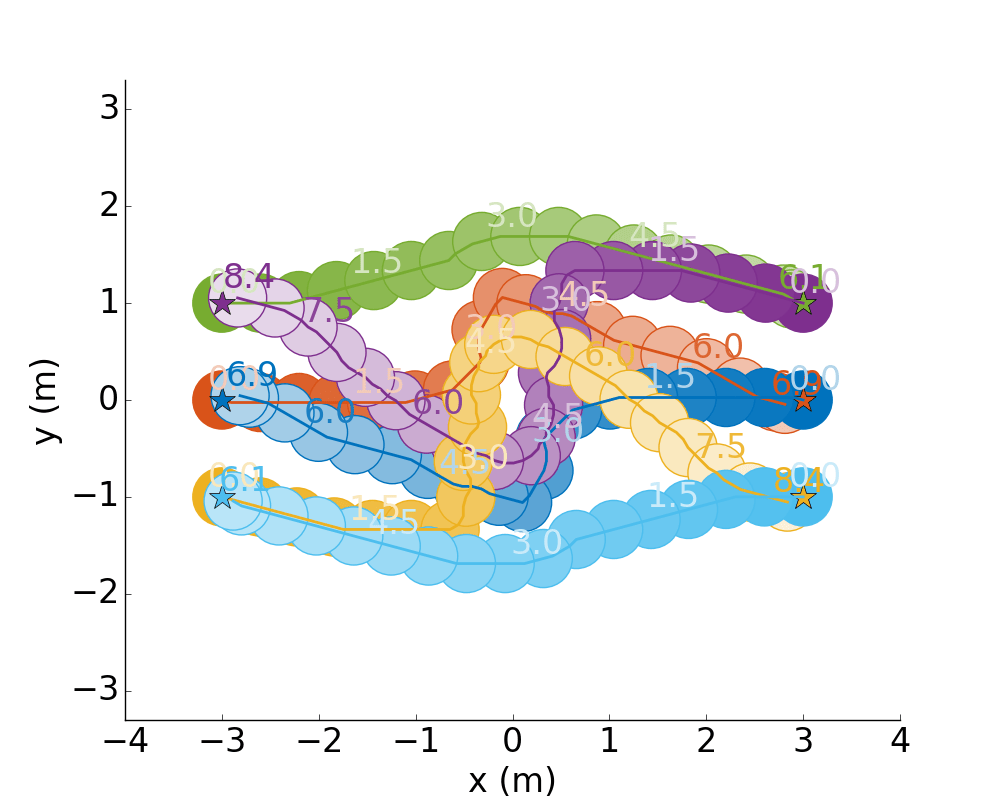}
		\caption{GA3C-CADRL: 3-pair swaps}
		\label{fig:ga3c_cadrl_traj_6_agent_swap} 
	\end{subfigure}
	\begin{subfigure}{0.30\textwidth}
		\centering
		\includegraphics [trim=20 0 50 40, clip, width=\textwidth, angle = 0]{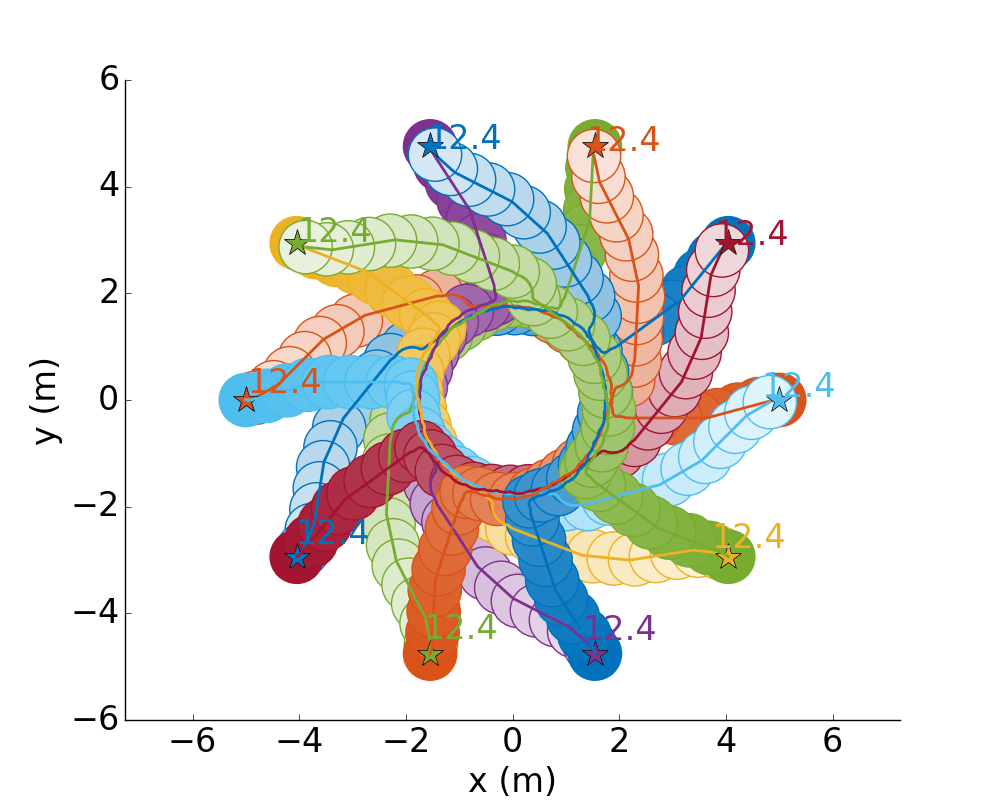}
		\caption{GA3C-CADRL: 10-agent circle}
		\label{fig:ga3c_cadrl_traj_10_agent_circle} 
	\end{subfigure}
	\begin{subfigure}{0.30\textwidth}
		\centering
		\includegraphics [trim=20 0 50 40, clip, width=\textwidth, angle = 0]{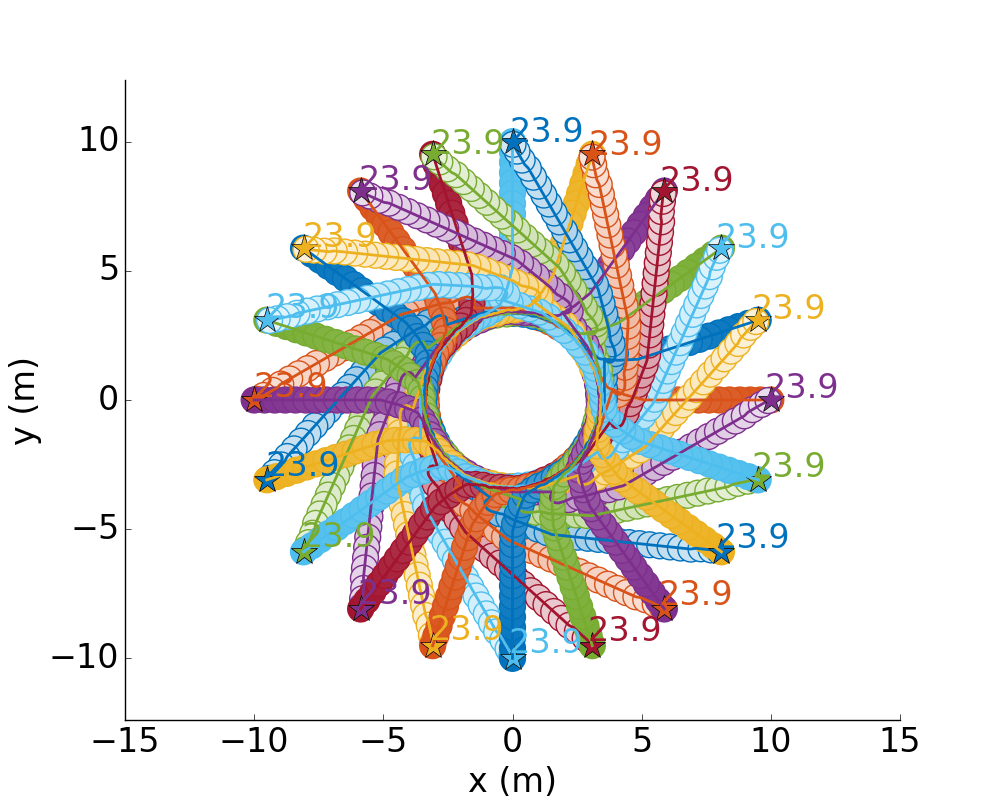}
		\caption{GA3C-CADRL: 20-agent circle}
		\label{fig:ga3c_cadrl_traj_20_agent_circle} 
	\end{subfigure}
	\begin{subfigure}{0.30\textwidth}
		\centering
		\includegraphics [trim=20 0 50 40, clip, width=\textwidth, angle = 0]{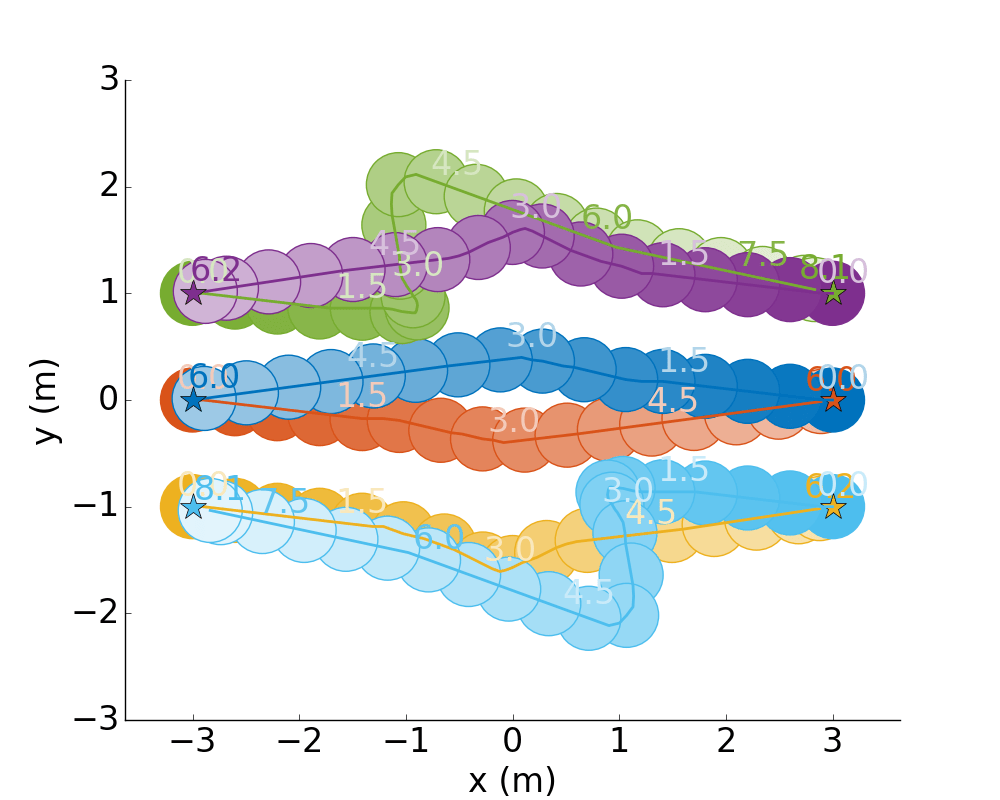}
		\caption{SA-CADRL: 3-pair swaps}
		\label{fig:sa_cadrl_traj_6_agent_swap} 
	\end{subfigure}
	\begin{subfigure}{0.30\textwidth}
		\centering
		\includegraphics [trim=20 0 50 40, clip, width=\textwidth, angle = 0]{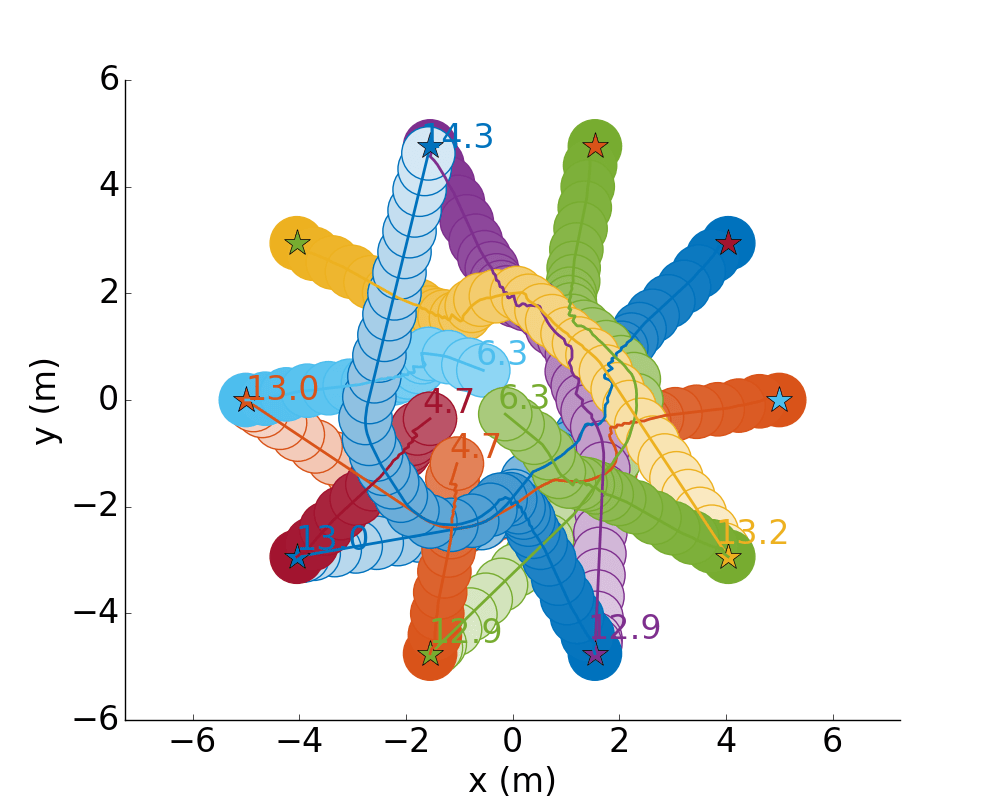}
		\caption{SA-CADRL: 10-agent circle}
		\label{fig:sa_cadrl_traj_10_agent_circle} 
	\end{subfigure}
	\begin{subfigure}{0.30\textwidth}
		\centering
		\includegraphics [trim=20 0 50 40, clip, width=\textwidth, angle = 0]{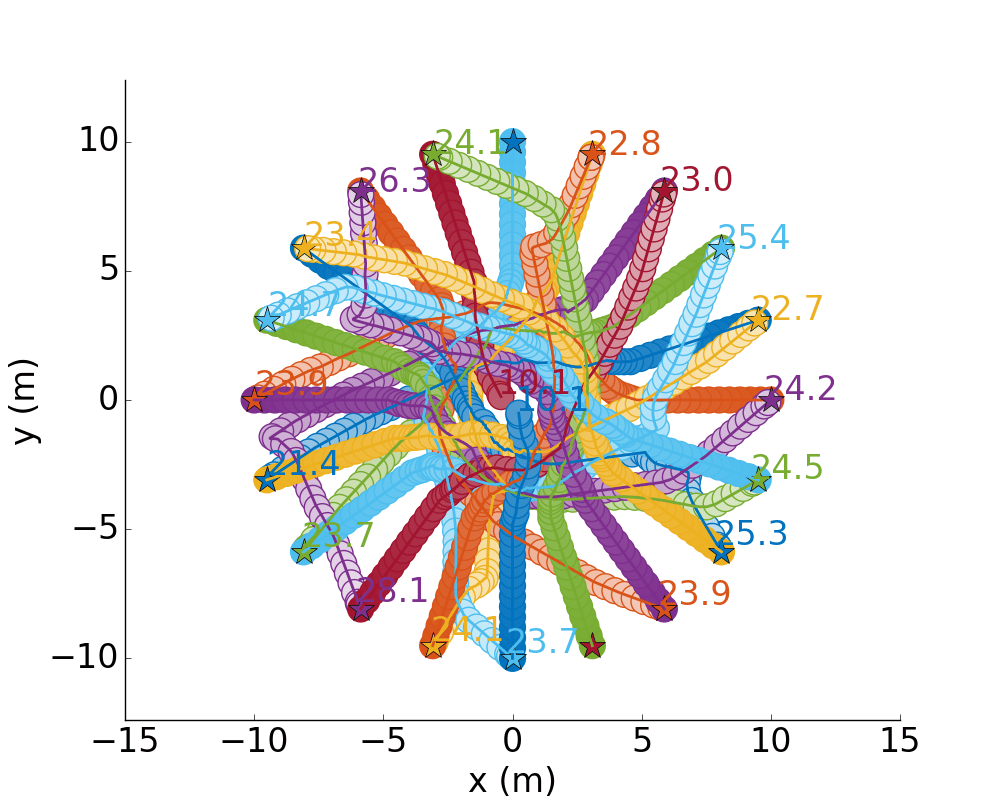}
		\caption{SA-CADRL: 20-agent circle}
		\label{fig:sa_cadrl_traj_20_agent_cirlce} 
	\end{subfigure}
	\caption{Scenarios with $n>4$ agents.
	In the 3-pair swap~\cref{fig:sa_cadrl_traj_6_agent_swap,fig:ga3c_cadrl_traj_6_agent_swap}, GA3C-CADRL agents exhibit interesting multiagent behavior: two agents form a pair while passing the opposite pair of agents.
	SA-CADRL agents reach the goal more quickly than GA3C-CADRL agents, but such multiagent behavior is a result of GA3C-CADRL agents having the capacity to observe all of the other 5 agents each time step.
	In other scenarios, GA3C-CADRL agents successfully navigate the 10- and 20-agent circles, whereas some SA-CADRL agents collide (near $(-1,\,-1)$ and $(0,\,0)$ in~\cref{fig:sa_cadrl_traj_10_agent_circle} and $(0,\,0)$ in~\cref{fig:sa_cadrl_traj_20_agent_cirlce}).}
	\label{fig:more_than_4agent_traj}
\end{figure*}

\definecolor{tab20c_0}{rgb}{0.19215686274509805,0.5098039215686274,0.7411764705882353}
\definecolor{tab20c_1}{rgb}{0.4196078431372549,0.6823529411764706,0.8392156862745098}
\definecolor{tab20c_2}{rgb}{0.6196078431372549,0.792156862745098,0.8823529411764706}
\definecolor{tab20c_3}{rgb}{0.7764705882352941,0.8588235294117647,0.9372549019607843}
\definecolor{tab20c_4}{rgb}{0.9019607843137255,0.3333333333333333,0.050980392156862744}
\definecolor{tab20c_5}{rgb}{0.9921568627450981,0.5529411764705883,0.23529411764705882}
\definecolor{tab20c_6}{rgb}{0.9921568627450981,0.6823529411764706,0.4196078431372549}
\definecolor{tab20c_7}{rgb}{0.9921568627450981,0.8156862745098039,0.6352941176470588}
\definecolor{tab20c_8}{rgb}{0.19215686274509805,0.6392156862745098,0.32941176470588235}
\definecolor{tab20c_9}{rgb}{0.4549019607843137,0.7686274509803922,0.4627450980392157}
\definecolor{tab20c_10}{rgb}{0.6313725490196078,0.8509803921568627,0.6078431372549019}
\definecolor{tab20c_11}{rgb}{0.7803921568627451,0.9137254901960784,0.7529411764705882}
\definecolor{tab20c_12}{rgb}{0.4588235294117647,0.4196078431372549,0.6941176470588235}
\definecolor{tab20c_13}{rgb}{0.6196078431372549,0.6039215686274509,0.7843137254901961}
\definecolor{tab20c_14}{rgb}{0.7372549019607844,0.7411764705882353,0.8627450980392157}
\definecolor{tab20c_15}{rgb}{0.8549019607843137,0.8549019607843137,0.9215686274509803}
\definecolor{tab20c_16}{rgb}{0.38823529411764707,0.38823529411764707,0.38823529411764707}
\definecolor{tab20c_17}{rgb}{0.5882352941176471,0.5882352941176471,0.5882352941176471}
\definecolor{tab20c_18}{rgb}{0.7411764705882353,0.7411764705882353,0.7411764705882353}
\definecolor{tab20c_19}{rgb}{0.8509803921568627,0.8509803921568627,0.8509803921568627}

\begin{figure*}[t]
	\centering
	\begin{subfigure}{\textwidth}
		\centering
		\begin{tikzpicture}[framed, background rectangle/.style={draw=black, rounded corners}]
		    \node [text centered, name=prior_work] {\underline{Prior Work}};
		    \node [text centered, right=1.5cm of prior_work, name=trained_4] {\underline{Trained on 2-4 Agents}};
		    \node [text centered, right=1.5cm of trained_4, name=trained_10] {\underline{Trained on 2-10 Agents}};

		    \node [fill=tab20c_16, below=0.4cm of prior_work.west, name=rvo_color] {};
		    \node [right=0.1cm of rvo_color, name=rvo] {ORCA};
		    \node [fill=tab20c_0, below=0.1cm of rvo_color, name=sa_cadrl_color] {};
		    \node [right=0.1cm of sa_cadrl_color, name=sa_cadrl] {SA-CADRL};

		    \node [fill=tab20c_4, below=0.4cm of trained_4.west, name=ga3c_cadrl_4_ws_4_color] {};
		    \node [right=0.1cm of ga3c_cadrl_4_ws_4_color, name=ga3c_cadrl_4_ws_4] {GA3C-CADRL-4-WS-4};
		    \node [fill=tab20c_5, below=0.1cm of ga3c_cadrl_4_ws_4_color, name=ga3c_cadrl_4_ws_6_color] {};
		    \node [right=0.1cm of ga3c_cadrl_4_ws_6_color, name=ga3c_cadrl_4_ws_6] {GA3C-CADRL-4-WS-6};
		    \node [fill=tab20c_6, below=0.1cm of ga3c_cadrl_4_ws_6_color, name=ga3c_cadrl_4_ws_8_color] {};
		    \node [right=0.1cm of ga3c_cadrl_4_ws_8_color, name=ga3c_cadrl_4_ws_6] {GA3C-CADRL-4-WS-8};
		    \node [fill=tab20c_9, below=0.1cm of ga3c_cadrl_4_ws_8_color, name=ga3c_cadrl_4_lstm_color] {};
		    \node [right=0.1cm of ga3c_cadrl_4_lstm_color, name=ga3c_cadrl_4_lstm] {GA3C-CADRL-4-LSTM};

		    \node [fill=tab20c_12, below=0.4cm of trained_10.west, name=ga3c_cadrl_10_ws_4_color] {};
		    \node [right=0.1cm of ga3c_cadrl_10_ws_4_color, name=ga3c_cadrl_10_ws_4] {GA3C-CADRL-10-WS-4};
		    \node [fill=tab20c_13, below=0.1cm of ga3c_cadrl_10_ws_4_color, name=ga3c_cadrl_10_ws_6_color] {};
		    \node [right=0.1cm of ga3c_cadrl_10_ws_6_color, name=ga3c_cadrl_10_ws_6] {GA3C-CADRL-10-WS-6};
		    \node [fill=tab20c_14, below=0.1cm of ga3c_cadrl_10_ws_6_color, name=ga3c_cadrl_10_ws_8_color] {};
		    \node [right=0.1cm of ga3c_cadrl_10_ws_8_color, name=ga3c_cadrl_10_ws_6] {GA3C-CADRL-10-WS-8};
		    \node [fill=tab20c_8, below=0.1cm of ga3c_cadrl_10_ws_8_color, name=ga3c_cadrl_10_lstm_color] {};
		    \node [right=0.1cm of ga3c_cadrl_10_lstm_color, name=ga3c_cadrl_10_lstm] {GA3C-CADRL-10-LSTM};

		\end{tikzpicture}
		\label{fig:multi_agent_stats_legend} 
	\end{subfigure}\\
	\begin{subfigure}{\textwidth}
		\centering
		\includegraphics [trim=0 0 0 0, clip, width=\textwidth, angle = 0]{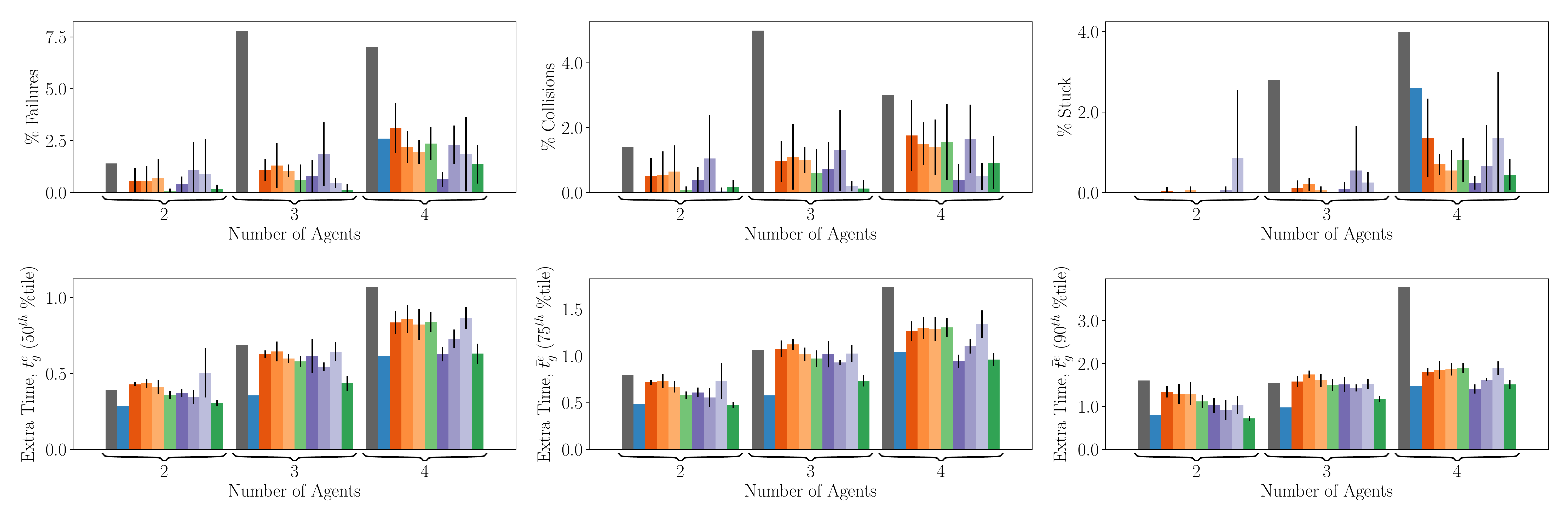}
		\caption{Comparison of Algorithms for $n\leq4$ agents}
		\label{fig:multi_agent_stats_nleq4} 
	\end{subfigure}\\
	\begin{subfigure}{\textwidth}
		\centering
		\includegraphics [trim=0 0 0 0, clip, width=\textwidth, angle = 0]{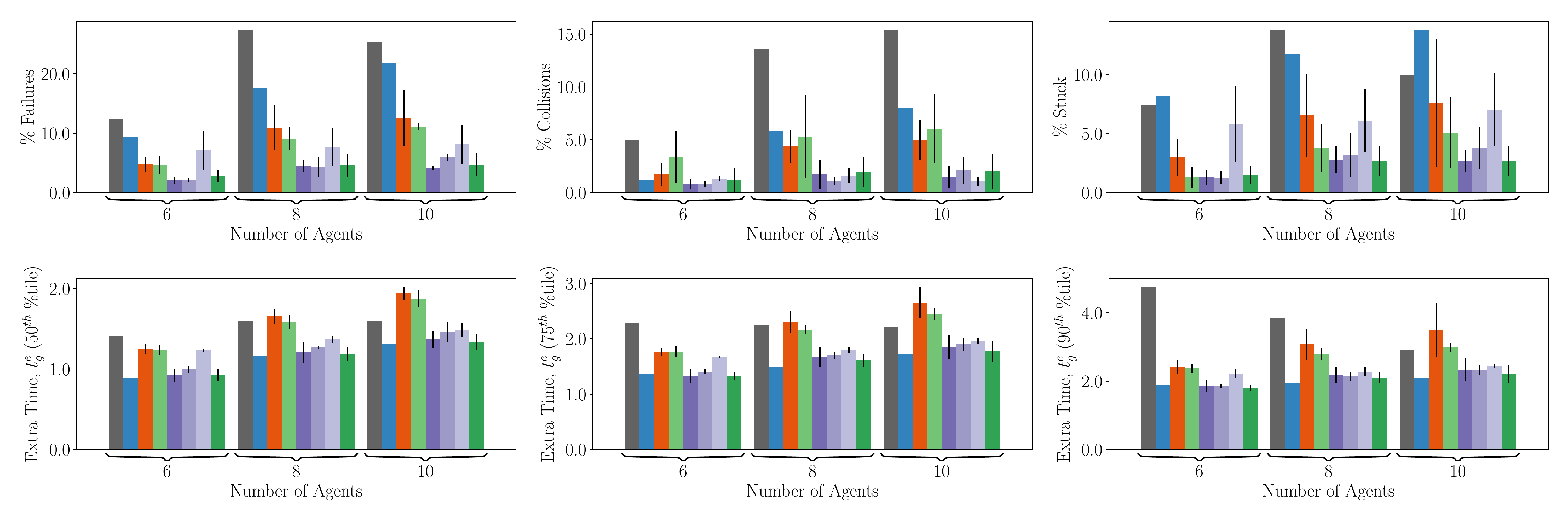}
		\caption{Comparison of Algorithms for $n>4$ agents}
		\label{fig:multi_agent_stats_ngt4}
	\end{subfigure}
	\caption{
	\changed{
	Numerical comparison on the same 500 random test cases (lower is better).
	The \textcolor{tab20c_8}{GA3C-CADRL-10-LSTM} network shows comparable performance to \textcolor{tab20c_0}{SA-CADRL} for small $n$ (a), much better performance for large $n$ (b), and better performance than model-based \textcolor{tab20c_16}{ORCA} for all $n$.
	Several ablations highlight SA-CADRL and GA3C-CADRL differences.
	With the same architecture (\textcolor{tab20c_0}{SA-CADRL} \& \textcolor{tab20c_4}{GA3C-CADRL-4}), the GA3C policy performs better for large $n$ (b), but worsens performance for small $n$ (a).
	Adding a second phase of training with up to 10 agents (\textcolor{tab20c_12}{GA3C-CADRL-10-4}) improves performance for all $n$ tested.
	Adding additional pre-defined agent capacity to the network (\textcolor{tab20c_13}{GA3C-CADRL-10-6}, \textcolor{tab20c_14}{GA3C-CADRL-10-8}) can degrade performance.
	The LSTM (\textcolor{tab20c_8}{GA3C-CADRL-10-LSTM}) adds flexibility in prior knowledge on number of agents, maintaining similar performance to the WS approaches for large $n$ and recovering comparable performance to \textcolor{tab20c_0}{SA-CADRL} for small $n$.
	}
	}
	\label{fig:multi_agent_stats}
\end{figure*}

\subsection{Computational Details}

The DNNs in this work were implemented with TensorFlow~\citep{abadi2016tensorflow} in Python.
Each query of the GA3C-CADRL network only requires the current state vector, and takes on average 0.4-0.5ms on a i7-6700K CPU, which is approximately 20 times faster than before~\citep{Chen17_IROS}.
Note that a GPU is not required for online inference in real time, and CPU-only training was faster than hybrid CPU-GPU training on our hardware.

In total, the RL training converges in about 24 hours (after $2~\cdot~10^{6}$ episodes) for the multiagent, LSTM network on a computer with an i7-6700K CPU with 32 parallel environment threads.
A limiting factor of the training time is the low learning rate required for stable training.
Recall that earlier approaches~\citep{Chen17_IROS} took 8 hours to train a 4-agent value network, but now the network learns both the policy and value function and without being provided any structure about the other agents' behaviors.
The larger number of training episodes can also be attributed to the stark contrast in initial policies upon starting RL between this and the earlier approach: CADRL was fine-tuning a decent policy, whereas GA3C-CADRL learns collision avoidance entirely in the RL phase.

The performance throughout the training procedure is shown as the ``closest last'' curve in~\cref{fig:training_curve} (the other curves are explained in~\cref{sec:results:lstm_ordering}).
The mean $\pm 1\sigma$ rolling reward over 5 training runs is shown.
After initialization, the agents receive on average 0.15 reward per episode.
After RL phase 1 (converges in $1.5\cdot10^6$ episodes), they average 0.90 rolling reward per episode. 
When RL phase 2 begins, the domain becomes much harder ($n_{max}$ increases from 4 to 10), and rolling reward increases until converging at 0.93 (after a total of $1.9\cdot10^6$ episodes).
Rolling reward is computed as the sum of the rewards accumulated in each episode, averaged across all GA3C-CADRL agents in that episode, averaged over a window of recent episodes.
Rolling reward is only a measure of success/failure, as it does not include the discount factor and thus is not indicative of time efficiency.
Because the maximum receivable reward on any episode is 1, an average reward $<1$ implies there are some collisions (or other penalized behavior) even after convergence.
This is expected, as agents sample from their policy distributions when selecting actions in training, so there is always a non-zero probability of choosing a sub-optimal action in training.
Later, when executing a trained policy, agents select the action with highest probability.

Key hyperparameter values include: learning rate $L_r = 2\cdot 10^{-5}$, entropy coefficient $\beta = 1 \cdot 10^{-4}$, discount $\gamma = 0.97$, training batch size $b_s = 100$, and we use the Adam optimizer~\citep{kingma2014adam}.

\subsection{Simulation Results}

\subsubsection{Baselines}
This section compares the proposed GA3C-CADRL algorithm to ORCA~\citep{berg_reciprocal_2011}, SA-CADRL~\citep{Chen17_IROS}, and, where applicable, DRLMACA~\citep{long2018towards}.

\added{
We first briefly summarize the ORCA, SA-CADRL, and DRLMACA algorithms.
In ORCA, agents solve a one-step optimization problem to make a minimal adjustment to the desired velocity vector, such that the new velocity does not collide with other agents in the future (assuming they travel at a constant velocity).
Because the one-step horizon and constant velocity assumption leads to mypoic planning, SA-CADRL improves on that approach by learning a value function that encodes the time-to-goal from various states.
Thus, the online optimization of SA-CADRL also considers the long-horizon impact of a local control command, via a quick lookup (DNN query).
Like SA-CADRL, DRLMACA also uses deep RL, but the inputs to the policy include raw sensor data (laserscans) to inform collision avoidance, rather than using agent position, velocity, and radius estimates as in SA-CADRL and this work.
}

In our experiments, ORCA agents must inflate agent radii by $5\%$ to reduce collisions caused by numerical issues.
Without this inflation, over $50\%$ of experiments with 10 ORCA agents had a collision.
This inflation led to more ORCA agents getting stuck, which is better than a collision in most applications.
The time horizon parameter in ORCA impacts the trajectories significantly; it was set to $5$ seconds.

Although the original 2-agent CADRL algorithm~\citep{chen_decentralized_2017} was also shown to scale to multiagent scenarios, its minimax implementation is limited in that it only considers one neighbor at a time as described in~\citep{Chen17_IROS}.
For that reason, this work focuses on the comparison against SA-CADRL which has better multiagent properties - the policy used for comparison is the same one that was used on the robotic hardware in~\citep{Chen17_IROS}.
That particular policy was trained with some noise in the environment ($\mathbf{p} = \mathbf{p}_{actual} + \sigma$) which led to slightly poorer performance than the ideally-trained network as reported in the results of~\citep{Chen17_IROS}, but more acceptable hardware performance.

The version of the new GA3C-CADRL policy after RL phase 2 is denoted GA3C-CADRL-10, as it was trained in scenarios of up to 10 agents.
To create a more fair comparison with SA-CADRL which was only trained with up to 4 agents, let GA3C-CADRL-4 denote the policy after RL phase 1 (which only involves scenarios of up to 4 agents). 
Recall GA3C-CADRL-4 can still be naturally implemented on $n>4$ agent cases, whereas SA-CADRL can only accept up to 3 nearby agents' states regardless of $n$.

\subsubsection{\texorpdfstring{$n \leq 4$}{n <= 4} agents: \changed{Numerical Comparison to Baselines}}
The previous approach (SA-CADRL) is known to perform well on scenarios involving a few agents ($n \leq 4$), as its trained network can accept up to 3 other agents' states as input.
Therefore, a first objective is to confirm that the new algorithm can still perform comparably with small numbers of agents.
This is not a trivial check, as the new algorithm is not provided with any structure/prior about the world's dynamics, so the learning is more difficult.

Trajectories are visualized in~\cref{fig:234agent_traj}: the top row shows scenarios with agents running the new policy (GA3C-CADRL-10-LSTM), and the bottom row shows agents in identical scenarios but using the old policy (SA-CADRL).
The colors of the circles (agents) lighten as time increases and the circle size represents agent radius.
The trajectories generally look similar for both algorithms, with SA-CADRL being slightly more efficient.
A rough way to assess efficiency in these plotted paths is time indicated when the agents reach their goals.

Although it is easy to pick out interesting pros/cons for any particular scenario, it is more useful to draw conclusions after aggregating over a large number of randomly-generated cases.
Thus, we created test sets of 500 random scenarios, defined by ($p_{start}$, $p_{goal}$, $r$, $v_{pref}$) per agent, for many different numbers of agents.
Each algorithm is evaluated on the same 500 test cases.
The comparison metrics are the percent of cases with a collision, percent of cases where an agent gets stuck and doesn't reach the goal, and the remaining cases where the algorithm was successful, the average extra time to goal, $\bar{t}^{e}_g$ beyond a straight path at $v_{pref}$.
These metrics provide measures of efficiency and safety.

\changed{
Aggregated results in~\cref{fig:multi_agent_stats} compare a model-based algorithm, ORCA~\citep{berg_reciprocal_2011}, SA-CADRL~\citep{Chen17_IROS}, and several variants of the new GA3C-CADRL algorithm.
With $n \leq 4$ agents in the environment (a), SA-CADRL has the lowest $\bar{t}^{e}_g$, and the agents rarely fail in these relatively simple scenarios.
}

\subsubsection{\changed{\texorpdfstring{$n \leq 4$}{n <= 4} agents: Ablation Study}}
\changed{
There are several algorithmic differences between SA-CADRL and GA3C-CADRL: we compare each ablation one-by-one.
With the same network architecture (pre-defined number of agents with weights shared (WS) for all agents), GA3C-CADRL-4-WS-4 loses some performance versus SA-CADRL, since GA3C-CADRL must learn the notion of a collision, whereas SA-CADRL's constant-velocity collision checking may be reasonable for small $n$.
Replacing the WS network head with LSTM improves the performance when the number of agents is below network capacity, potentially because the LSTM eliminates the need to pass ``dummy'' states to fill the network input vector.
Lastly, the second training phase (2-10 agents) improves policy performance even for small numbers of agents.
}

\changed{
Overall, the GA3C-CADRL-10-LSTM variant performs comparably, though slightly worse, than SA-CADRL for small numbers of agents, and outperforms the model-based ORCA algorithm.
}

\subsubsection{\texorpdfstring{$n > 4$}{n > 4} agents: \changed{Numerical Comparison to Baselines}}

A real robot will likely encounter more than 3 pedestrians at a time in a busy environment.
\changed{
However, experiments with the SA-CADRL algorithm suggest that increasing the network capacity beyond 4 total agents causes convergence issues.
}
Thus, the approach taken here for SA-CADRL is to supply only the closest 3 agents' states in crowded scenarios.
\changed{
The GA3C-CADRL policy's convergence is not as sensitive to the maximum numbers of agents, allowing an evaluation of whether simply expanding the network input size improves performance in crowded scenarios.
Moreover, the LSTM variant of GA3C-CADRL relaxes the need to pre-define a maximum number of agents, as any number of agents can be fed into the LSTM and the final hidden state can still be taken as a representation of the world configuration.
}

\begin{figure*}[t]
	\centering
	\begin{subfigure}{0.5\linewidth}
		\centering
		\includegraphics [trim=0 0 0 0, clip, width=\textwidth, angle = 0]{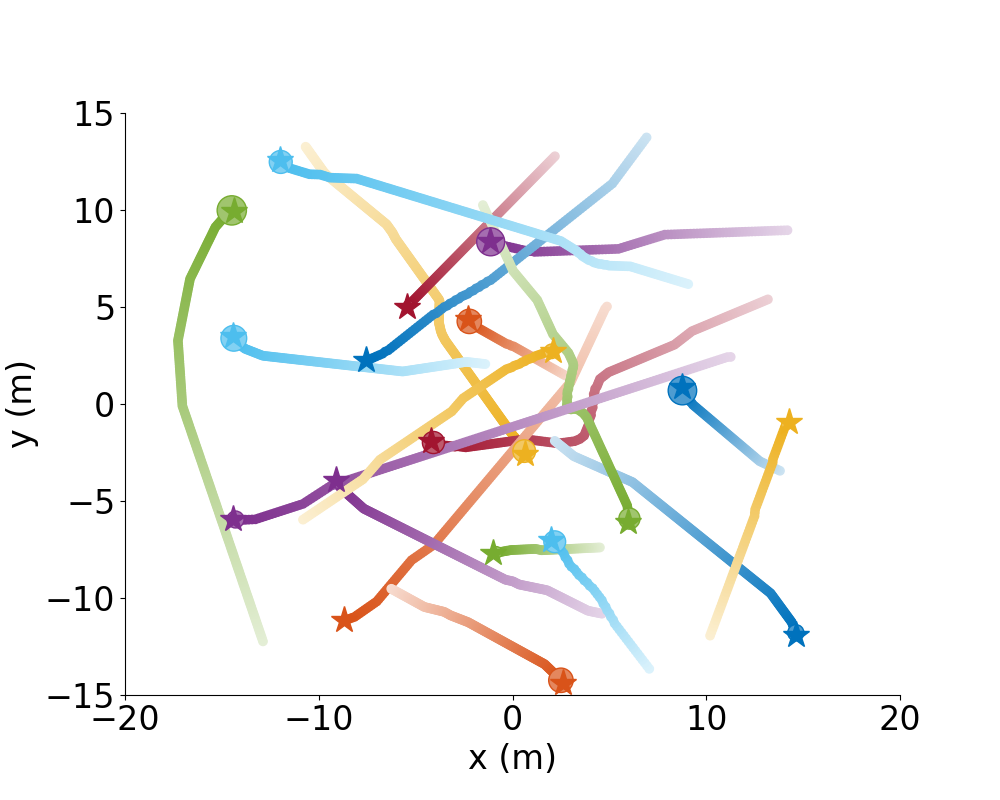}
		\caption{20-agent Scenario 1}
		\label{fig:20_agents_scenario1}
	\end{subfigure}%
	\begin{subfigure}{0.5\linewidth}
		\centering
		\includegraphics [trim=0 0 0 0, clip, width=\textwidth, angle = 0]{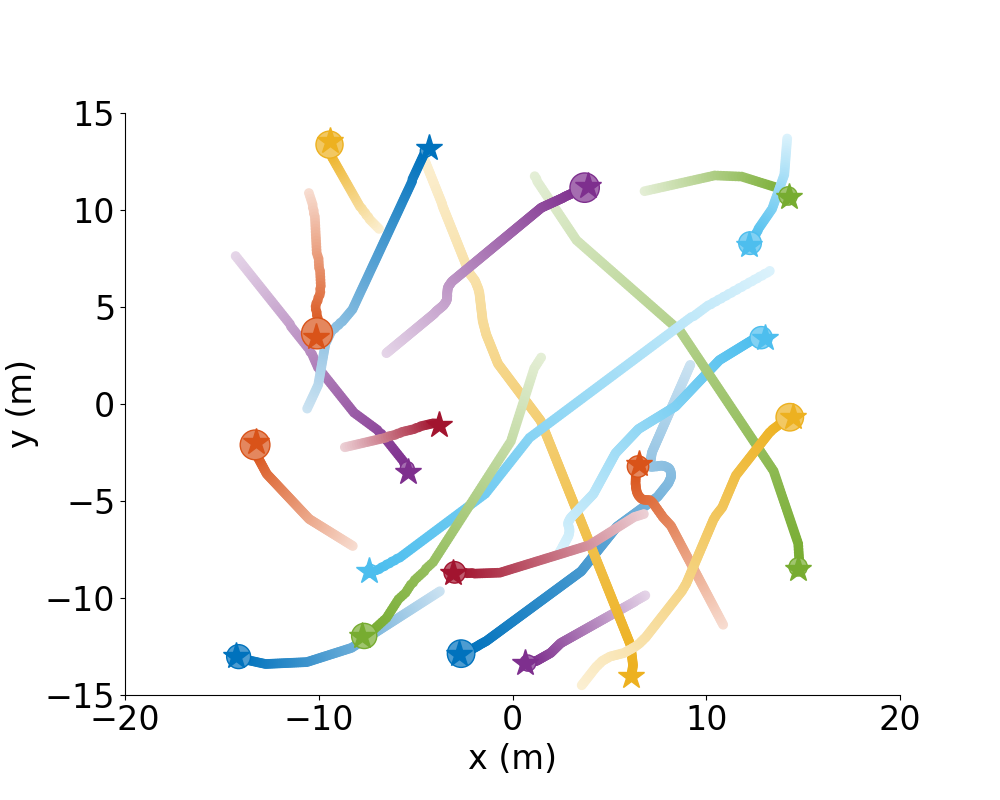}
		\caption{20-agent Scenario 2}
		\label{fig:20_agents_scenario2}
	\end{subfigure}\\
	\begin{subfigure}{0.5\linewidth}
		\centering
		\includegraphics [trim=0 0 0 0, clip, width=\textwidth, angle = 0]{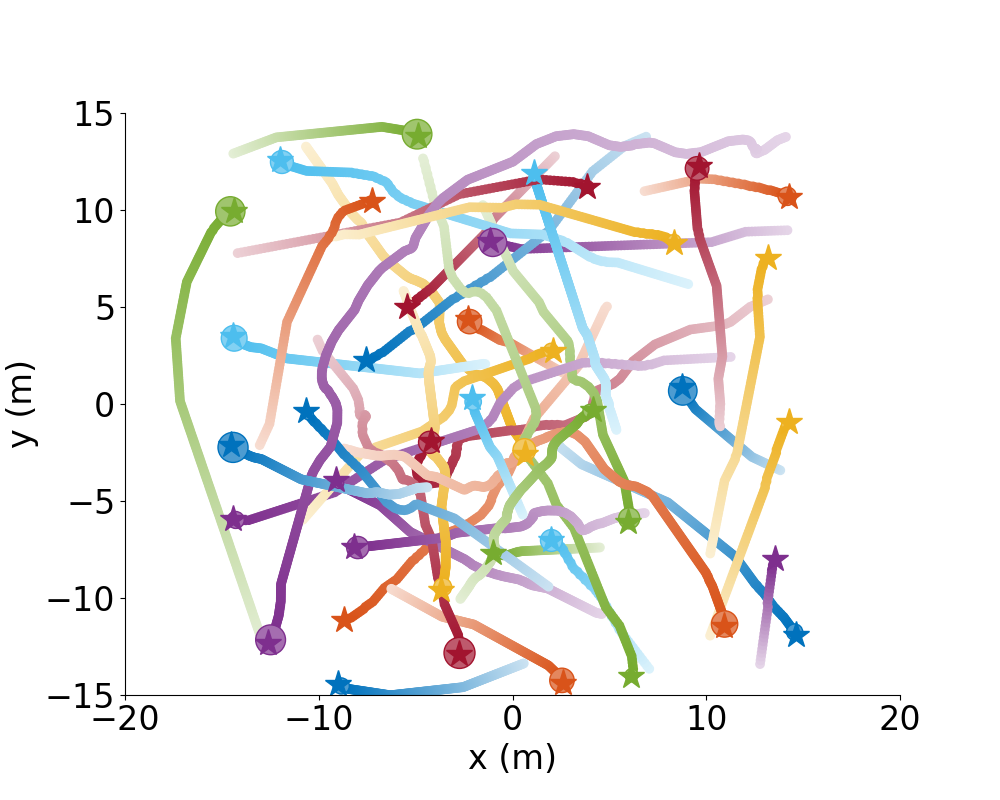}
		\caption{40-agent Scenario 1}
		\label{fig:40_agents_scenario1}
	\end{subfigure}%
	\begin{subfigure}{0.5\linewidth}
		\centering
		\includegraphics [trim=0 0 0 0, clip, width=\textwidth, angle = 0]{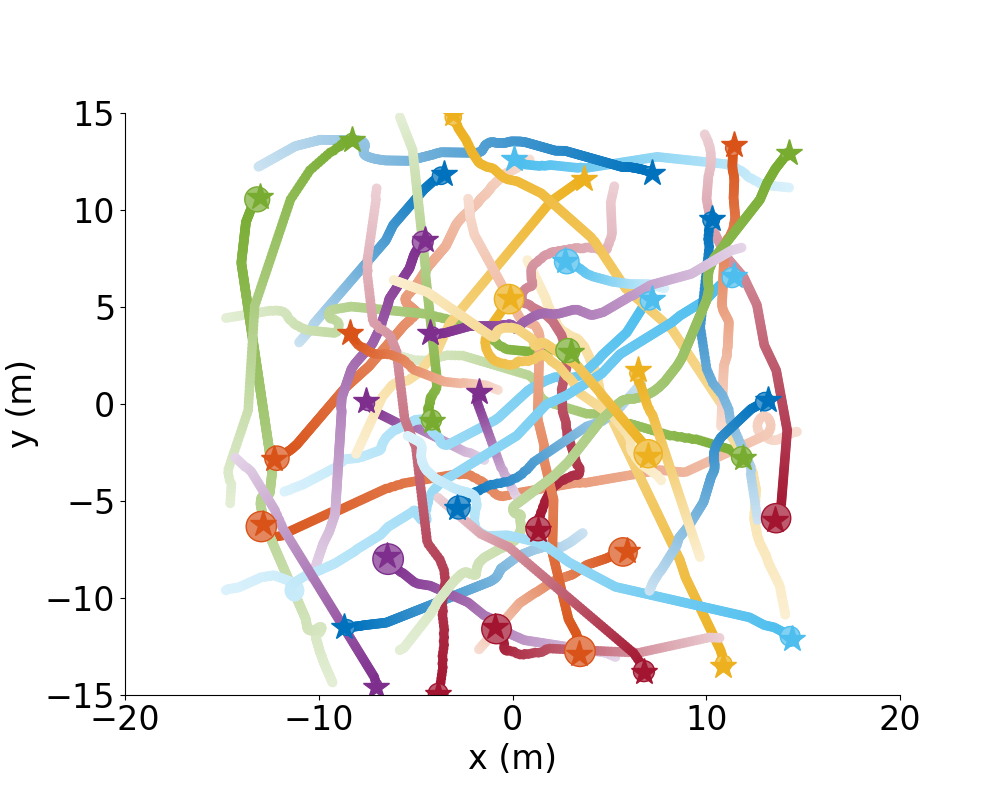}
		\caption{40-agent Scenario 2}
		\label{fig:40_agents_scenario2}
	\end{subfigure}
	\caption{
	\added{Random 20- and 40-agent scenarios.
	These figures highlight the learned policy's ability to handle large numbers of agents with a single LSTM-based representation (GA3C-CADRL-10).
	All agents reach their goals without collision in (a)-(d).}
	}
	\label{fig:many_agents_cases}
\end{figure*}

Even in $n>4$-agent environments, interactions still often only involve a couple of agents at a time.
Some specific cases where there truly are many-agent interactions are visualized in~\cref{fig:more_than_4agent_traj}.
In the 6-agent swap (left), GA3C-CADRL agents exhibit interesting multiagent behavior: the bottom-left and middle-left agents form a pair while passing the top-right and middle-right agents.
This phenomenon leads to a particularly long path for bottom-left and top-right agents, but also allows the top-left and bottom-right agents to not deviate much from a straight line.
In contrast, in SA-CADRL the top-left agent starts moving right and downward, until the middle-right agent becomes one of the closest 3 neighbors.
The top-left agent then makes an escape maneuver and passes the top-right on the outside.
In this case, SA-CADRL agents reach the goal more quickly than GA3C-CADRL agents, but the interesting multiagent behavior is a result of GA3C-CADRL agents having the capacity to observe all of the other 5 agents each time step, rather than SA-CADRL which just uses the nearest 3 neighbors. GA3C-CADRL agents successfully navigate the 10- and 20-agent circles (antipodal swaps), whereas several SA-CADRL agents get stuck or collide\footnote{Note there is not perfect symmetry in these SA-CADRL cases: small numerical fluctuations affect the choice of the closest agents, leading to slightly different actions for each agent. And after a collision occurs with a pair of agents, symmetry will certainly be broken for future time steps.}.

\changed{Statistics across 500 random cases of 6, 8, and 10 agents are shown in~\cref{fig:multi_agent_stats_ngt4}}.
The performance gain by using GA3C-CADRL becomes larger as the number of agents in the environment increases.
\changed{For $n=6,8,10$, GA3C-CADRL-10-LSTM shows a 3-4x reduction in failed cases with similar $\bar{t}^{e}_g$ compared to SA-CADRL.}
GA3C-CADRL-10-LSTM's percent of success remains above 95\% across any $n\leq10$, whereas SA-CADRL drops to under 80\%.
It is worth noting that SA-CADRL agents' failures are more often a result of getting stuck rather than colliding with others, however neither outcomes are desirable.
\changed{The GA3C-CADRL variants outperform model-based ORCA for large $n$ as well.}
The domain size of $n=10$ agent scenarios is set to be larger ($12\times12$ vs. $8\times8m$) than cases with smaller $n$ to demonstrate cases where $n$ is large but the world is not necessarily more densely populated with agents.

\added{
Furthermore, \cref{fig:many_agents_cases} shows that the learned policy generalizes beyond the 2-10 agent scenarios it was trained on.
In particular, \cref{fig:20_agents_scenario1,fig:20_agents_scenario2} show 2 random 20-agent scenarios and \cref{fig:40_agents_scenario1,fig:40_agents_scenario2} show 2 random 40-agent scenarios.
In addition to the examples shown, across 10 random trials of 20-agent scenarios, all agents reached their goals successfully (0 collisions or stuck agents).
Across 10 random trials of 40-agent scenarios, only 4/400 trajectories ended in collisions, and 1 agent became stuck, with 395/400 agents reaching their goals successfully.
}

\subsubsection{\changed{\texorpdfstring{$n > 4$}{n > 4} agents: Ablation Study}}
\changed{
We now discuss the GA3C-CADRL variants.
For large $n$, GA3C-CADRL-WS-4 strongly outperforms SA-CADRL.
Since the network architectures and number of agents trained on are the same, the performance difference highlights the benefit of the policy-based learning framework.
Particularly for large $n$, the multiagent interactions cause SA-CADRL's constant velocity assumption about other agents (to convert the value function to a policy in~\cref{eqn:cadrl_policy}) to become unrealistic.
Replacing the WS network head with LSTM (which accepts observations of any number of agents) causes slight performance improvement for $n=6,8$ (GA3C-CADRL-4-WS-4 vs. GA3C-CADRL-4-LSTM).
Since GA3C-CADRL-4-WS-6,8 never saw $n>4$ agents in training, these are omitted from~\cref{fig:multi_agent_stats_ngt4} (as expected, their performance is awful).
The second training phase (on up to 10 agents) leads to another big performance improvement (GA3C-CADRL-4-* vs. GA3C-CADRL-10-*).
The additional network capacity of the WS approaches (GA3C-CADRL-WS-4,6,8) appears to have some small, in some cases negative, performance impact.
That observation suggests that simply increasing the maximum number of agents input to the network does not address the core issues with multiagent collision avoidance.
The GA3C-CADRL-10-LSTM variant performs similarly to GA3C-CADRL-10-WS-4, while providing a more flexible network architecture (and better performance for small $n$, as described earlier).
}

\begin{figure*}[t]
	\centering
	\begin{subfigure}{0.49\textwidth}
		\centering
		\includegraphics [trim=20 0 50 60, clip, width=\textwidth, angle = 0]{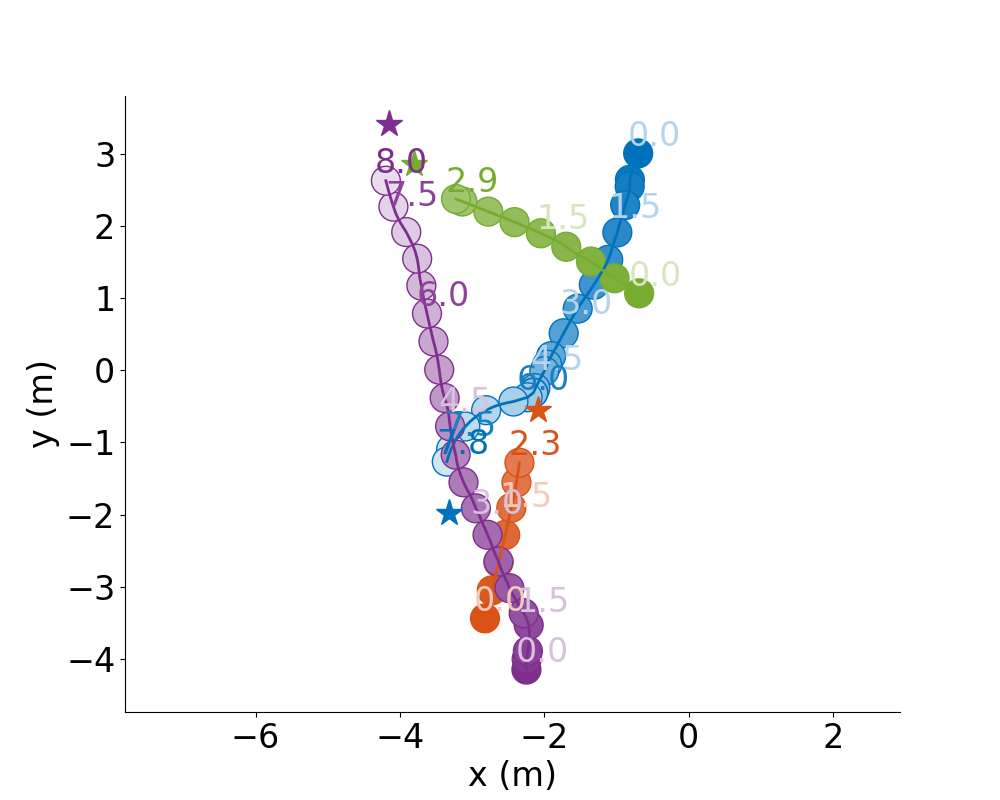}
		\caption{DRLMACA trajectory with $4$ agents}
		\label{fig:drlmaca_traj_r0.2} 
	\end{subfigure}
	\begin{subfigure}{0.49\textwidth}
		\centering
		\includegraphics [trim=20 0 50 60, clip, width=\textwidth, angle = 0]{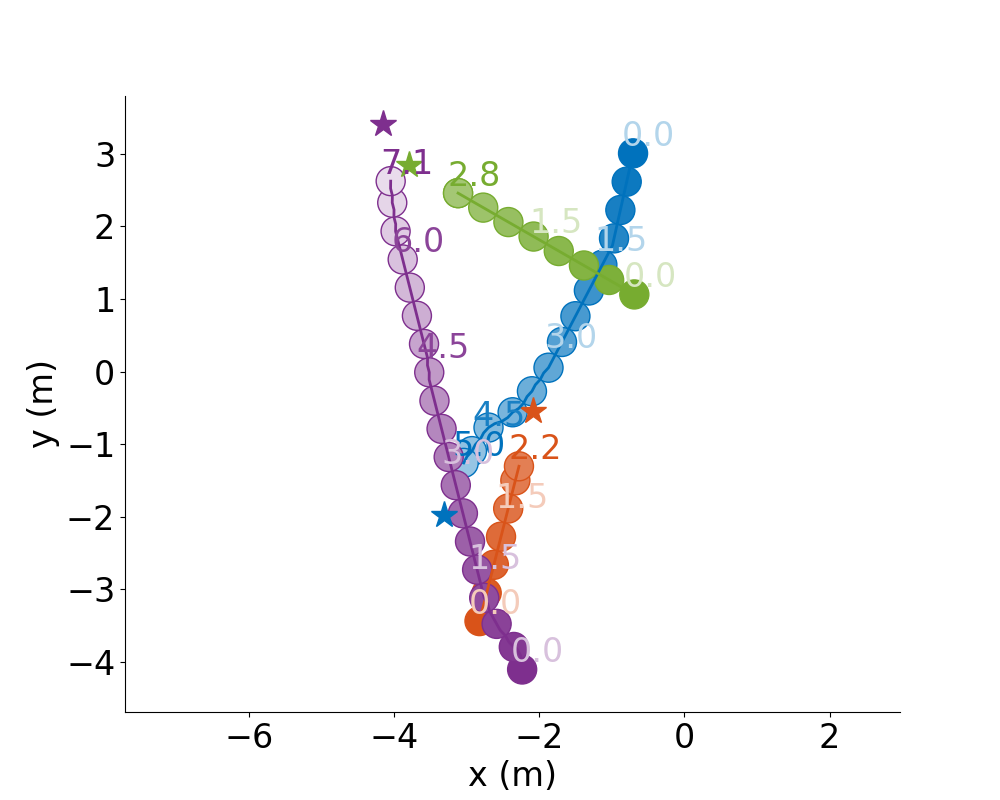}
		\caption{GA3C-CADRL trajectory with $4$ agents}
		\label{fig:ga3c_cadrl_traj_r0.2}
	\end{subfigure}
	\caption{
	GA3C-CADRL and DRLMACA 4-agent trajectories.
	Both algorithms produce collision-free paths; numbers correspond to the timestamp agents achieved that position.
	GA3C-CADRL agents are slightly faster; the bottom-most DRLMACA agent (left) slows down near the start of the trajectory, leading to a larger time to goal (8.0 vs. 7.1 seconds), whereas the bottom-most GA3C-CADRL agent (right) cuts behind another agent near its $v_{pref}$.
	Similarly, the top-right DRLMACA agent slows down near $(-2,0)$ (overlapping circles), whereas the top-right GA3C-CADRL agent maintains high speed and reaches the goal faster (7.8 vs. 5.0 seconds).
	Agents stop moving once within $0.8m$ of their goal position in these comparisons.
	}
	\label{fig:drlmaca_ga3c_trajs}
\end{figure*}

\begin{table*}[t]
	\centering
	\caption[]{
	Performance of ORCA~\citep{berg_reciprocal_2011}, SA-CADRL~\citep{Chen17_IROS}, DRLMACA~\citep{long2018towards}, and GA3C-CADRL (new) algorithms on the same 100 random test cases, for various agent radii, with $v_{pref}=1.0 m/s$ for all agents.
	For both $r=0.2m$ and $r=0.5m$, GA3C-CADRL outperforms DRLMACA, and DRLMACA performance drops heavily for $r=0.5m$, with $69\%$ collisions in random 4-agent scenarios.
	}
	\scriptsize
	\begin{tabular}{|c||c|c|c||c|c|c|}
	\hline
	& \multicolumn{6}{c|}{Test Case Setup} \\ \hline
	size (m) & & 8 x 8 & 8 x 8 & & 8 x 8 & 8 x 8 \\
	\# agents & & 2 & 4 & & 2 & 4 \\
	\hline
	& \multicolumn{6}{c|}{\textbf{Extra time to goal $\bar{t}^{e}_{g}$ (s) (Avg / 75th / 90th percentile)} $\Rightarrow$ smaller is better} \\ \hline
	ORCA & \multirow{4}{*}{$r=0.2m$}
	& 0.18 / 0.39 / 0.82 & 0.4 / 0.62 / 2.4 & \multirow{4}{*}{$r=0.5m$} & 0.43 / 1.11 / 1.65 & 0.95 / 1.22 / 1.86 \\
	SA-CADRL & & 0.2 / 0.29 / 0.43 & \textbf{0.26 / 0.38 / 0.75} & & \textbf{0.27 / 0.37 / 0.66} & \textbf{0.48 / 1.03 / 1.71} \\
	DRLMACA & & 0.91 / 1.33 / 4.82 & 1.46 / 2.16 / 3.35 & & 0.72 / 1.12 / 1.33 & 1.26 / 2.42 / 2.57 \\
	GA3C-CADRL-10-LSTM & & 0.17 / 0.24 / 0.71 & \textbf{0.25 / 0.53 / 0.7} & & \textbf{0.27 / 0.37 / 0.57} & \textbf{0.6 / 1.24 / 1.69} \\
	\hline
	& \multicolumn{6}{c|}{\textbf{\% failures  (\% collisions / \% stuck)} $\Rightarrow$ smaller is better} \\ \hline
	ORCA & \multirow{4}{*}{$r=0.2m$}
	& 4 (4 / 0) & 7 (7 / 0) & \multirow{4}{*}{$r=0.5m$} & 4 (4 / 0) & 12 (9 / 3) \\
	SA-CADRL & & \textbf{0 (0 / 0)} & 2 (0 / 2) & & \textbf{0 (0 / 0)} & 2 (0 / 2)\\
	DRLMACA & & 3 (0 / 3) & 14 (8 / 6) & & 23 (23 / 0) & 71 (69 / 2)\\
	GA3C-CADRL-10-LSTM & & \textbf{0 (0 / 0)} & \textbf{0 (0 / 0)} & & \textbf{0 (0 / 0)} & \textbf{1 (1 / 0)} \\
	\hline
	\end{tabular}
	\label{tab:comparison_to_long}
\end{table*}

The ability for GA3C-CADRL to retrain in complex scenarios after convergence in simple scenarios, and yield a significant performance increase, is a key benefit of the new framework.
This result suggests there could be other types of complexities in the environment (beyond increasing $n$) that the general GA3C-CADL framework could also learn about after being initially trained on simple scenarios.

\subsubsection{Comparison to Other RL Approach}

\cref{tab:comparison_to_long} shows a comparison to another deep RL policy, DRLMACA~\citep{long2018towards}.
DRLMACA stacks the latest 3 laserscans as the observation of other agents; other algorithms in the comparisons use the exact other agent states.
DRLMACA assumes $v_{pref}=1m/s$ for all agents, so all test cases used in~\cref{tab:comparison_to_long} share this setting ($v_{pref}$ is random in~\cref{fig:multi_agent_stats}, explaining the omission of DRLMACA).

During training, all DRLMACA agents are discs of the same radius, $R$, and some reported trajectories from~\citep{long2018towards} suggest the policy can generalize to other agent sizes.
However, our experiments with a trained DRLMACA policy~\citep{drl_long_github_3rd_party} suggest the policy does not generalize to other agent radii, as the number of collisions increases with agent radius.
In particular, $69\%$ of experiments ended in a collision for 4 agents running DRLMACA with $r=0.5m$.
Moreover, a qualitative look at DRLMACA trajectories in~\cref{fig:drlmaca_ga3c_trajs} demonstrates how agents often slow down and stop to wait for other agents, whereas GA3C-CADRL agents often move out of other agents' paths before needing to stop.
Even though the implementation in~\citep{drl_long_github_3rd_party} uses the same hyperparameters, training scheme, network architecture, and reward function from the paper, these results are worse than what was reported in~\citep{long2018towards}.

\subsubsection{Formation Control}

\begin{figure*}[t]
	\captionsetup[subfigure]{labelformat=empty, justification=centering}
	\centering
	\begin{subfigure}{0.33\textwidth}
		\centering
		\includegraphics [trim=0 0 4000 0, clip, width=\textwidth, angle = 0]{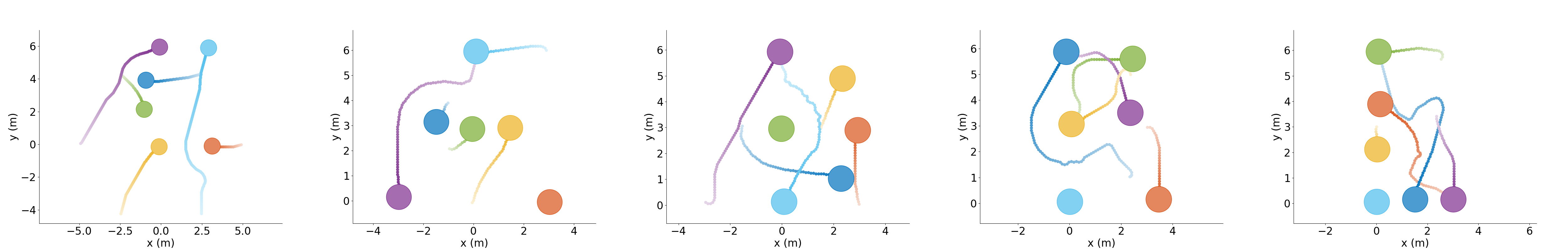}
		\caption{``C''}
		\label{fig:6_agents_spelling_cadrl_c} 
	\end{subfigure}
	\begin{subfigure}{0.33\textwidth}
		\centering
		\includegraphics [trim=1000 0 3000 0, clip, width=\textwidth, angle = 0]{figures/ijrr/sim_formation_trajs/GA3C_CADRL_6agents}
		\caption{``A''}
		\label{fig:6_agents_spelling_cadrl_a}
	\end{subfigure}
	\begin{subfigure}{0.33\textwidth}
		\centering
		\includegraphics [trim=2000 0 2000 0, clip, width=\textwidth, angle = 0]{figures/ijrr/sim_formation_trajs/GA3C_CADRL_6agents}
		\caption{``D''}
		\label{fig:6_agents_spelling_cadrl_d} 
	\end{subfigure}
	\begin{subfigure}{0.33\textwidth}
		\centering
		\includegraphics [trim=3000 0 1000 0, clip, width=\textwidth, angle = 0]{figures/ijrr/sim_formation_trajs/GA3C_CADRL_6agents}
		\caption{``R''}
		\label{fig:6_agents_spelling_cadrl_r}
	\end{subfigure}
	\begin{subfigure}{0.33\textwidth}
		\centering
		\includegraphics [trim=4000 0 0 0, clip, width=\textwidth, angle = 0]{figures/ijrr/sim_formation_trajs/GA3C_CADRL_6agents}
		\caption{``L''}
		\label{fig:6_agents_spelling_cadrl_l} 
	\end{subfigure}
	\caption{
	6 agents spelling out ``CADRL''.
	Each agent is running the same GA3C-CADRL-10-LSTM policy.
	A centralized system randomly assigns agents to goal positions (random to ensure interaction), and each agent selects its action in a decentralized manner, using knowledge of other agents' current positions, velocities, and radii.
	Collision avoidance is an essential aspect of formation control.
	}
	\label{fig:6_agents_spelling_cadrl}
\end{figure*}

Formation control is one application of multiagent robotics that requires collision avoidance: recent examples include drone light shows~\citep{intel_drones}, commercial airplane formations~\citep{airbus_formation}, robotic soccer~\citep{kitano1997robocup}, and animations~\citep{finding_nemo_formation}.
One possible formation objective is to assign a team of agents to goal coordinates, say to spell out letters or make a shape.

\cref{fig:6_agents_spelling_cadrl} shows 6 agents spelling out the letters in ``CADRL''.
Each agent uses GA3C-CADRL-10-LSTM and knowledge of other agents' current positions, velocities, and radii, to choose a collision-free action toward its own goal position.
All goal coordinates lie within a $6\times6m$ region, and goal coordinates are randomly assigned to agents.
Each agent has a radius of $0.5m$ and a preferred speed of $1.0m/s$.
Agents start in random positions before the first letter, ``C'', then move from ``C'' to ``A'', etc.
Agent trajectories darken as time increases, and the circles show the final agent positions.
Multiple iterations are animated in the video attachment.

\subsection{Hardware Experiments}

\begin{figure}[t]
	\centering
	\includegraphics [trim=0 120 220 0, clip, angle=0, width=0.8\columnwidth, keepaspectratio]{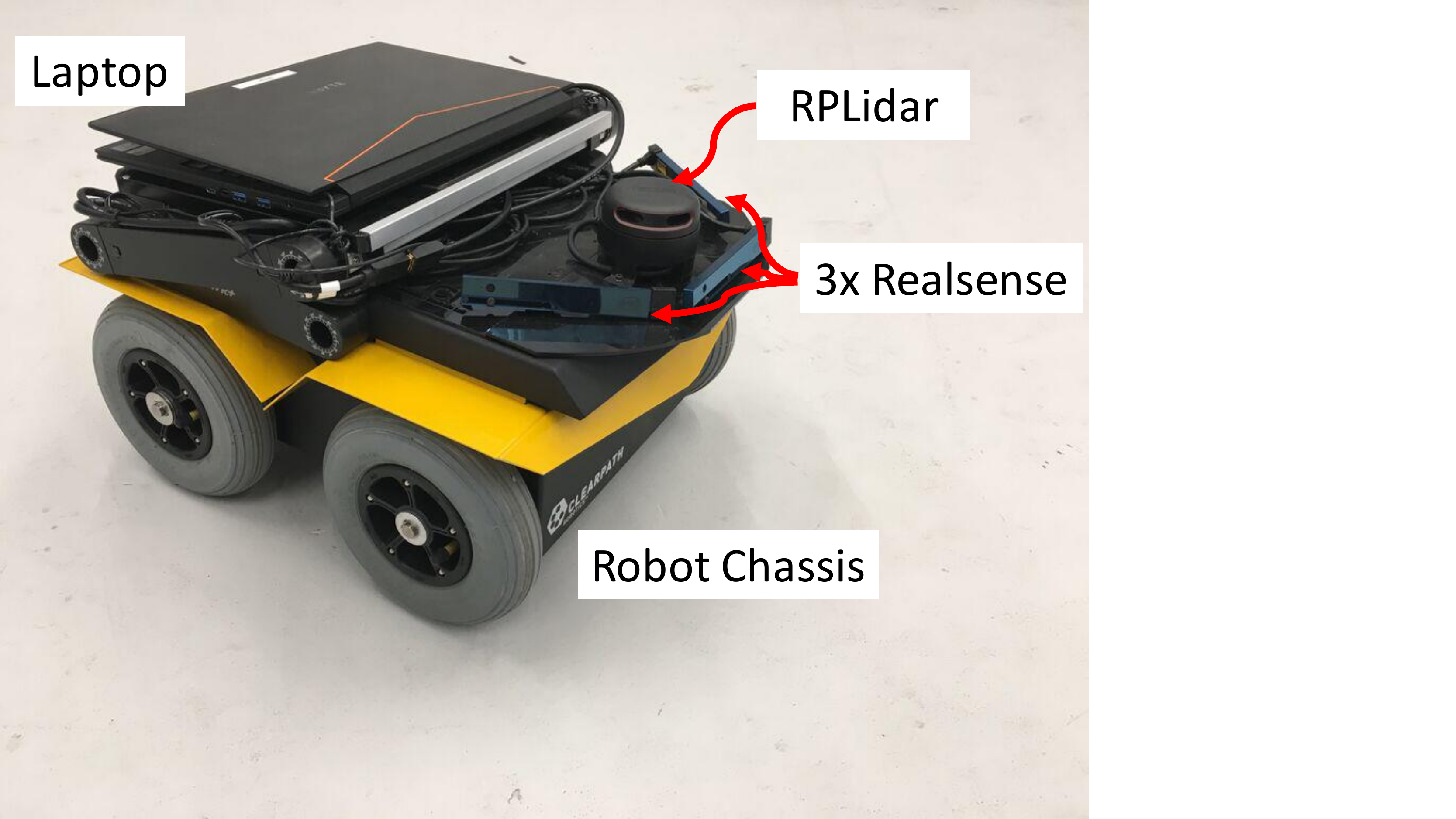}
	\caption{Robot hardware. The compact, low-cost ($<~\$1000$) sensing package uses a single 2D Lidar and 3 Intel Realsense R200 cameras. The total sensor and computation assembly is less than 3 inches tall, leaving room for cargo.}
	\label{fig:robot_sensors} 
\end{figure}

\begin{figure*}[t!]
	\centering
	\begin{subfigure}{0.3\textwidth}
		\centering
		\includegraphics [trim=0 0 3840 0, clip, width=\textwidth, angle = 0]{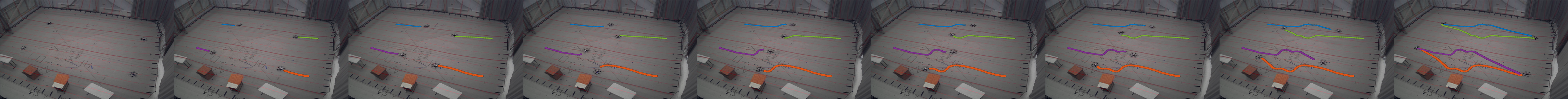}
		\caption{}
		\label{fig:4_hexes_parallel_0} 
	\end{subfigure}
	\begin{subfigure}{0.3\textwidth}
		\centering
		\includegraphics [trim=480 0 3360 0, clip, width=\textwidth, angle = 0]{figures/ijrr/drone_trajs/four_hexes_traj}
		\caption{}
		\label{fig:4_hexes_parallel_1}
	\end{subfigure}
	\begin{subfigure}{0.3\textwidth}
		\centering
		\includegraphics [trim=960 0 2880 0, clip, width=\textwidth, angle = 0]{figures/ijrr/drone_trajs/four_hexes_traj}
		\caption{}
		\label{fig:4_hexes_parallel_2} 
	\end{subfigure}
	\begin{subfigure}{0.3\textwidth}
		\centering
		\includegraphics [trim=1440 0 2400 0, clip, width=\textwidth, angle = 0]{figures/ijrr/drone_trajs/four_hexes_traj}
		\caption{}
		\label{fig:4_hexes_parallel_3} 
	\end{subfigure}
	\begin{subfigure}{0.3\textwidth}
		\centering
		\includegraphics [trim=1920 0 1920 0, clip, width=\textwidth, angle = 0]{figures/ijrr/drone_trajs/four_hexes_traj}
		\caption{}
		\label{fig:4_hexes_parallel_4} 
	\end{subfigure}
	\begin{subfigure}{0.3\textwidth}
		\centering
		\includegraphics [trim=2400 0 1440 0, clip, width=\textwidth, angle = 0]{figures/ijrr/drone_trajs/four_hexes_traj}
		\caption{}
		\label{fig:4_hexes_parallel_5} 
	\end{subfigure}
	\begin{subfigure}{0.3\textwidth}
		\centering
		\includegraphics [trim=2880 0 960 0, clip, width=\textwidth, angle = 0]{figures/ijrr/drone_trajs/four_hexes_traj}
		\caption{}
		\label{fig:4_hexes_parallel_6} 
	\end{subfigure}
	\begin{subfigure}{0.3\textwidth}
		\centering
		\includegraphics [trim=3360 0 480 0, clip, width=\textwidth, angle = 0]{figures/ijrr/drone_trajs/four_hexes_traj}
		\caption{}
		\label{fig:4_hexes_parallel_7} 
	\end{subfigure}
	\begin{subfigure}{0.3\textwidth}
		\centering
		\includegraphics [trim=3840 0 0 0, clip, width=\textwidth, angle = 0]{figures/ijrr/drone_trajs/four_hexes_traj}
		\caption{}
		\label{fig:4_hexes_parallel_8} 
	\end{subfigure}
	\caption{4 Multirotors running GA3C-CADRL: 2 parallel pairs. Each vehicle's on-board controller tracks the desired speed and heading angle produced by each vehicle's GA3C-CADRL-10-LSTM policy.}
	\label{fig:4_hexes_parallel_swaps}

	\begin{subfigure}{0.3\textwidth}
		\centering
		\includegraphics [trim=0 0 3840 0, clip, width=\textwidth, angle = 0]{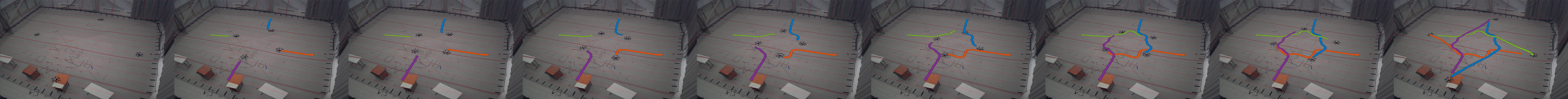}
		\caption{}
		\label{fig:4_hexes_square_0} 
	\end{subfigure}
	\begin{subfigure}{0.3\textwidth}
		\centering
		\includegraphics [trim=480 0 3360 0, clip, width=\textwidth, angle = 0]{figures/ijrr/drone_trajs/four_hexes_square_traj}
		\caption{}
		\label{fig:4_hexes_square_1}
	\end{subfigure}
	\begin{subfigure}{0.3\textwidth}
		\centering
		\includegraphics [trim=960 0 2880 0, clip, width=\textwidth, angle = 0]{figures/ijrr/drone_trajs/four_hexes_square_traj}
		\caption{}
		\label{fig:4_hexes_square_2} 
	\end{subfigure}
	\begin{subfigure}{0.3\textwidth}
		\centering
		\includegraphics [trim=1440 0 2400 0, clip, width=\textwidth, angle = 0]{figures/ijrr/drone_trajs/four_hexes_square_traj}
		\caption{}
		\label{fig:4_hexes_square_3} 
	\end{subfigure}
	\begin{subfigure}{0.3\textwidth}
		\centering
		\includegraphics [trim=1920 0 1920 0, clip, width=\textwidth, angle = 0]{figures/ijrr/drone_trajs/four_hexes_square_traj}
		\caption{}
		\label{fig:4_hexes_square_4} 
	\end{subfigure}
	\begin{subfigure}{0.3\textwidth}
		\centering
		\includegraphics [trim=2400 0 1440 0, clip, width=\textwidth, angle = 0]{figures/ijrr/drone_trajs/four_hexes_square_traj}
		\caption{}
		\label{fig:4_hexes_square_5} 
	\end{subfigure}
	\begin{subfigure}{0.3\textwidth}
		\centering
		\includegraphics [trim=2880 0 960 0, clip, width=\textwidth, angle = 0]{figures/ijrr/drone_trajs/four_hexes_square_traj}
		\caption{}
		\label{fig:4_hexes_square_6} 
	\end{subfigure}
	\begin{subfigure}{0.3\textwidth}
		\centering
		\includegraphics [trim=3360 0 480 0, clip, width=\textwidth, angle = 0]{figures/ijrr/drone_trajs/four_hexes_square_traj}
		\caption{}
		\label{fig:4_hexes_square_7} 
	\end{subfigure}
	\begin{subfigure}{0.3\textwidth}
		\centering
		\includegraphics [trim=3840 0 0 0, clip, width=\textwidth, angle = 0]{figures/ijrr/drone_trajs/four_hexes_square_traj}
		\caption{}
		\label{fig:4_hexes_square_8} 
	\end{subfigure}
	\caption{4 Multirotors running GA3C-CADRL: 2 orthogonal pairs. The agents form a symmetric ``roundabout'' pattern in the center of the room, even though each vehicle has slightly different dynamics and observations of neighbors.}
	\label{fig:4_hexes_square}
\end{figure*}

\begin{figure*}[t]
	\centering
	\begin{subfigure}{0.33\textwidth}
		\centering
		\includegraphics [trim=0 200 5440 0, clip, width=\textwidth, angle = 0]{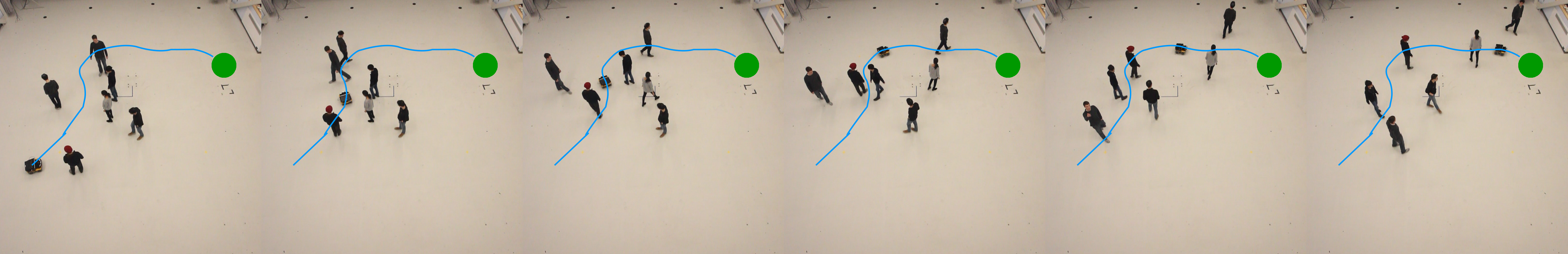}
		\caption{}
		\label{fig:robot_pedestrian_trajs0} 
	\end{subfigure}
	\begin{subfigure}{0.33\textwidth}
		\centering
		\includegraphics [trim=1088 200 4352 0, clip, width=\textwidth, angle = 0]{figures/ijrr/robot_trajs/robot_traj}
		\caption{}
		\label{fig:robot_pedestrian_trajs1}
	\end{subfigure}
	\begin{subfigure}{0.33\textwidth}
		\centering
		\includegraphics [trim=2176 200 3264 0, clip, width=\textwidth, angle = 0]{figures/ijrr/robot_trajs/robot_traj}
		\caption{}
		\label{fig:robot_pedestrian_trajs2} 
	\end{subfigure}
	\begin{subfigure}{0.33\textwidth}
		\centering
		\includegraphics [trim=3264 200 2176 0, clip, width=\textwidth, angle = 0]{figures/ijrr/robot_trajs/robot_traj}
		\caption{}
		\label{fig:robot_pedestrian_trajs3} 
	\end{subfigure}
	\begin{subfigure}{0.33\textwidth}
		\centering
		\includegraphics [trim=4352 200 1088 0, clip, width=\textwidth, angle = 0]{figures/ijrr/robot_trajs/robot_traj}
		\caption{}
		\label{fig:robot_pedestrian_trajs4} 
	\end{subfigure}
	\begin{subfigure}{0.33\textwidth}
		\centering
		\includegraphics [trim=5440 200 0 0, clip, width=\textwidth, angle = 0]{figures/ijrr/robot_trajs/robot_traj}
		\caption{}
		\label{fig:robot_pedestrian_trajs5} 
	\end{subfigure}
	\caption{Ground robot among pedestrians. The on-board sensors are used to estimate pedestrian positions, velocities, and radii. An on-board controller tracks the GA3C-CADRL-10-LSTM policy output. The vehicle moves at human walking speed (1.2m/s), nominally.}
	\label{fig:robot_pedestrian_trajs}
\end{figure*}

This work implements the policy learned in simulation on two different hardware platforms to demonstrate the flexibility in the learned collision avoidance behavior and that the learned policy enables real-time decision-making.
The first platform, a fleet of 4 multirotors, highlights the transfer of the learned policy to vehicles with more complicated dynamics than the unicycle kinematic model used in training.
The second platform, a ground robot operating among pedestrians, highlights the policy's robustness to both imperfect perception from low-cost, on-board perception, and to heterogeneity in other agent policies, as none of the pedestrians follow one of the policies seen in training.

\changed{The hardware experiments were designed to demonstrate that the new algorithm could be deployed using realistic sensors, vehicle dynamics, and computational resources.
Combined with the numerical experiments in~\cref{fig:multi_agent_stats,tab:comparison_to_long}, the hardware experiments provide evidence of an algorithm that exceeds the state of the art and can be deployed on real robots.}

\subsubsection{Multiple Multirotors}
A fleet of 4 multirotors with on-board velocity and position controllers resemble the agents in the simulated training environment.
\added{These experiments consider the case of multirotors flying within the same plane (roughly 1m above the ground).}
Each vehicle's planner receives state measurements \added{of the ego vehicle (position, velocity, heading angle)} and of the other vehicles (positions, velocities) from a motion capture system at 200Hz~\citep{Omidshafiei15_Infotech}.
At each planning step (10Hz), the planners build a state vector using other agent states, an assumed agent radius (0.5m), a preferred ego speed (0.5m/s), and knowledge of their own goal position in global coordinates.
Each vehicle's planner queries the learned policy and selects the action with highest probability: a desired heading angle change and speed.
A desired velocity vector with magnitude equal to the desired speed, and in the direction of the desired heading angle, is sent to the velocity controller.
To smooth the near-goal behavior, the speed and heading rates decay linearly with distance to goal within 2m, and agents simply execute position control on the goal position when within 0.3m of the goal.
Throughout flight, the multirotors also control to their desired heading angle; this would be necessary with a forward-facing sensor, but is somewhat extraneous given that other agents' state estimates are provided externally.

The experiments included two challenging 4-agent scenarios.
In~\cref{fig:4_hexes_parallel_swaps}, two pairs of multirotors swap positions in parallel, much like the third column of~\cref{fig:234agent_traj}.
This policy was not trained to prefer a particular directionality -- the agents demonstrate clockwise/left-handed collision avoidance behavior in the center of the room.
In~\cref{fig:4_hexes_square}, the 4 vehicles swap positions passing through a common center, like the fourth column of~\cref{fig:234agent_traj}.
Unlike in simulation, where the agents' dynamics, observations, and policies (and therefore, actions) are identical, small variations in vehicle states lead to slightly different actions for each agent.
However, even with these small fluctuations, the vehicles still perform ``roundabout'' behavior in the center.

Further examples of 2-agent swaps, and multiple repeated trials of the 4-agent scenarios are included in the video attachment.

\subsubsection{Ground Robot Among Pedestrians}

A GA3C-CADRL policy implemented on a ground robot demonstrates the algorithm's performance among pedestrians.
We designed a compact, low-cost (${<\$1000}$) sensing suite with sensors placed as to not limit the robot's cargo-carrying capability~(\cref{fig:robot_sensors}).
The sensors are a 2D Lidar (used for localization and obstacle detection), and 3 Intel Realsense R200 cameras (used for pedestrian classification and obstacle detection).
Pedestrian positions and velocities are estimated by clustering the 2D Lidar's scan~\citep{Campbell13_NIPS}, and clusters are labeled as pedestrians using a classifier~\citep{liu2016ssd} applied to the cameras' RGB images~\citep{miller_dynamic_2016}.
A detailed description of the software architecture is in~\citep{everett_robot_2017}.
\added{Despite not having perfect state knowledge, as was available in the simulations, the robot is able to avoid collisions using only on-board sensing.}

\begin{figure*}[t]
    \includegraphics[width=\textwidth]{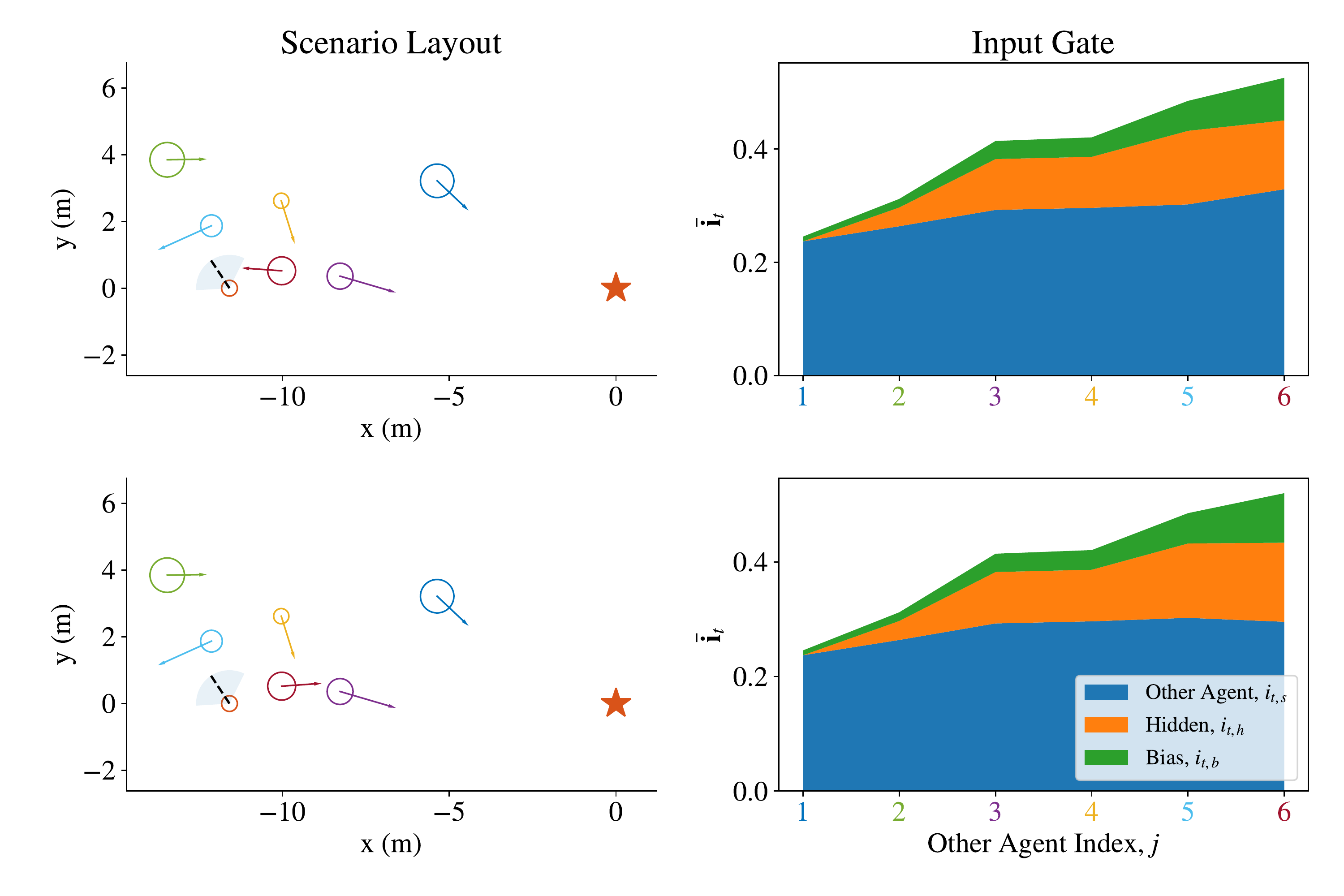}
    \caption{
    Gate Dynamics on Single Timestep.
    In the top row, one agent, near (-12,0), observes 6 neighboring agents.
    Other agent states are passed into the LSTM sequentially, starting with the furthest agent, near (-5, 3).
    The top right plot shows the impact of each LSTM cell input on the input gate as each agent is added to the LSTM: other agent state (bottom slice), previous hidden state (middle slice), and bias (top slice).
    The bottom row shows the same scenario, but the closest agent, near (-10,0), has a velocity vector away from the ego agent.
    Accordingly, the bottom right plot's bottom slice slightly declines from $j=5$ to $j=6$, but the corresponding slice increases in the top right plot.
    This suggests the LSTM has learned to put more emphasis on agents heading toward the ego agent, as they are more relevant for collision avoidance.
    }
    \label{fig:lstm_gate_dynamics}
\end{figure*}

Snapshots of a particular sequence are shown in~\cref{fig:robot_pedestrian_trajs}: 6 pedestrians move around between the robot's starting position and its goal (large circle) about 6m away.
Between the first two frames, 3 of the pedestrians remain stationary, and the other 3 move with varying levels of cooperativeness, but these roles were not assigned and change stochastically throughout the scenario.
The robot successfully navigates to its goal in the proximity of many heterogeneous agents.
Other examples of safe robot navigation in challenging scenarios are available in the video attachment.

\subsection{LSTM Analysis}\label{sec:results:lstm_analysis}

This section provides insights into the design and inner workings of the LSTM module in~\cref{fig:nn_arch} in two ways: how agent states affect the LSTM gates, and how the ordering of agents affects performance during training.

\subsubsection{LSTM Gate Dynamics}

We first analyze the LSTM of the trained GA3C-CADRL-10-LSTM network, building on~\citep{omidshafiei2017crossmodal}, using notation from~\citep{Olah15}.
The LSTM weights, $\{W_i, W_f, W_o\}$, and biases, $\{b_i, b_f, b_o\}$ are updated during the training process and fixed during inference.

Recall the LSTM has three inputs: the state of agent $j$ at time $t$, the previous hidden state, and the previous cell state, which are denoted $\tilde{\mathbf{s}}^o_{j,t}$, $\mathbf{h}_{t-1}$, $C_{t-1}$, respectively.
Of the four gates in an LSTM, we focus on the input gate here.
The input gate, $\mathbf{i}_t \in [0,1]^{n_h}$, is computed as, 
\begin{equation}
\mathbf{i}_{t} = \sigma([W_{i,s}, \, W_{i,h}, \, W_{i,b}]^T\cdot[\tilde{\mathbf{s}}^o_{j,t}, \, \mathbf{h}_t, \, b_i]),
\end{equation}
where $W_i = [W_{i,h}, \, W_{i,s}]$, \, $W_{i,b}=\textrm{diag}(b_i)$, and $n_h=64$ is the hidden state size.

Thus, $\mathbf{i}_t = [1]^{n_h}$ when the entire \textit{new} candidate cell state, $\tilde{C}_t$, should be used to update the cell state, $C_t$,
\begin{align}
\tilde{C}_t &= \textrm{tanh}([W_{C,s}, W_{C,h}, W_{C,b}]^T \cdot [\tilde{\mathbf{s}}^o_{j,t}, \mathbf{h}_{t-1}, b_t]) \\
C_t &= \mathbf{f}_t * C_{t-1} + \mathbf{i}_t * \tilde{C}_t,
\end{align}
where $\mathbf{f}_t$ is the value of the forget gate, computed analgously to $\mathbf{i}_t$.
In other words, $\mathbf{i}_t$ with elements near 1 means that agent $j$ is particularly important in the context of agents ${[1,\ldots,j-1]}$, and it will have a large impact on $C_t$.
Contrarily, $\mathbf{i}_t$ with elements near 0 means very little of the observation about agent $j$ will be added to the hidden state and will have little impact on the downstream decision-making.

\allowdisplaybreaks
Because $\mathbf{i}_t$ is a 64-element vector, we must make some manipulations to visualize it.
First, we separate $\mathbf{i}_t$ into quantities that measure how much it is affected by each component, $\{\tilde{\mathbf{s}}^o_{j,t}, \mathbf{h}_{t-1}, b_i\}$:
\begin{align}
\tilde{i}_{t,s} &= ||\mathbf{i}_{t} - \sigma([W_{i,h}, W_{i,b}]^T\cdot[\mathbf{h}_t, b_i])||_2\\
\tilde{i}_{t,h} &= ||\mathbf{i}_{t} - \sigma([W_{i,s}, W_{i,b}]^T\cdot[\tilde{\mathbf{s}}^o_{j,t}, b_i])||_2\\
\tilde{i}_{t,b} &= ||\mathbf{i}_{t} - \sigma([W_{i,s}, W_{i,h}]^T\cdot[\tilde{\mathbf{s}}^o_{j,t}, \mathbf{h}_t])||_2\\
\bar{\mathbf{i}}_t &= \begin{bmatrix} i_{t,s} \\ i_{t,h} \\ i_{t,b} \\ \end{bmatrix} = k \cdot \begin{bmatrix} \tilde{i}_{t,s} \\ \tilde{i}_{t,h} \\ \tilde{i}_{t,b} \\ \end{bmatrix} \\
k &= \frac{||\mathbf{i}_{t}||_1}{\tilde{i}_{t,s}+\tilde{i}_{t,h}+\tilde{i}_{t,b}},
\end{align}
where the constant, $k$, normalizes the sum of the three components contributing to $\bar{\mathbf{i}}_t$, and scales each by the average of all elements in $\mathbf{i}_t$.

An example scenario is shown in~\cref{fig:lstm_gate_dynamics}.
A randomly generated 7-agent scenario is depicted on the left column, where the ego agent is at $(-12, \,0)$, and its goal is the star at $(0,\,0)$.
The 6 other agents in the neighborhood are added to the LSTM in order of furthest distance to the ego agent, so the tick marks on the x-axis of the right-hand figures correspond to each neighboring agent.
That is, the agent at $(-5,\,3)$ is furthest from the ego agent, so it is agent $j=0$, and the agent at $(-10,\,0)$ is closest and is agent $j=5$.

For this scenario, $\bar{\mathbf{i}}_t$ (top of the stack of three slices) starts about 0.3, and goes up and down (though trending slightly upward) as agents are added.
The bottom slice corresponds to $i_{t,s}$, middle to $i_{t,h}$, and top to $i_{t,b}$.

The top and middle slices are tiny compared to the bottom slice for agent 0.
This corresponds to the fact that, for $j=0$, all information about whether that agent is relevant for decision-making is in $\tilde{\mathbf{s}}^o_{0,0}$ (bottom), since the hidden and cell states are initially blank ($h_{-1}=\mathbf{0}$).
As more agents are added, the LSTM considers \textit{both} the hidden state and current observation to decide how much of the candidate cell state to pass through -- this intuition matches up with the relatively larger middle slices for subsequent agents.

The importance of the contents of $\tilde{\mathbf{s}}^o_{j,t}$ is demonstrated in the bottom row of~\cref{fig:lstm_gate_dynamics}.
It considers the same scenario as the top row, but with the closest agent's velocity vector pointing away from the ego agent, instead of toward.
The values of $\bar{\mathbf{i}}_t$ for all previous agents are unchanged, but the value of $\bar{\mathbf{i}}_t$ is larger when the agent is heading toward the ego agent.
This is seen as an uptick between $j=5$ and $j=6$ in the bottom slice of the top-right figure, and a flat/slightly decreasing segment for the corresponding piece of the bottom-right figure.
This observation agrees with the intuition that agents heading toward the ego agent should have a larger impact on the hidden state, and eventually on collision avoidance decision-making.

This same behavior of increased $\bar{\mathbf{i}}_t$ (specifically $i_{t_s}$) when the last agent was heading roughly toward the ego agent was observed in most randomly generated scenarios.
Our software release will include an interactive Jupyter notebook so researchers can analyze other scenarios of interest, or do similar analysis on their networks.

\subsubsection{Agent Ordering Strategies}\label{sec:results:lstm_ordering}

\begin{figure}[t]
	\vspace{-0.1in}
    \includegraphics[trim=0 0 0 20, clip, width=\columnwidth]{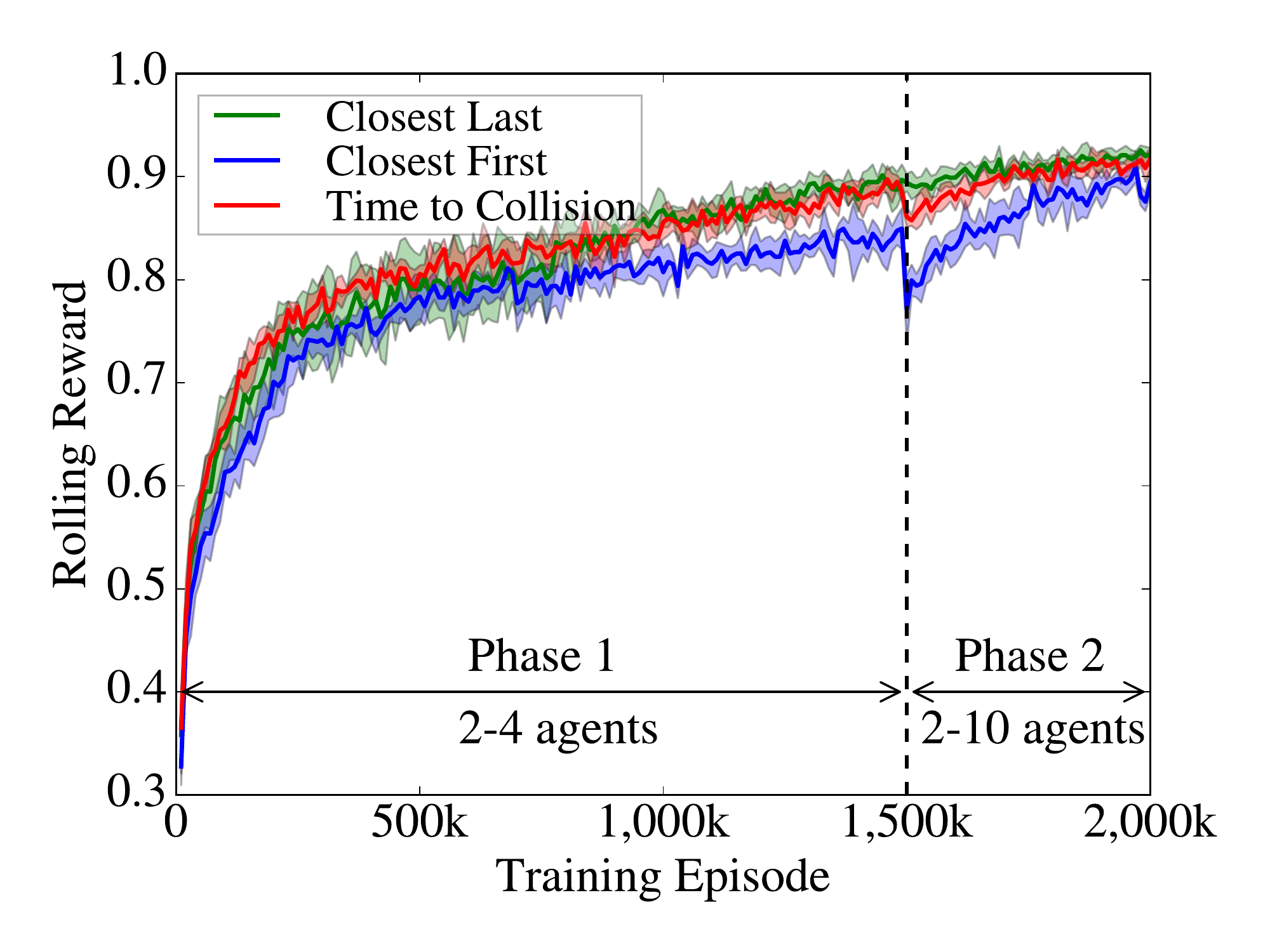}
    \caption{
    Training performance and LSTM ordering effect on training.
    The first phase of training uses random scenarios with 2-4 agents; the final 500k episodes use random scenarios with 2-10 agents.
    Three curves corresponding to three heuristics for ordering the agent sequences all converge to a similar reward after 2M episodes.
    The ``closest last'' ordering has almost no dropoff between phases and achieves the highest final performance.
    The ``closest first'' ordering drops off substantially between phases, suggesting the ordering scheme has a second-order effect on training.
    Curves show the mean $\pm 1\sigma$ over 5 training runs per ordering.
    }
    \label{fig:training_curve}
\end{figure}

The preceding discussion on LSTM gate dynamics assumed agents are fed into the LSTM in the order of ``closest last.''
However, there are many ways of ordering the agents.

\cref{fig:training_curve} compares the performance throughout the training process of networks with three different ordering strategies.
``Closest last'' is sorted in decreasing order of agent distance to the ego agent, and ``closest first'' is the reverse of that.
``Time to collision'' is computed as the minimum time at the current velocities for two agents to collide and is often infinite when agents are not heading toward one another.
The secondary ordering criterion of ``closest last'' was used as a tiebreaker.
In all cases, $\tilde{p}_x$ (in the ego frame) was used as a third tiebreaker to preserve symmetry.

The same network architecture, differing only in the LSTM agent ordering, was trained for 1.5M episodes in 2-4 agent scenarios (Phase 1) and 0.5M more episodes in 2-10 agent scenarios (Phase 2).
All three strategies yield similar performance over the first 1M training episodes.
By the end of phase 1, the ``closest first'' strategy performs slightly worse than the other two, which are roughly overlapping.

At the change in training phase, the ``closest first'' performance drops substantially, and the ``time to collision'' curve has a small dip.
This suggests that the first training phase did not produce an LSTM that efficiently combines previous agent summaries with an additional agent for these two heuristics.
On the other hand, there is no noticeable dip with the ``closest last'' strategy.
All three strategies converge to a similar final performance.

In conclusion, the choice of ordering has a second order effect on the reward curve during training, and the ``closest last'' strategy employed throughout this work was better than the tested alternatives.
This evidence aligns with the intuition from~\cref{sec:approach:variable_num_agents}.

\section{Conclusion} \label{sec:conclusion}
This work presented a collision avoidance algorithm, GA3C-CADRL, that is trained in simulation with deep RL without requiring any knowledge of other agents' dynamics.
It also proposed a strategy to enable the algorithm to select actions based on observations of a large (possibly varying) number of nearby agents, using LSTM at the network's input.
The new approach was shown to outperform a classical method, another deep RL-based method, and scales better than our previous deep RL-based method as the number of agents in the environment increased.
These results support the use of LSTMs to encode a varying number of agent states into a fixed-length representation of the world.
Analysis of the trained LSTM provides deeper introspection into the effect of agent observations on the hidden state vector, and quantifies the effect of agent ordering heuristics on performance throughout training.
The work provided an application of the algorithm for formation control, and the algorithm was implemented on two hardware platforms: a fleet of 4 fully autonomous multirotors successfully avoided collisions across multiple scenarios, and a small ground robot was shown to navigate at human walking speed among pedestrians.
\changed{
Combined with the numerical comparisons to prior works, the hardware experiments provide evidence of an algorithm that exceeds the state of the art and can be deployed on real robots.
}

\bibliographystyle{IEEEtran}
\bibliography{biblio}

\begin{thebibliography}{10}
\providecommand{\url}[1]{#1}
\csname url@samestyle\endcsname
\providecommand{\newblock}{\relax}
\providecommand{\bibinfo}[2]{#2}
\providecommand{\BIBentrySTDinterwordspacing}{\spaceskip=0pt\relax}
\providecommand{\BIBentryALTinterwordstretchfactor}{4}
\providecommand{\BIBentryALTinterwordspacing}{\spaceskip=\fontdimen2\font plus
\BIBentryALTinterwordstretchfactor\fontdimen3\font minus
  \fontdimen4\font\relax}
\providecommand{\BIBforeignlanguage}[2]{{%
\expandafter\ifx\csname l@#1\endcsname\relax
\typeout{** WARNING: IEEEtran.bst: No hyphenation pattern has been}%
\typeout{** loaded for the language `#1'. Using the pattern for}%
\typeout{** the default language instead.}%
\else
\language=\csname l@#1\endcsname
\fi
#2}}
\providecommand{\BIBdecl}{\relax}
\BIBdecl

\bibitem{kummerle2013navigation}
R.~K{\"u}mmerle, M.~Ruhnke, B.~Steder, C.~Stachniss, and W.~Burgard, ``A
  navigation system for robots operating in crowded urban environments,'' in
  \emph{2013 IEEE International Conference on Robotics and Automation}.\hskip
  1em plus 0.5em minus 0.4em\relax IEEE, 2013, pp. 3225--3232.

\bibitem{trautman_unfreezing_2010}
P.~Trautman and A.~Krause, ``Unfreezing the robot: {{Navigation}} in dense,
  interacting crowds,'' in \emph{2010 {{IEEE}}/{{RSJ International Conference}}
  on {{Intelligent Robots}} and {{Systems}} ({{IROS}})}, Oct. 2010, pp.
  797--803.

\bibitem{snape_hybrid_2011}
J.~Snape, J.~{Van den Berg}, S.~J. Guy, and D.~Manocha, ``The hybrid reciprocal
  velocity obstacle,'' \emph{IEEE Transactions on Robotics}, vol.~27, no.~4,
  pp. 696--706, Aug. 2011.

\bibitem{ferrer_social-aware_2013}
G.~Ferrer, A.~Garrell, and A.~Sanfeliu, ``Social-aware robot navigation in
  urban environments,'' in \emph{2013 {{European Conference}} on {{Mobile
  Robots}} ({{ECMR}})}, Sep. 2013, pp. 331--336.

\bibitem{berg_reciprocal_2011}
J.~{Van den Berg}, S.~J. Guy, M.~Lin, and D.~Manocha,
  ``\BIBforeignlanguage{en}{Reciprocal n-body collision avoidance},'' in
  \emph{\BIBforeignlanguage{en}{Robotics {{Research}}}}, ser. Springer Tracts
  in Advanced Robotics.\hskip 1em plus 0.5em minus 0.4em\relax {Springer Berlin
  Heidelberg}, 2011, no.~70, pp. 3--19.

\bibitem{alonso2013optimal}
J.~Alonso-Mora, A.~Breitenmoser, M.~Rufli, P.~Beardsley, and R.~Siegwart,
  ``Optimal reciprocal collision avoidance for multiple non-holonomic robots,''
  in \emph{Distributed Autonomous Robotic Systems}.\hskip 1em plus 0.5em minus
  0.4em\relax Springer, 2013, pp. 203--216.

\bibitem{kretzschmar_socially_2016}
H.~Kretzschmar, M.~Spies, C.~Sprunk, and W.~Burgard,
  ``\BIBforeignlanguage{en}{Socially compliant mobile robot navigation via
  inverse reinforcement learning},'' \emph{\BIBforeignlanguage{en}{The
  International Journal of Robotics Research}}, Jan. 2016.

\bibitem{trautman_robot_2013}
P.~Trautman, J.~Ma, R.~M. Murray, and A.~Krause, ``Robot navigation in dense
  human crowds: the case for cooperation,'' in \emph{Proceedings of the 2013
  {{IEEE International Conference}} on {{Robotics}} and {{Automation}}
  ({{ICRA}})}, May 2013, pp. 2153--2160.

\bibitem{kuderer_feature-based_2012}
M.~Kuderer, H.~Kretzschmar, C.~Sprunk, and W.~Burgard, ``Feature-based
  prediction of trajectories for socially compliant navigation,'' in
  \emph{Robotics:{{Science}} and {{Systems}}}, 2012.

\bibitem{long2018towards}
P.~Long, T.~Fanl, X.~Liao, W.~Liu, H.~Zhang, and J.~Pan, ``Towards optimally
  decentralized multi-robot collision avoidance via deep reinforcement
  learning,'' in \emph{2018 IEEE International Conference on Robotics and
  Automation (ICRA)}.\hskip 1em plus 0.5em minus 0.4em\relax IEEE, 2018, pp.
  6252--6259.

\bibitem{tai2017virtual}
L.~Tai, G.~Paolo, and M.~Liu, ``Virtual-to-real deep reinforcement learning:
  Continuous control of mobile robots for mapless navigation,'' in
  \emph{Intelligent Robots and Systems (IROS), 2017 IEEE/RSJ International
  Conference on}.\hskip 1em plus 0.5em minus 0.4em\relax IEEE, 2017, pp.
  31--36.

\bibitem{sutskever2014sequence}
I.~Sutskever, O.~Vinyals, and Q.~V. Le, ``Sequence to sequence learning with
  neural networks,'' in \emph{Advances in neural information processing
  systems}, 2014, pp. 3104--3112.

\bibitem{cho2014learning}
K.~Cho, B.~Van~Merri{\"e}nboer, C.~Gulcehre, D.~Bahdanau, F.~Bougares,
  H.~Schwenk, and Y.~Bengio, ``Learning phrase representations using rnn
  encoder-decoder for statistical machine translation,'' \emph{arXiv preprint
  arXiv:1406.1078}, 2014.

\bibitem{hochreiter1997long}
S.~Hochreiter and J.~Schmidhuber, ``Long short-term memory,'' \emph{Neural
  computation}, vol.~9, no.~8, pp. 1735--1780, 1997.

\bibitem{chen_decentralized_2017}
Y.~Chen, M.~Liu, M.~Everett, and J.~P. How, ``Decentralized, non-communicating
  multiagent collision avoidance with deep reinforcement learning,'' in
  \emph{Proceedings of the 2017 {{IEEE International Conference}} on
  {{Robotics}} and {{Automation}} ({{ICRA}})}, Singapore, 2017.

\bibitem{Chen17_IROS}
Y.~F. Chen, M.~Everett, M.~Liu, and J.~P. How, ``Socially aware motion planning
  with deep reinforcement learning,'' in \emph{IEEE/RSJ International
  Conference on Intelligent Robots and Systems (IROS)}, Vancouver, BC, Canada,
  September 2017.

\bibitem{Everett18_IROS}
M.~Everett, Y.~F. Chen, and J.~P. How, ``Motion planning among dynamic,
  decision-making agents with deep reinforcement learning,'' in \emph{IEEE/RSJ
  International Conference on Intelligent Robots and Systems (IROS)}, Madrid,
  Spain, September 2018.

\bibitem{rawlings2000tutorial}
J.~B. Rawlings, ``Tutorial overview of model predictive control,'' \emph{IEEE
  control systems magazine}, vol.~20, no.~3, pp. 38--52, 2000.

\bibitem{fox_dynamic_1997}
D.~Fox, W.~Burgard, and S.~Thrun, ``The dynamic window approach to collision
  avoidance,'' \emph{IEEE Robotics Automation Magazine}, vol.~4, no.~1, pp.
  23--33, Mar. 1997.

\bibitem{phillips_sipp:_2011}
M.~Phillips and M.~Likhachev, ``{{SIPP}}: {{Safe}} interval path planning for
  dynamic environments,'' in \emph{2011 {{IEEE International Conference}} on
  {{Robotics}} and {{Automation}} ({{ICRA}})}, May 2011, pp. 5628--5635.

\bibitem{aoude_probabilistically_2013}
G.~S. Aoude, B.~D. Luders, J.~M. Joseph, N.~Roy, and J.~P. How,
  ``\BIBforeignlanguage{en}{Probabilistically safe motion planning to avoid
  dynamic obstacles with uncertain motion patterns},''
  \emph{\BIBforeignlanguage{en}{Autonomous Robots}}, vol.~35, no.~1, pp.
  51--76, May 2013.

\bibitem{sutton_introduction_1998}
R.~S. Sutton and A.~G. Barto, \emph{Introduction to {{Reinforcement
  Learning}}}, 1st~ed.\hskip 1em plus 0.5em minus 0.4em\relax Cambridge, MA,
  USA: {MIT Press}, 1998.

\bibitem{mnih-dqn-2015}
V.~Mnih, K.~Kavukcuoglu, D.~Silver, A.~A. Rusu, J.~Veness, M.~G. Bellemare,
  A.~Graves, M.~Riedmiller, A.~K. Fidjeland, G.~Ostrovski, S.~Petersen,
  C.~Beattie, A.~Sadik, I.~Antonoglou, H.~King, D.~Kumaran, D.~Wierstra,
  S.~Legg, and D.~Hassabis, ``Human-level control through deep reinforcement
  learning,'' \emph{Nature}, vol. 518, no. 7540, pp. 529--533, Feb. 2015.

\bibitem{mnih2016asynchronous}
V.~Mnih, A.~P. Badia, M.~Mirza, A.~Graves, T.~Lillicrap, T.~Harley, D.~Silver,
  and K.~Kavukcuoglu, ``Asynchronous methods for deep reinforcement learning,''
  in \emph{International Conference on Machine Learning}, 2016, pp. 1928--1937.

\bibitem{babaeizadeh2017ga3c}
M.~Babaeizadeh, I.~Frosio, S.~Tyree, J.~Clemons, and J.~Kautz, ``Reinforcement
  learning thorugh asynchronous advantage actor-critic on a gpu,'' in
  \emph{ICLR}, 2017.

\bibitem{omidshafiei2017crossmodal}
S.~Omidshafiei, D.-K. Kim, J.~Pazis, and J.~P. How, ``Crossmodal attentive
  skill learner,'' \emph{arXiv preprint arXiv:1711.10314}, 2017.

\bibitem{schulman2017proximal}
J.~Schulman, F.~Wolski, P.~Dhariwal, A.~Radford, and O.~Klimov, ``Proximal
  policy optimization algorithms,'' \emph{arXiv preprint arXiv:1707.06347},
  2017.

\bibitem{fujimoto2018addressing}
S.~Fujimoto, H.~van Hoof, and D.~Meger, ``Addressing function approximation
  error in actor-critic methods,'' \emph{arXiv preprint arXiv:1802.09477},
  2018.

\bibitem{hessel2018rainbow}
M.~Hessel, J.~Modayil, H.~Van~Hasselt, T.~Schaul, G.~Ostrovski, W.~Dabney,
  D.~Horgan, B.~Piot, M.~Azar, and D.~Silver, ``Rainbow: Combining improvements
  in deep reinforcement learning,'' in \emph{Thirty-Second AAAI Conference on
  Artificial Intelligence}, 2018.

\bibitem{bojarski2016end}
M.~Bojarski, D.~Del~Testa, D.~Dworakowski, B.~Firner, B.~Flepp, P.~Goyal, L.~D.
  Jackel, M.~Monfort, U.~Muller, J.~Zhang \emph{et~al.}, ``End to end learning
  for self-driving cars,'' \emph{arXiv preprint arXiv:1604.07316}, 2016.

\bibitem{kim_socially_2015}
B.~Kim and J.~Pineau, ``\BIBforeignlanguage{en}{Socially adaptive path planning
  in human environments using inverse reinforcement learning},''
  \emph{\BIBforeignlanguage{en}{International Journal of Social Robotics}},
  vol.~8, no.~1, pp. 51--66, Jun. 2015.

\bibitem{pfeiffer2016predicting}
M.~Pfeiffer, U.~Schwesinger, H.~Sommer, E.~Galceran, and R.~Siegwart,
  ``Predicting actions to act predictably: Cooperative partial motion planning
  with maximum entropy models,'' in \emph{2016 IEEE/RSJ International
  Conference on Intelligent Robots and Systems (IROS)}.\hskip 1em plus 0.5em
  minus 0.4em\relax IEEE, 2016, pp. 2096--2101.

\bibitem{Baker2020Emergent}
\BIBentryALTinterwordspacing
B.~Baker, I.~Kanitscheider, T.~Markov, Y.~Wu, G.~Powell, B.~McGrew, and
  I.~Mordatch, ``Emergent tool use from multi-agent autocurricula,'' in
  \emph{International Conference on Learning Representations}, 2020. [Online].
  Available: \url{https://openreview.net/forum?id=SkxpxJBKwS}
\BIBentrySTDinterwordspacing

\bibitem{garcia2015comprehensive}
J.~Garc{\i}a and F.~Fern{\'a}ndez, ``A comprehensive survey on safe
  reinforcement learning,'' \emph{Journal of Machine Learning Research},
  vol.~16, no.~1, pp. 1437--1480, 2015.

\bibitem{alahi2016social}
A.~Alahi, K.~Goel, V.~Ramanathan, A.~Robicquet, L.~Fei-Fei, and S.~Savarese,
  ``Social lstm: Human trajectory prediction in crowded spaces,'' in
  \emph{Proceedings of the IEEE Conference on Computer Vision and Pattern
  Recognition}, 2016, pp. 961--971.

\bibitem{Olah15}
C.~Olah, ``Understanding lstm networks,'' COURSERA: Neural Networks for Machine
  Learning, 2015.

\bibitem{abadi2016tensorflow}
M.~Abadi, P.~Barham, J.~Chen, Z.~Chen, A.~Davis, J.~Dean, M.~Devin,
  S.~Ghemawat, G.~Irving, M.~Isard \emph{et~al.}, ``Tensorflow: A system for
  large-scale machine learning.'' in \emph{OSDI}, vol.~16, 2016, pp. 265--283.

\bibitem{kingma2014adam}
D.~P. Kingma and J.~Ba, ``Adam: A method for stochastic optimization,''
  \emph{arXiv preprint arXiv:1412.6980}, 2014.

\bibitem{drl_long_github_3rd_party}
U.~Lau, ``{rl-collision-avoidance},''
  \url{https://github.com/Acmece/rl-collision-avoidance}, 2019, [Online;
  accessed 10-Sep-2019].

\bibitem{intel_drones}
Intel, ``{Intel Drones Light Up the Sky},''
  \url{https://www.intel.com/content/www/us/en/technology-innovation/aerial-technology-light-show.html},
  2019, [Online; accessed 4-Sep-2019].

\bibitem{airbus_formation}
Airbus, ``{Airbus Commercial Aircraft formation flight: 50-year anniversary},''
  \url{https://www.youtube.com/watch?v=JS6w-DXiZpk}, 2019, [Online; accessed
  4-Sep-2019].

\bibitem{kitano1997robocup}
H.~Kitano, M.~Asada, Y.~Kuniyoshi, I.~Noda, and E.~Osawa, ``Robocup: The robot
  world cup initiative,'' in \emph{Proceedings of the first international
  conference on Autonomous agents}, 1997, pp. 340--347.

\bibitem{finding_nemo_formation}
Pixar, ``{Finding Nemo (School of Fish Scene)},''
  \url{https://www.youtube.com/watch?v=Le13by2WM70}, 2003, [Online; accessed
  4-Sep-2019].

\bibitem{Omidshafiei15_Infotech}
S.~Omidshafiei, A.~akbar Agha-mohammadi, Y.~F. Chen, N.~K. Ure, J.~How,
  J.~Vian, and R.~Surati, ``{MAR-CPS: Measurable Augmented Reality for
  Prototyping Cyber-Physical Systems},'' in \emph{AIAA Infotech@ Aerospace},
  2015.

\bibitem{Campbell13_NIPS}
T.~Campbell, M.~Liu, B.~Kulis, J.~P. How, and L.~Carin, ``Dynamic clustering
  via asymptotics of the dependent dirichlet process mixture,'' in
  \emph{Advances in Neural Information Processing Systems}, 2013, pp. 449--457.

\bibitem{liu2016ssd}
W.~Liu, D.~Anguelov, D.~Erhan, C.~Szegedy, S.~Reed, C.-Y. Fu, and A.~C. Berg,
  ``Ssd: Single shot multibox detector,'' in \emph{European conference on
  computer vision}.\hskip 1em plus 0.5em minus 0.4em\relax Springer, 2016, pp.
  21--37.

\bibitem{miller_dynamic_2016}
J.~Miller, A.~Hasfura, S.~Y. Liu, and J.~P. How, ``Dynamic arrival rate
  estimation for campus {{Mobility On Demand}} network graphs,'' in \emph{2016
  {{IEEE}}/{{RSJ International Conference}} on {{Intelligent Robots}} and
  {{Systems}} ({{IROS}})}, Oct. 2016, pp. 2285--2292.

\bibitem{everett_robot_2017}
M.~Everett, ``Robot designed for socially acceptable navigation,'' Master
  Thesis, MIT, Cambridge, MA, USA, Jun. 2017.

\end{thebibliography}
\balance


\begin{IEEEbiography}[{\includegraphics[width=1in,height=1.25in,clip,keepaspectratio]{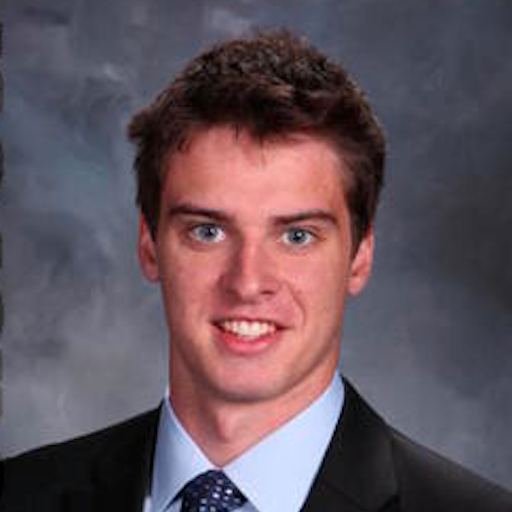}}]{Michael Everett}
is a Postdoctoral Associate at the MIT Department of Aeronautics and Astronautics and conducts research in the Aerospace Controls Laboratory.
He received the PhD (2020), SM (2017), and SB (2015) degrees from MIT in Mechanical Engineering.
His research addresses fundamental gaps in the connection of machine learning and real mobile robotics, with recent emphasis on developing the theory of safety/robustness of learned modules.
He was an author of works that won the Best Paper Award on Cognitive Robotics at IROS 2019, the Best Student Paper Award and finalist for the Best Paper Award on Cognitive Robotics at IROS 2017, and finalist for the Best Multi-Robot Systems Paper Award at ICRA 2017.
He has been interviewed live on the air by BBC Radio and his team's robots were featured by Today Show and the Boston Globe.
\end{IEEEbiography}

\begin{IEEEbiography}[{\includegraphics[width=1in,height=1.25in,clip,keepaspectratio]{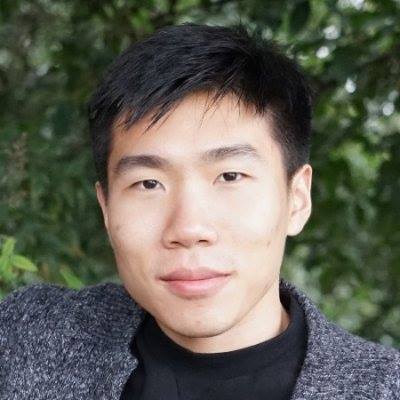}}]{Yu Fan (Steven) Chen} is a research scientist at Facebook Reality Labs, formerly known as Oculus Research. He received a B.A.Sc. degree from the University of Toronto in 2012, and his SM and PhD degrees in Aeronautics and Astronautics from MIT in 2014 and 2017. His research interest spans perception and decision-making for robotics and augmented reality applications. In particular, his current work focuses on self-supervised learning by aggregating information from multiple views and enforcing geometric consistency. His earlier works on multi-agent collision avoidance have won the Best Student Paper Award at IROS 2017, and finalist for the Best Multi-Robot Systems Paper Award at ICRA 2017.
\end{IEEEbiography}

\begin{IEEEbiography}[{\includegraphics[width=1in,height=1.25in,clip,keepaspectratio]{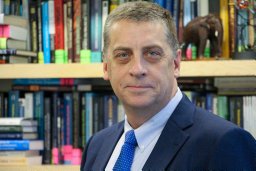}}]{Jonathan P. How} is the Richard C. Maclaurin Professor of Aeronautics and Astronautics at the Massachusetts Institute of Technology.  He received a B.A.Sc. (aerospace) from the University of Toronto in 1987, and his S.M. and Ph.D. in Aeronautics and Astronautics from MIT in 1990 and 1993, respectively, and then studied for 1.5 years at MIT as a postdoctoral associate. Prior to joining MIT in 2000, he was an assistant professor in the Department of Aeronautics and Astronautics at Stanford University.  Dr. How was the editor-in-chief of the IEEE Control Systems Magazine (2015-19) and is an associate editor for the AIAA Journal of Aerospace Information Systems and the IEEE Transactions on Neural Networks and Learning Systems. He was an area chair for International Joint Conference on Artificial Intelligence (2019) and will be the program vice-chair (tutorials) for the Conference on Decision and Control (2021).  He was elected to the Board of Governors of the IEEE Control System Society (CSS) in 2019 and is a member of the IEEE CSS Technical Committee on Aerospace Control and the Technical Committee on Intelligent Control. He is the Director of the Ford-MIT Alliance and was a member of the USAF Scientific Advisory Board (SAB) from 2014-17. His research focuses on robust planning and learning under uncertainty with an emphasis on multiagent systems, and he was the planning and control lead for the MIT DARPA Urban Challenge team.  His work has been recognized with multiple awards, including the 2020 AIAA Intelligent Systems Award. He is a Fellow of IEEE and AIAA.   
\end{IEEEbiography}

\EOD

\end{document}